\pdfoutput=1
\documentclass{article}
\PassOptionsToPackage{numbers}{natbib}
\usepackage[final]{neurips_2022}

\usepackage[utf8]{inputenc} 
\usepackage[T1]{fontenc}    
\usepackage{hyperref}       
\usepackage{url}            
\usepackage{booktabs}       
\usepackage{amsfonts}       
\usepackage{nicefrac}       
\usepackage{microtype}      
\usepackage{xcolor}         

\usepackage{amsmath}
\usepackage{amsfonts}
\usepackage{bm}

\def\eqref#1{equation~\ref{#1}}
\def\Eqref#1{Equation~\ref{#1}}

\def\1{\bm{1}}

\DeclareMathAlphabet{\mathsfit}{\encodingdefault}{\sfdefault}{m}{sl}
\SetMathAlphabet{\mathsfit}{bold}{\encodingdefault}{\sfdefault}{bx}{n}

\DeclareMathOperator*{\argmax}{arg\,max}

\usepackage{hyperref}
\usepackage{url}
\usepackage{siunitx}
\usepackage{booktabs}
\usepackage{wrapfig,lipsum,booktabs}
\usepackage{multirow}
\usepackage{graphicx}
\usepackage{cleveref}
\usepackage{tablefootnote}
\usepackage[normalem]{ulem}
\usepackage{xcolor}
\usepackage[flushleft]{threeparttable}
\usepackage{xspace}
\usepackage{float}
\usepackage{caption}
\usepackage{subcaption}
\usepackage{scrextend}
\usepackage{mathtools}
\usepackage{enumitem}
\usepackage{makecell}
\usepackage{bm}
\usepackage{algorithm}
\usepackage{algorithmic}

\usepackage{bbding}

\title{Reinforcement Learning with\\ Automated Auxiliary Loss Search}

\author{
Tairan He$^{1}$\thanks{The work was conducted during Tairan He’s internship at Microsoft Research.} \quad Yuge Zhang$^{2}$ \quad Kan Ren$^2$\thanks{The corresponding author is Kan Ren.} \quad Minghuan Liu$^1$ \quad\\
\textbf{Che Wang}$^3$ \quad \textbf{Weinan Zhang}$^1$ \quad \textbf{Yuqing Yang}$^2$ \quad \textbf{Dongsheng Li}$^2$\\
$^1$Shanghai Jiao Tong University \quad $^2$Microsoft Research Asia \quad $^3$New York University\\
\texttt{whynot@sjtu.edu.cn \quad kan.ren@microsoft.com}\\
}

\newcommand{\method}{\textsc{A2LS}\xspace}

\newcommand{\minisection}[1]{\vspace{1pt}\noindent\textbf{#1}}

\begin{document}

\maketitle

\begin{abstract}
A good state representation is crucial to solving complicated reinforcement learning (RL) challenges. Many recent works focus on designing auxiliary losses for learning informative representations. Unfortunately, these handcrafted objectives rely heavily on expert knowledge and may be sub-optimal.
In this paper, we propose a principled and universal method for learning better representations with auxiliary loss functions, named Automated Auxiliary Loss Search (\method), which automatically searches for top-performing auxiliary loss functions for RL. Specifically, based on the collected trajectory data, we define a general auxiliary loss space of size $7.5 \times 10^{20}$ and explore the space with an efficient evolutionary search strategy. Empirical results show that the discovered auxiliary loss (namely, \texttt{A2-winner}) significantly improves the performance on both high-dimensional (image) and low-dimensional (vector) unseen tasks with much higher efficiency, showing promising generalization ability to different settings and even different benchmark domains. We conduct a statistical analysis to reveal the relations between patterns of auxiliary losses and RL performance. The codes and supplementary materials are available at \url{https://seqml.github.io/a2ls}.
\end{abstract}

\section{Introduction}
\label{sec:introduction}
Reinforcement learning (RL) has achieved remarkable progress in games~\citep{mnih2013dqn,vinyals2019starcraft,guan2022perfectdou}, financial trading~\citep{fang2021universal} and robotics~\citep{gu2017deep}.
However, in its core part, without designs tailored to specific tasks, general RL paradigms are still learning implicit representations from critic loss (value predictions) and actor loss (maximizing cumulative reward). 
In many real-world scenarios where observations are complicated (e.g., images) or incomplete (e.g., partial observable),
training an agent that is able to extract informative signals from those inputs becomes incredibly sample-inefficient. 

Therefore, many recent works have been devoted to obtaining a good state representation, which is believed to be one of the key solutions to improve the efficacy of RL~\citep{laskin2020curl,laskin2020reinforcement}. 
One of the main streams is adding auxiliary losses to update the state encoder. Under the hood, it resorts to informative and dense learning signals in order to encode various prior knowledge and regularization~\citep{shelhamer2016loss}, and obtain better latent representations.
Over the years, a series of works have attempted to figure out the form of the most helpful auxiliary loss for RL. 
Quite a few advances have been made, including observation reconstruction~\citep{yarats2019sac_ae}, reward prediction~\citep{DBLP:conf/iclr/JaderbergMCSLSK17}, environment dynamics prediction~\citep{shelhamer2016loss, de2018integrating, ota2020ofe}, etc. 
But we note two problems in this evolving process: (i) each of the loss designs listed above are obtained through empirical trial-and-errors based on expert designs, thus heavily relying on human labor and expertise; (ii) few works have used the final performance of RL as an optimization objective to directly search the auxiliary loss, indicating that these designs could be sub-optimal.
To resolve the issues of the existing handcrafted solution mentioned above, we decide to automate the process of designing the auxiliary loss functions of RL and propose a principled solution named Automated Auxiliary Loss Search (A2LS). 
A2LS formulates the problem as a bi-level optimization where we try to find the best auxiliary loss, which, to the most extent, helps train a good RL agent. The outer loop searches for auxiliary losses based on RL performance to ensure the searched losses align with the RL objective, while the inner loop performs RL training with the searched auxiliary loss function.
Specifically, A2LS utilizes an evolutionary strategy to search the configuration of auxiliary losses over a novel search space of size $7.5 \times 10^{20}$ that covers many existing solutions. By searching on a small set of simulated \textit{training environments} of continuous control from Deepmind Control suite (DMC)~\citep{tassa2018deepmind}, \method finalizes a loss, namely \texttt{A2-winner}.

To evaluate the generalizability of the discovered auxiliary loss \texttt{A2-winner}, we test \texttt{A2-winner} on a wide set of \textit{test environments}, including both image-based and vector-based (with proprioceptive features like positions, velocities and accelerations as inputs) tasks. Extensive experiments show the searched loss function is highly effective and largely outperforms strong baseline
methods. More importantly, the searched auxiliary loss generalizes well to unseen settings such as (i) different robots of control; (ii) different data types of observation; (iii) partially observable settings; (iv) different network architectures; and (v) even to a totally different discrete control domain (Atari 2600 games~\citep{bellemare2013arcade}).
In the end, we make detailed statistical analyses on the relation between RL performance and patterns of auxiliary losses based on the data of whole evolutionary search process, providing useful insights on future studies of auxiliary loss designs and representation learning in RL. 

\begin{figure}[t]
    \centering
    \includegraphics[width=1\textwidth]{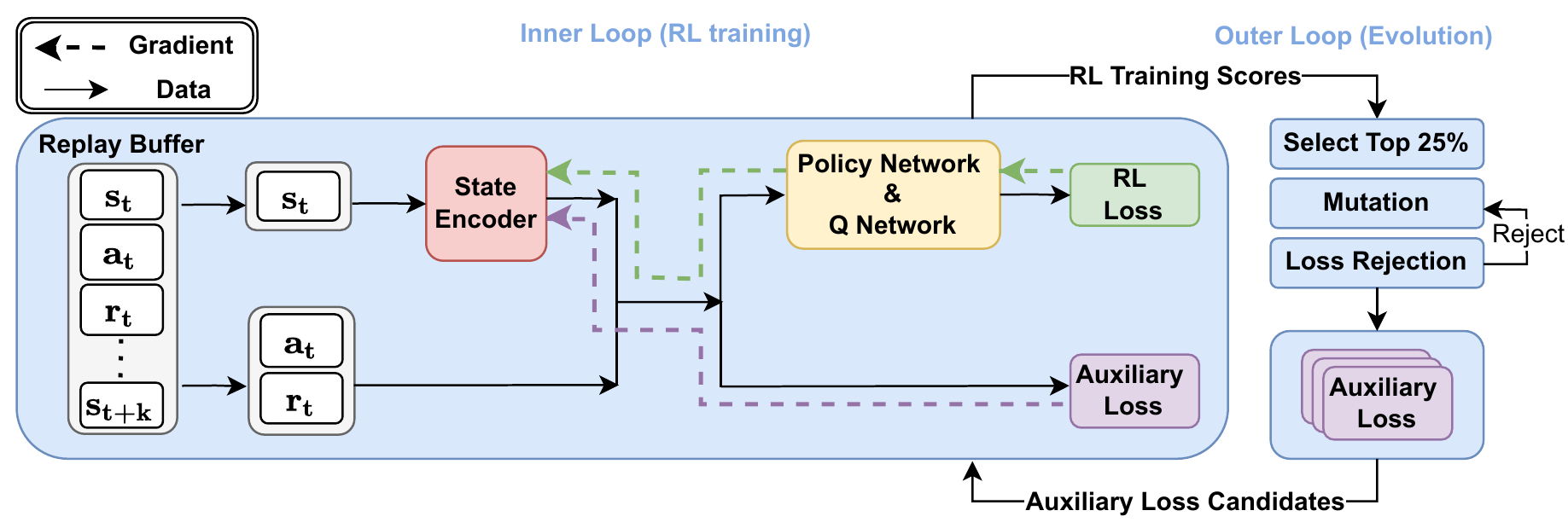}
    \caption{Overview of \method. \method contains an inner loop (left) and an outer loop (right). The inner loop performs an RL training procedure with searched auxiliary loss functions. The outer loop searches auxiliary loss functions using an evolutionary algorithm to select the better auxiliary losses.}
    \label{fig:auxi_flow}
    \vspace{-10pt}
\end{figure}
\section{Problem Formulation and Background}
We consider the standard Markov Decision Process (MDP) $\mathcal{E}$ where the state, action and reward at time step $t$ are denoted as $(s_t, a_t, r_t)$.
The sequence of rollout data sampled by the agent in the episodic environment is $( {s_0, \ldots, s_t, a_t, r_t, s_{t+1}, \cdots, s_{T}} )$, where $T$ represents the episode length. 
Suppose the RL agent is parameterized by $\omega$ (either the policy $\pi$ or the state-action value function $Q$), with a state encoder $g_\theta$ parameterized by $\theta\subseteq\omega$ which plays a key role for representation learning in RL.
The agent is required to maximize its cumulative rewards in environment $\mathcal{E}$  by optimizing $\omega$, noted as $\mathcal{R}(\omega; \mathcal{E})=\mathbb{E}_{\pi}[\sum_{t=0}^{T-1} r_t]$.

In this paper, we aim to find the optimal auxiliary loss function $\mathcal{L}_{\text{Aux}}$ such that the agent can reach the best performance by optimizing $\omega$ under a combination of an arbitrary RL loss function $\mathcal{L}_{\text{RL}}$ together with an auxiliary loss $\mathcal{L}_{\text{Aux}}$. Formally, our optimization goal is:
\begin{equation}
\max_{\mathcal{L}_{\text{Aux}}} \quad \mathcal{R}(\min_{\omega}\mathcal{L}_{\text{RL}}(\omega; \mathcal{E}) + \lambda \mathcal{L}_{\text{Aux}}(\theta; \mathcal{E}); \mathcal{E})~,
\label{eq:nested_optimization}
\end{equation}
where $\lambda$ is a hyper-parameter balancing the relative weight of the auxiliary loss. The left part (inner loop) of \Cref{fig:auxi_flow} illustrates how data and gradients flow in RL training when an auxiliary loss is enabled. 
Some instances of $\mathcal{L}_{\text{RL}}$ and $\mathcal{L}_{\text{Aux}}$ are given in \Cref{appendix: loss_example}. Unfortunately, existing auxiliary losses $\mathcal{L}_{\text{Aux}}$ are handcrafted, which heavily rely on expert knowledge, and may not generalize well in different scenarios as shown in the experiment part. 
To find better auxiliary loss functions for representation learning in RL, we introduce our principled solution in the following section.

\section{Automated Auxiliary Loss Search}
To meet our goal of finding top-performing auxiliary loss functions without expert assignment, we turn to the help of automated loss search, which has shown promising results in the automated machine learning (AutoML) community~\citep{li2019lfs,li2021autoloss_zero,wang2020loss}. Correspondingly, we propose Automated Auxiliary Loss Search (A2LS), a principled solution for resolving the above bi-level optimization problem in \Eqref{eq:nested_optimization}. A2LS resolves the inner problem as a standard RL training procedure; for the outer one, A2LS defines a finite and discrete search space (\Cref{sec:search_space}), and designs a novel evolution strategy to efficiently explore the space (\Cref{sec:search_algorithm}).

\begin{table}[t]
\center
\caption{Typical solution with auxiliary loss and their common elements.}
\resizebox{.9\textwidth}{!}{
\centering
\begin{threeparttable}
\begin{tabular}{c|cccc}
\toprule
\multirow{2}{*}{Auxiliary Loss}  &
\multirow{2}{*}{Operator} &
\multicolumn{3}{c}{Input Elements} \\
\cmidrule(lr){3-5}
& &  Horizon & Source & Target \\
\hline
Forward dynamics~\citep{ota2020ofe,shelhamer2016loss,de2018integrating} & MSE & 1 & $\{s_t, a_t\}$  & $\{s_{t+1}\}$\\
Inverse dynamics & MSE & 1 & $\{a_t, s_{t+1}\}$  & $\{s_t\}$\\
Reward prediction~\citep{DBLP:conf/iclr/JaderbergMCSLSK17,de2018integrating} & MSE & 1 & $\{s_t, a_t\}$ & $\{r_t\}$\\
Action inference~\citep{shelhamer2016loss,de2018integrating} & MSE & 1 & $\{s_t, s_{t+1}\}$  & $\{a_t\}$\\
CURL~\citep{laskin2020curl} & Bilinear & 1 & $\{s_t\}$  & $\{s_t\}$\\
ATC~\citep{stooke2021decoupling}& Bilinear & k & $\{s_t\}$  & $\{s_{t+1}, \cdots, s_{t+k}\}$\\
SPR~\citep{schwarzer2020spr}& N-MSE & k & $\{s_t, a_t, a_{t+1}, \cdots, a_{t+k-1}\}$  & $\{s_{t+1}, \cdots, s_{t+k}\}$\\

\bottomrule
\end{tabular}
\end{threeparttable}
}
\label{tab:existing_in_search_space}
\end{table} 
\subsection{Search Space Design}
\label{sec:search_space}
\begin{figure}[t]
    \centering
    \includegraphics[width=0.9\textwidth]{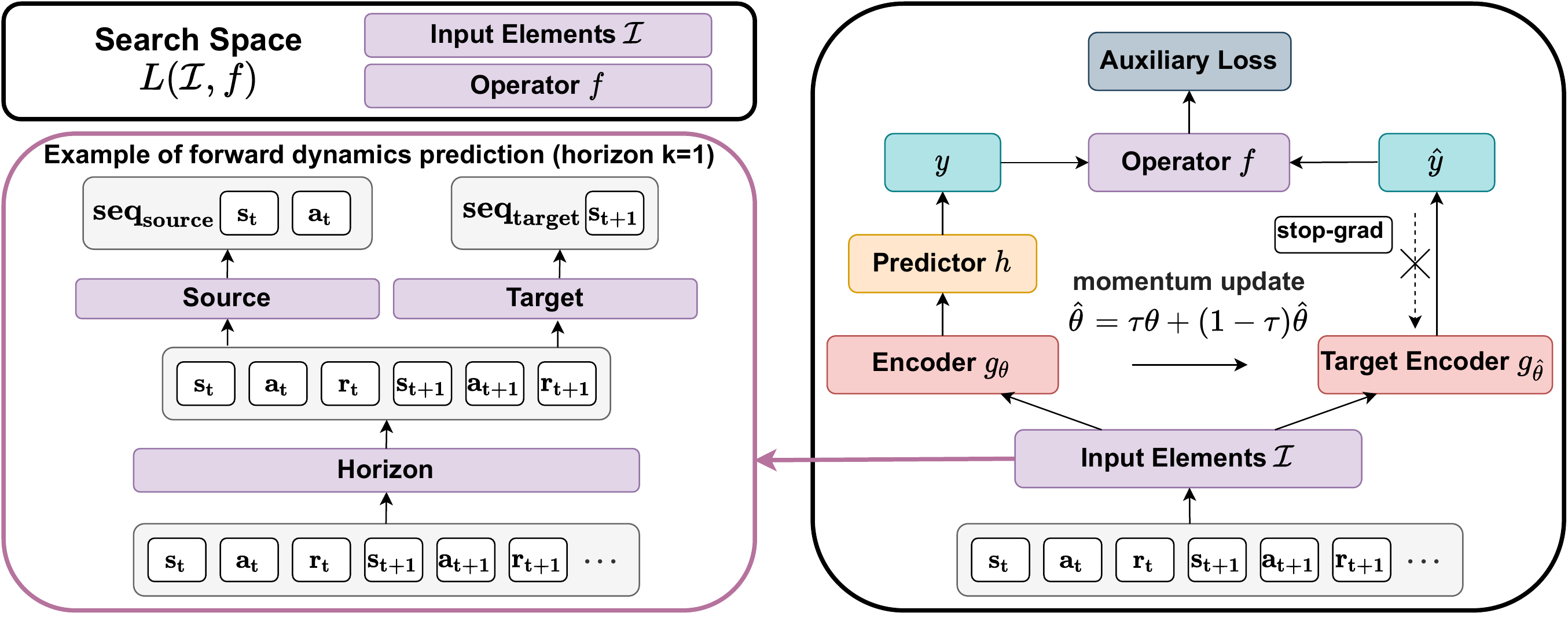}
    \caption{Overview of the search space $\{\mathcal{I}, f\}$ and the computation graph of auxiliary loss functions. $\mathcal{I}$ selects a candidate sequence $\{s_{t}, a_{t}, r_{t}\}_{t=i}^{i+k}$ with \textit{horizon} $k$; then determine a \textit{source} and a \textit{target} as arbitrary subsets of the sequence; an encoder $g_\theta$ first encodes the \textit{source} into latent representations, which is fed into a predictor $h$ to get a prediction $y$; the auxiliary loss is computed over the prediction $y$ and the ground truth $\hat{y}$ that is translated from the \textit{target} by a target encoder $g_{\hat{\theta}}$, using a operator $f$.
    }
    \vspace{-10pt}
    \label{fig:auxi_computation_graph}
\end{figure}

We have argued that almost all existing auxiliary losses require expert knowledge, and we expect to search for a better one automatically. To this end, it is clear that we should design a search space that satisfies the following desiderata.
\begin{itemize}[leftmargin=*]
    \item \textbf{Generalization}: the search space should cover most of the existing handcrafted auxiliary losses to ensure the searched results can be no worse than handcrafted losses;
    \item \textbf{Atomicity}: the search space should be composed of several independent dimensions to fit into any general search algorithm~\cite{configspace} and support an efficient search scheme;
    \item \textbf{Sufficiency}: the search space should be large enough to contain the top-performing solutions.
\end{itemize}
Given the criteria, we conclude and list some existing auxiliary losses in \Cref{tab:existing_in_search_space} and find their commonalities, as well as differences. We realize that these losses share similar components and computation flow. 
As shown in \Cref{fig:auxi_computation_graph}, when training the RL agent, the loss firstly selects a sequence $\{s_{t}, a_{t}, r_{t}\}_{t=i}^{i+k}$ from the replay buffer, when $k$ is called \textit{horizon}. The agent then tries to predict some elements in the sequence (called \emph{target}) based on another picked set of elements from the sequence (called \emph{source}). Finally, the loss calculates and minimizes the prediction error (rigorously defined with \emph{operator}). 
To be more specific, the encoder part $g_\theta$ of the agent, first encodes the \textit{source} into latent representations, which is further fed into a predictor $h$ to get a prediction $y$; the auxiliary loss is computed by the prediction $y$ and the target $\hat{y}$ that is translated from the \textit{target} by a target encoder $g_{\hat{\theta}}$, using an \textit{operator} $f$. The target encoder is updated in an momentum manner as shown in \Cref{fig:auxi_computation_graph} (details are given in \Cref{sec:ablation_target_encoders}).
Formally,
\begin{equation}
\mathcal{L}_{\text{Aux}}(\theta; \mathcal{E}) = f\Big(h\big(g_\theta(\text{seq}_\text{source})\big), g_{\hat{\theta}}(\text{seq}_{\text{target}})\Big)~,
\label{eq:computational_graph}
\end{equation}
where $\text{seq}_\text{source},\text{seq}_\text{target}\subseteq\{s_{t}, a_{t}, r_{t}\}_{t=i}^{i+k}$ are both subsets of the candidate sequence. And for simplicity, we will denote $g_\theta(s_t, a_t, r_t, s_{t+1}, \cdots)$ as short for $[g_\theta(s_t), a_t, r_t, g_\theta(s_{t+1}), \cdots]$ for the rest of this paper (the encoder $g$ only deals with states $\{s_i\}$).
Thereafter, we observe that these existing auxiliary losses differ in two dimensions, i.e., \textit{input elements} and \textit{operator}, where \textit{input elements} are further combined by \textit{horizon}, \textit{source} and \textit{target}.
These differences compose our search dimensions of the whole space. 
We then illustrate the search ranges of these dimensions in detail.

\minisection{Input elements.}
The \textit{input elements} denote all inputs to the loss functions, which can be further disassembled as \textit{horizon}, \textit{source} and \textit{target}. 
Different from previous automated loss search works, the \emph{target} here is not ``ground-truth'' because auxiliary losses in RL have no labels beforehand. 
Instead, both \emph{source} and \emph{target} are generated via interacting with the environment in a self-supervised manner. Particularly, the \emph{input elements} first determine a candidate sequence $\{s_{t}, a_{t}, r_{t}\}_{t=i}^{i+k}$ with \textit{horizon} $k$. Then, it chooses two subsets from the candidate sequence as \textit{source} and \textit{target} respectively. For example, the subsets can be $\{s_t\}, \{s_t,s_{t+1}\}$, or $\{s_t,r_{t+1},a_{t+2}\}, \{s_t,s_{t+1},a_{t+1}\}$, etc. 

\minisection{Operator.} 
Given a prediction $y$ and its target $\hat{y}$, the auxiliary loss is computed by an operator $f$, which is often a similarity measure. 
In our work, we cover all different operators $f$ used by the previous works,
including inner product (Inner)~\citep{he2020moco, stooke2021decoupling}, 
bilinear inner product (Bilinear)~\citep{laskin2020curl},
cosine similarity (Cosine)~\citep{chen2020simCLR}, mean squared error (MSE)~\citep{ota2020ofe, de2018integrating} and normalized mean squared error (N-MSE)~\citep{schwarzer2020spr}. 
Additionally, other works also utilize contrastive objectives, e.g., InfoNCE loss~\citep{oord2018cpc}, incorporating the trick to sample un-paired predictions and targets as negative samples and maximize the distances between them. 
This technique is orthogonal to the five similarity measures mentioned above, so we make it optional and create $5 \times 2 = 10$ different operators in total. 

\minisection{Final design.}
In the light of preceding discussion, with the definition of \textit{input elements} and \textit{operator}, we finish the design of the search space, which satisfactorily meets the desiderata mentioned above. Specifically, the space is \textbf{generalizable} to cover most of the existing handcrafted auxiliary losses; additionally, the \textbf{atomicity} is embodied by the compositionality that all \textit{input elements} work with any \textit{operator}; most importantly, the search space is \textbf{sufficiently} large with a total size of $7.5 \times 10^{20}$ (detailed calculation can be found in \Cref{sec:calculation_complexity}) to find better solutions.

\subsection{Search Strategy}
\label{sec:search_algorithm}
The success of evolution strategies in exploring large,  multi-dimensional search space has been proven in many works~\citep{DBLP:conf/nips/HouthooftEPG18,co2020evolvingRL}. Similarly, A2LS adopts an evolutionary algorithm~\citep{real2019regularized} to search for top-performing auxiliary loss functions over the designed search space. In its essence, the evolutionary algorithm (i) keeps a population of loss function candidates; (ii) evaluates their performance; (iii) eliminates the worst and evolves into a new better population. Note that step (ii) of ``evaluating'' is very costly because it needs to train the RL agents with dozens of different auxiliary loss functions. Therefore, our key technical contribution contains how to further reduce the search cost (\Cref{sec:pruning})  and how to make an efficient search procedure (\Cref{sec:evolution}). 

\subsubsection{Search Space Pruning}
\label{sec:pruning}
In our preliminary experiment, we find out the dimension of \textit{operator} in the search space can be simplified.
In particular, MSE outperforms all the others by significant gaps in most cases. So we effectively prune other choices of \textit{operators} except MSE. See \Cref{Ablation Study on Search Space Pruning} for complete comparative results and an ablation study on the effectiveness of search space pruning. 

\subsubsection{Evolution Procedure}
\label{sec:evolution}
Our evolution procedure roughly contains four important components:
(i) \textbf{evaluation and selection}: a population of candidate auxiliary losses is evaluated through an inner loop of RL training, then we select the top candidates for the next evolution stage (i.e., generation); (ii) \textbf{mutation}: the selected candidates mutate to form a new population and move to the next stage; (iii) \textbf{loss rejection}: filter out and skip evaluating invalid auxiliary losses for the next stage; and (iv) \textbf{bootstrapping initial population}: assign more chance to initial auxiliary losses that may contain useful patterns by prior knowledge for higher efficiency. 
The step-by-step evolution algorithm is provided in \Cref{alg:method} in the appendix, and 
an overview of the \method pipeline is illustrated in \Cref{fig:auxi_flow}.
We next describe them in detail.

\minisection{Evaluation and selection.}
At each evolution stage, we first train a population of candidates with a population size $P=100$ by the inner loop of RL training.
The candidates are then sorted by computing the approximated \textit{area under learning curve} (AULC)~\citep{ghiassian2020improving,stadie2015incentivizing}, which is a single metric reflecting both the convergence speed and the final performance~\citep{viering2021shape} with low variance of results. After each training stage, the top-25\% candidates are selected to generate the population for the next stage.
We include an ablation study on the effectiveness of AULC in \Cref{Effectiveness of AULC scores}. 

\begin{figure}[t]
     \centering
     \includegraphics[width=1.0\textwidth]{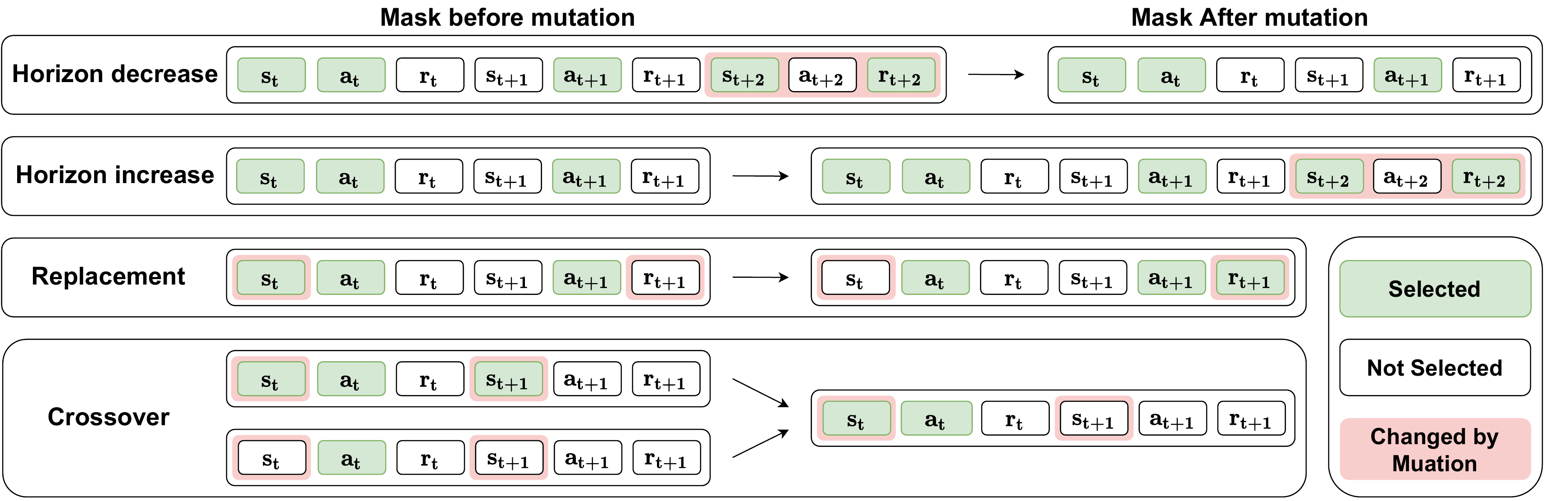}
     \caption{
     Four types of mutation strategy for evolution. We represent both the \textit{source} and the \textit{target} of the input elements as a pair of binary masks, where each bit of the binary mask represents \emph{selected} (green block) by 1 or \emph{not selected} (white block) by 0.
     }
     \label{fig:evolution_operations}
\end{figure}

\minisection{Mutation.}
To obtain a new population of auxiliary loss functions, we propose a novel mutation strategy. First, we represent both the \textit{source} and the \textit{target} of the input elements as a pair of binary masks, where each bit of the mask represents \emph{selected} by 1 or \emph{not selected} by 0. For instance, given a candidate sequence $\{s_t,a_t,r_t,s_{t+1},a_{t+1},r_{t+1}\}$, the binary mask of this subset sequence $\{s_t,a_t,r_{t+1}\}$ is denoted as $110001$.
Afterward, we adopt four types of mutations, also shown in \Cref{fig:evolution_operations}: (i) replacement (50\% of the population): flip the given binary mask with probability $p = \frac{1}{2 \cdot (3k + 3)}$ with the horizon length $k$; (ii) crossover (20\%): generate a new candidate by randomly combining the mask bits of two candidates with the same horizon length in the population; (iii) horizon decrease and horizon increase (10\%): append new binary masks to the tail or delete existing binary masks at the back. (iv) random generation (20\%): every bit of the binary mask is generated from a Bernoulli distribution $\mathcal{B}(0.5)$. 

\minisection{Loss rejection protocol.}
Since the auxiliary loss needs to be differentiable with respect to the parameters of the state encoder, 
we perform a gradient flow check on randomly generated loss functions during evolution and skip evaluating invalid auxiliary losses. 
Concretely, the following conditions must be satisfied to make a valid loss function: (i) having at least one state element in $\text{seq}_{\text{source}}$ to make sure the gradient of auxiliary loss can propagate back to the state encoder; (ii) $\text{seq}_{\text{target}}$ is not empty; (iii) the horizon should be within a reasonable range ($1\leq k\leq10$ in our experiments). If a loss is rejected, we repeat the mutation to fill the population.

\minisection{Bootstrapping initial population.}
To improve the computational efficiency so that the algorithm can find reasonable loss functions quickly, we incorporate prior knowledge into the initialization of the search. Particularly, before the first stage of evolution, we bootstrap the initial population with a prior distribution that assigns high probability to auxiliary loss functions containing useful patterns like dynamics and reward prediction. More implementation details are provided in \Cref{sec:evolution_details}.

\section{Evolution and Searched Results}
\begin{figure}[htbp]
     \centering
     \includegraphics[width=\textwidth]{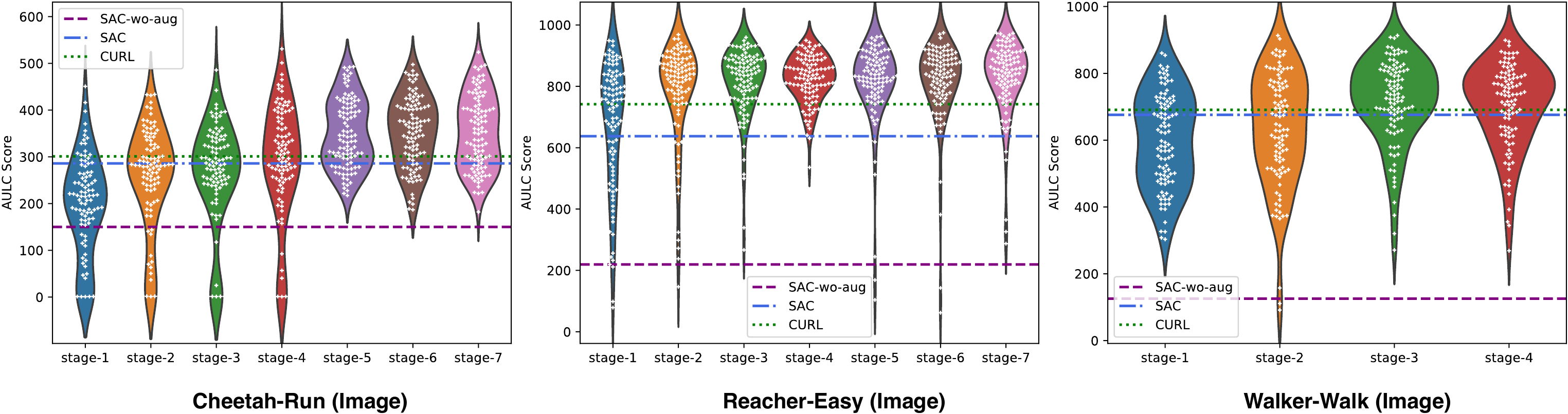}
     \caption{
     Evolution process in the three training (image-based) environments. Every white dot represents a candidate of auxiliary loss, and y-axis shows its corresponding approximated AULC score \citep{ghiassian2020improving,stadie2015incentivizing}. The horizontal lines show the scores of the baselines. 
    The AULC score is approximated with the average evaluation score at 100k, 200k, 300k, 400k, 500k time steps.
    }
     \label{fig:image_evolution_process}
\end{figure} 
As mentioned in \Cref{sec:introduction}, 
we expect to find auxiliary losses that align with the RL objective and generalize well to unseen \textit{test environments}. To do so, we use A2LS to search over a small set of \textit{training environments}, and then test the searched results on a wide range of \textit{test environments}. In this section, we first introduce the evolution on \textit{training environments} and search results.

\subsection{Evolution on Training Environments}
\label{sec: Evolution on Image-based RL}

The \textit{training environments} are chosen as three image-based (observations for agents are images) continuous control tasks in DMC benchmark~\citep{tassa2018deepmind}, Cheetah-Run, Reacher-Easy, and Walker-Walk. 
For each environment, we set the total budget to 16k GPU hours (on NVIDIA P100) and terminate the search when the resource is exhausted. Due to computation complexity, we only run one seed for each inner loop RL training, but we try to prevent such randomness by cross validation (see \Cref{sec:search-reasults}).
We use the same network architecture and hyperparameters config as CURL~\cite{laskin2020curl} (see \Cref{sec:hyper_image} for details) to train the RL agents. 
To evaluate the population during evolution, we measure \method as compared to SAC, SAC-wo-aug,
and CURL, where we randomly crop images from $100 \times 100$ to $84 \times 84$ as data augmentation (the same technique used in CURL\citep{laskin2020curl}) for all methods except SAC-wo-aug.
The whole evolution process on three environments is demonstrated in \Cref{fig:image_evolution_process}.
Even in the early stages (e.g., stage 1), some of the auxiliary loss candidates already surpass baselines, indicating the high potential of automated loss search. 
The overall AULC scores of the population continue to improve when more stages come in (detailed numbers are summerized in \Cref{appendix:trend}). Judging from the trend, we believe the performances could improve even more if we had further increased the budget. 

\begin{wrapfigure}{r}{0.37\textwidth}
    \vspace{-12pt}
    \centering
    \includegraphics[width=0.37\textwidth]{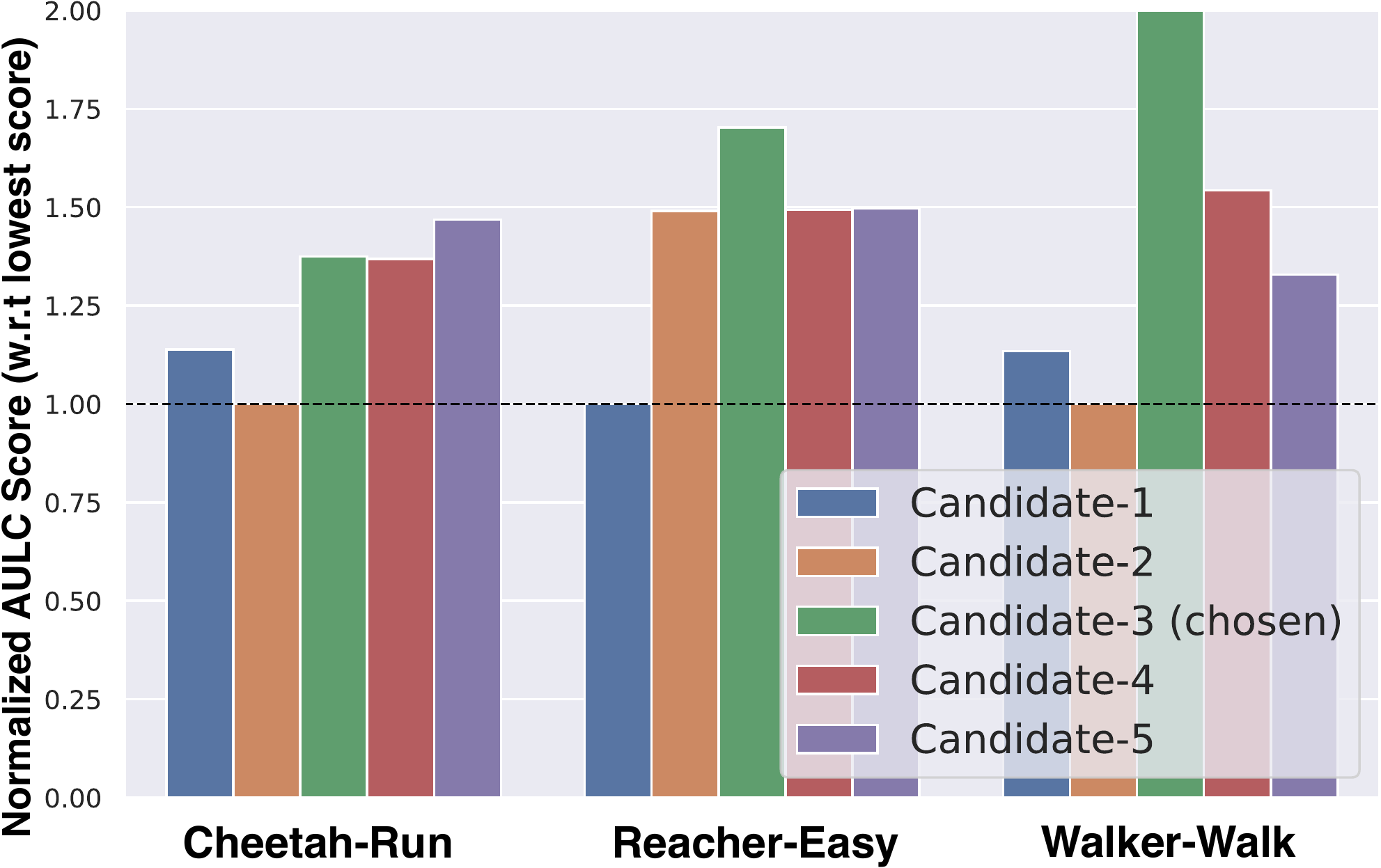}
    \caption{Cross validation on image-based \textit{training environments}.}
    \label{fig:bagging-image}
    \vspace{-12pt}
\end{wrapfigure}
\subsection{Searched Results: \texttt{A2-winner}} 
\label{sec:search-reasults}
Although some candidates in the population have achieved remarkable AULC scores in the evolution (\Cref{fig:image_evolution_process}), they were only evaluated with one random seed in one environment, making their robustness under question. 
To ensure that we find a consistently-useful auxiliary loss, we conduct a cross validation.
We first choose the top 5 candidates of stage-5 of the evolution on Cheetah-Run (detailed top candidates during the whole evolution procedure are provided in \Cref{sec:top_losses}). For each of the five candidates, we repeat the RL training on all three \textit{training environments}, shown in \Cref{fig:bagging-image}. Finally, we mark the best among five (green bar in \Cref{fig:bagging-image}) as our final searched result. We call it \texttt{A2-winner}, which has the following form:
\begin{equation}
\mathcal{L}_{\text{Aux}}(\theta; \mathcal{E}) = 
\|h\big(g_\theta(s_{t+1}, a_{t+1}, a_{t+2}, a_{t+3})\big) - g_{\hat{\theta}}(r_t, r_{t+1}, s_{t+2}, s_{t+3})\|_2~.
\label{eq:method-image}
\end{equation}

\section{Generalization Experiments}
\label{Generalization of method}
\label{sec:experiments}
To verify the effectiveness of the searched results, we conduct various generalization experiments on a wide range of \textit{test environments} in depth.
Implementation details and more ablation studies are given in \Cref{sec:implementation_details} and \Cref{Further Experiment Results}.

\minisection{Generalize to unseen image-based tasks.} 
We first investigate the generalizability of \texttt{A2-winner} to unseen image-based tasks by training agents with \texttt{A2-winner} on common DMC tasks and compare with model-based and model-free baselines that use different auxiliary loss functions (see \Cref{sec:baselines} for details about baseline methods). 
The results are summarized in \Cref{tab:image_generalization} where \texttt{A2-winner} greatly outperforms other baseline methods on most tasks, including unseen \textit{test environments}. 
This implies that \texttt{A2-winner} is a robust and effective auxiliary loss for image-based continuous control tasks to improve both the efficiency and final performance. 

\begin{table}[tb]
\caption{Episodic rewards (mean \& standard deviation for 10 seeds) on DMC100K (100K time steps) and DMC500K (500K time steps). Note that the optimal score of DMC is 1000 for all environments. The baseline methods are PlaNet~\citep{hafner2019planet}, Dreamer~\citep{hafner2019dreamer}, SAC+AE~\citep{yarats2019sac_ae}, SLAC~\citep{lee2019slac}, image-based SAC~\citep{haarnoja2018sac}. 
Performance values of all baselines are referred to~\cite{laskin2020curl}, 
except for Image SAC.
Learning curves of all 12 DMC environments are included in  \Cref{sec:learning_curves_image}.}
\begin{center}
\resizebox{1\textwidth}{!}{
\begin{threeparttable}
\begin{tabular}{c|ccccccc}
\toprule
\textbf{500K} Steps Scores &   \texttt{A2-winner}   &  CURL$^\S$  & PlaNet$^\S$ & Dreamer$^\S$ & SAC+AE$^\S$ & SLACv1$^\S$ & Image SAC \\
\midrule
Cheetah-Run$^\dag$ & 613 $\pm$ 39 & $518 \pm 28$ & $ 305 \pm 131$ & $570 \pm 253 $ & $550 \pm 34$ & \textbf{640 $\pm$ 19} &  99 $\pm$ 28 \\
Reacher-Easy$^\dag$ & \textbf{938 $\pm$ 46} &  \textbf{$929 \pm 44$}  & $ 210 \pm 390$ & $793 \pm 164 $ & $627 \pm 58$ & - & 312 $\pm$ 132 \\
Walker-Walk$^\dag$ & \textbf{917 $\pm$ 18} & $902 \pm 43$ & $ 351 \pm 58$ & $897 \pm 49 $ & $847 \pm 48$ & $842 \pm 51$ &  76 $\pm$ 44 \\
\hline
Finger-Spin$^*$ & \textbf{983 $\pm$ 4} & $926 \pm 45$  & $ 561 \pm 284$ & $796 \pm 183 $ & $884 \pm 128$ & $673 \pm 92$ &  282 $\pm$ 102 \\
Cartpole-Swingup$^*$ & \textbf{864 $\pm$ 19} &  $841 \pm 45$ & $ 475 \pm 71$ & $762 \pm 27 $ & $735 \pm 63$ & - &  344 $\pm$ 104 \\
Ball in cup-Catch$^*$ & \textbf{970 $\pm$ 8} &  $959 \pm 27$  & $ 460 \pm 380$ & $897 \pm 87 $ & $794 \pm 58$ & $852 \pm 71$ &  200 $\pm$ 114 \\
\toprule
\textbf{100K} Steps Scores \\
\midrule
Cheetah-Run$^\dag$ & \textbf{449 $\pm$ 34} &  $299 \pm 48$  & $ 138 \pm 88$ & $235 \pm 137 $ & $267 \pm 24$ & $319 \pm 56$ &  128 $\pm$ 12 \\
Reacher-Easy$^\dag$ & \textbf{778 $\pm$ 164} &  $538 \pm 223$ & $ 20 \pm 50$ & $314 \pm 155 $ & $274 \pm 14$ & - &  277 $\pm$ 69 \\
Walker-Walk$^\dag$ & \textbf{510 $\pm$ 151} &  $403 \pm 24$  & $ 224 \pm 48$ & $277 \pm 12 $ & $394 \pm 22$ & $361 \pm 73$ &  127 $\pm$ 28 \\
\hline
Finger-Spin$^*$ & \textbf{872 $\pm$ 27} &  $767 \pm 56$   & $ 136 \pm 216$ & $341 \pm 70 $ & $740 \pm 64$ & $693 \pm 141$ & 160 $\pm$ 138 \\
Cartpole-Swingup$^*$ & \textbf{815 $\pm$ 66} &  $582 \pm 146$  & $ 297  \pm 39$ & $326 \pm 27 $ & $311 \pm 11$ & - & 243 $\pm$ 19 \\
Ball in cup-Catch$^*$ & \textbf{862 $\pm$ 167} &  $769 \pm 43$  & $ 0 \pm 0$ & $246 \pm 174 $ & $391 \pm 82$ & $512 \pm 110$ &  100 $\pm$ 90 \\
\bottomrule
\end{tabular}
\begin{tablenotes}
    \item[] $\dag$: \textit{Training environments}. $*$: Unseen \textit{test environments}. $\S$: Results reported in ~\cite{laskin2020curl}.
\end{tablenotes}
\end{threeparttable}
}
\label{tab:image_generalization}
\end{center}
\vspace{-20pt}
\end{table}
\begin{wraptable}{r}{0.6\textwidth}
\vspace{-13pt}
\caption{Mean and Median scores (normalized by human score and random score) achieved by \method and baselines on 26 Atari games benchmarked at 100k time-steps (Atari100k).}
\vspace{-10pt}
\label{tab:atari}
\begin{center}
\resizebox{0.99\linewidth}{!}{
\begin{tabular}{c|cccc|cc}
\toprule
Metric &  \texttt{A2-winner}  &  CURL  &  Eff. Rainbow & DrQ~\citep{kostrikov2020drq} & Random & Human \\
\midrule
Mean Human-Norm'd & \textbf{0.568} & 0.381 & 0.285 & 0.357 & 0.000 & 1.000\\
Median Human-Norm'd & \textbf{0.317} & 0.175 & 0.161 & 0.268 & 0.000 & 1.000 \\
\bottomrule
\end{tabular}}
\end{center}
\end{wraptable}
\minisection{Generalize to totally different benchmark domains.} 
To further verify the generalizability of \texttt{A2-winner} on totally different benchmark domains other than DMC tasks, we conduct experiments on the Atari 2600 Games~\citep{bellemare2013arcade}, where we take Efficient Rainbow~\citep{van2019effrainbow} as the base RL algorithm and add \texttt{A2-winner} to obtain a better state representation.
Results are shown in \Cref{tab:atari} where \texttt{A2-winner} outperforms all baselines, showing strong evidence of the generalization and potential usages of \texttt{A2-winner}. Note that the base RL algorithm used in Atari is a value-based method, indicating that \texttt{A2-winner} generalizes well to both value-based and policy-based RL algorithms.

\begin{table}{r}
\vspace{-15pt}
\caption{Episodic rewards (mean \& standard deviation for 10 seeds) on DMC100K (easy tasks) and DMC1000K (difficult tasks) with vector inputs.} 
\vspace{-10pt}
\label{tab:state_generalization}
\center
\resizebox{0.8\linewidth}{!}{
\begin{threeparttable}
\begin{tabular}{c|ccccc}
\toprule
\textbf{100K} Steps Scores & \texttt{A2-winner} & \texttt{A2-winner-v} &  SAC-Identity  & SAC & CURL\\
\midrule
Cheetah-Run$^\dag$ & \textbf{529 $\pm$ 76} & 472 $\pm$ 30 & 237 $\pm$ 27 & 172 $\pm$ 29 & 190 $\pm$ 32\\
\hline
Finger-Spin$^*$ & 790 $\pm$ 128 & \textbf{837 $\pm$ 52} & 805 $\pm$ 32 & 785 $\pm$ 106 & 712 $\pm$ 83 \\
Finger-Turn hard$^*$ & 272 $\pm$ 149 & 218 $\pm$ 117 & \textbf{347 $\pm$ 150} & 174 $\pm$ 94 & 43 $\pm$ 42 \\
Cartpole-Swingup$^*$ & 866 $\pm$ 24 & \textbf{877 $\pm$ 5} & 873 $\pm$ 10 & 866 $\pm$ 7 & 854 $\pm$ 17\\
Cartpole-Swingup sparse$^*$ &  634 $\pm$ 226 & \textbf{695 $\pm$ 147} & 455 $\pm$ 359 & 627 $\pm$ 307 & 446 $\pm$ 196\\
Reacher-Easy$^*$ & 818 $\pm$ 211 & \textbf{934 $\pm$ 38} &  697 $\pm$ 192 & 874 $\pm$ 87 & 749 $\pm$ 183\\
Walker-Stand$^*$ &  935 $\pm$ 32 & \textbf{948 $\pm$ 7} & 940 $\pm$ 10 & 862 $\pm$ 196 & 767 $\pm$ 104 \\
Walker-Walk$^*$ & \textbf{932 $\pm$ 39} & 906 $\pm$ 78 & 873 $\pm$ 89 & 925 $\pm$ 22 & 852 $\pm$ 64 \\
Walker-Run$^*$ & \textbf{616 $\pm$ 52} & 564 $\pm$ 45 & 559 $\pm$ 34 & 403 $\pm$ 43 & 289 $\pm$ 61 \\
Ball in cup-Catch$^*$ & 964 $\pm$ 7 & \textbf{965 $\pm$ 7} & 954 $\pm$ 12 & 962 $\pm$ 13 & 941 $\pm$ 32 \\
Fish-Upright$^*$ & \textbf{586 $\pm$ 128} & 498 $\pm$ 88 & 471 $\pm$ 62 & 400 $\pm$ 62 & 295 $\pm$ 117 \\
Hopper-Stand$^*$ & 177 $\pm$ 257 & \textbf{311 $\pm$ 177} &  14 $\pm$ 16 & 26 $\pm$ 40 & 6 $\pm$ 3  \\
\bottomrule
\toprule
\textbf{1,000K} Steps Scores &  \texttt{A2-winner-v} & \texttt{A2-winner}  &  SAC-Identity  & SAC & CURL\\
\midrule
Quadruped-Run$^\dag$ & \textbf{863 $\pm$ 50} & 838 $\pm$ 58 & 345 $\pm$ 157 & 707 $\pm$ 148 & 497 $\pm$ 128 \\
Hopper-Hop$^\dag$ &  213 $\pm$ 31 &  \textbf{278 $\pm$ 106} & 121 $\pm$ 51 & 134 $\pm$ 93 & 60 $\pm$ 22  \\
\hline
Pendulum-Swingup$^*$ & 200 $\pm$ 322 & \textbf{579 $\pm$ 410} &  506 $\pm$ 374 & 379 $\pm$ 391 & 363 $\pm$ 366  \\
Humanoid-Stand$^*$ & \textbf{329 $\pm$ 35} & 286 $\pm$ 15 & 9 $\pm$ 2 & 7 $\pm$ 1 & 7 $\pm$ 1 \\
Humanoid-Walk$^*$ & \textbf{311 $\pm$ 36} & 299 $\pm$ 55 & 16 $\pm$ 28 & 2 $\pm$ 0 & 2 $\pm$ 0 \\
Humanoid-Run$^*$ &  75 $\pm$ 37 & \textbf{88 $\pm$ 2} & 1 $\pm$ 0 & 1 $\pm$ 0 & 1 $\pm$ 0 \\
\bottomrule
\end{tabular}
\begin{tablenotes}
    \item[] $\dag$: \textit{Training environments}. $*$: Unseen \textit{test environments}. 
\end{tablenotes}
\end{threeparttable}
}
\vspace{-15pt}
\end{table}
\minisection{Generalize to different observation types.} 
To see whether \texttt{A2-winner} (searched in image-based environments) is able to generalize to the environments with different observation types, we test \texttt{A2-winner} on vector-based (inputs for RL agents are proprioceptive features such as positions, velocities and accelerations) tasks of DMC and list the results in \Cref{tab:state_generalization}. Concretely, we compare \texttt{A2-winner} with SAC-Identity, SAC and CURL, where SAC-Identity does not have state encoder while the others share the same state encoder architecture (See \Cref{sec:implementation_encoder_architectures} and \Cref{sec:ablation_encoder_architectures} for detailed implementations). To our delight, \texttt{A2-winner} still outperforms all baselines in 12 out of 18 environments, showing \texttt{A2-winner} can also benefit RL performance in vector-based observations.
Moreover, the performance gain is particularly significant in more complex environments like Humanoid, where SAC barely learns anything at 1000K time steps. In order to get a deeper understanding of this phenomenon, we additionally visualize the Q loss landscape for both methods in \Cref{sec:landscape}. 

\begin{wrapfigure}{r}{0.45\textwidth}
     \vspace{0pt}
     \centering
     \includegraphics[width=0.45\textwidth]{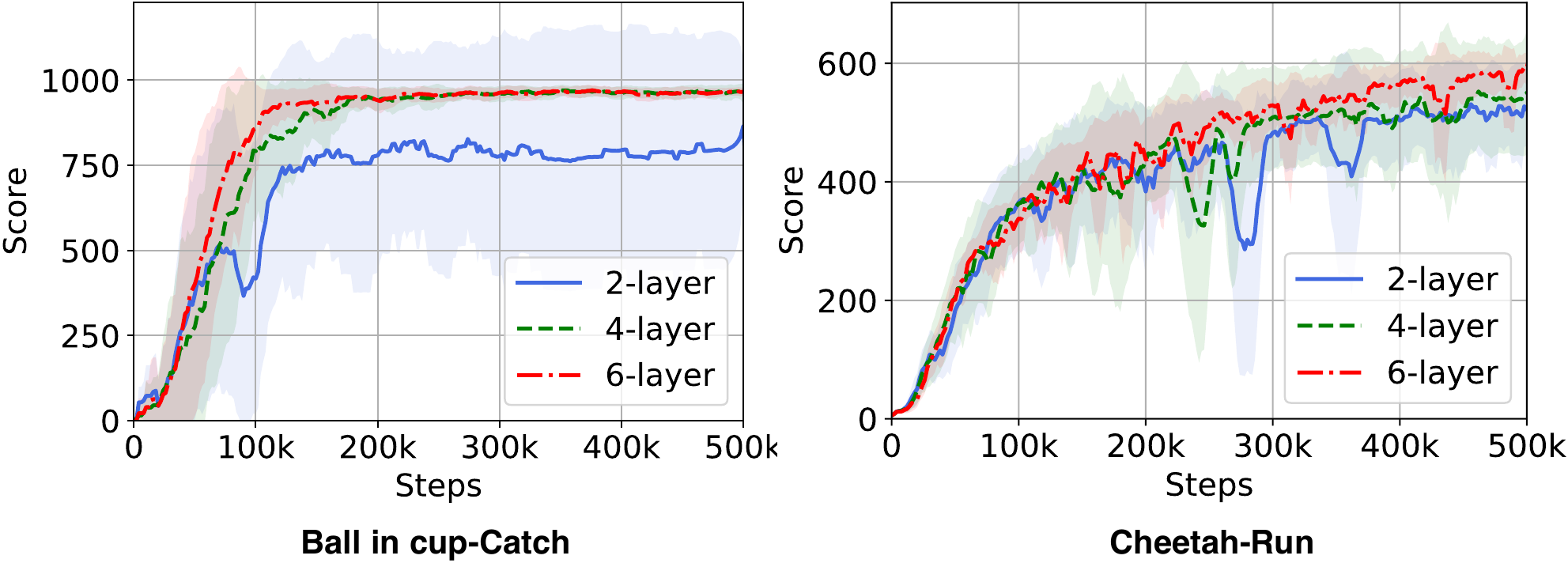}
     \caption{
     Comparison of \texttt{A2-winner} with different depth of convolutional encoder in image-based DMC environments.}
     \label{fig:method_different_encoder_depth_image}
     \vspace{-11pt}
\end{wrapfigure}
\paragraph{Generalize to different hypothesis spaces.}
The architecture of a neural network defines a hypothesis space of functions to be optimized. During the evolutionary search in \Cref{sec: Evolution on Image-based RL}, 
the encoder architecture has been kept static as a 4-layer convolutional neural network.
Since encoder architecture may have a large impact on the RL training process \citep{ota2021training, bjorck2021towards}, 
we test  \texttt{A2-winner} with three encoders with different depth of neural networks. The result is shown in \Cref{fig:method_different_encoder_depth_image}.
Note that even though the auxiliary loss is searched with a 4-layer encoder, the 6-layer convolutional encoder is able to perform better in both two environments. This proves that the auxiliary loss function of \texttt{A2-winner} is able to improve RL performance with a deeper and more expressive image encoder. Moreover, the ranking of RL performance (6-layer > 4-layer > 2-layer) is consistent across the two environments. This shows that the auxiliary loss function of \texttt{A2-winner} does not overfit one specific architecture of the encoder.
\begin{wrapfigure}{r}{0.45\textwidth}
    \centering
    \vspace{-10pt}
    \includegraphics[width=0.45\textwidth]{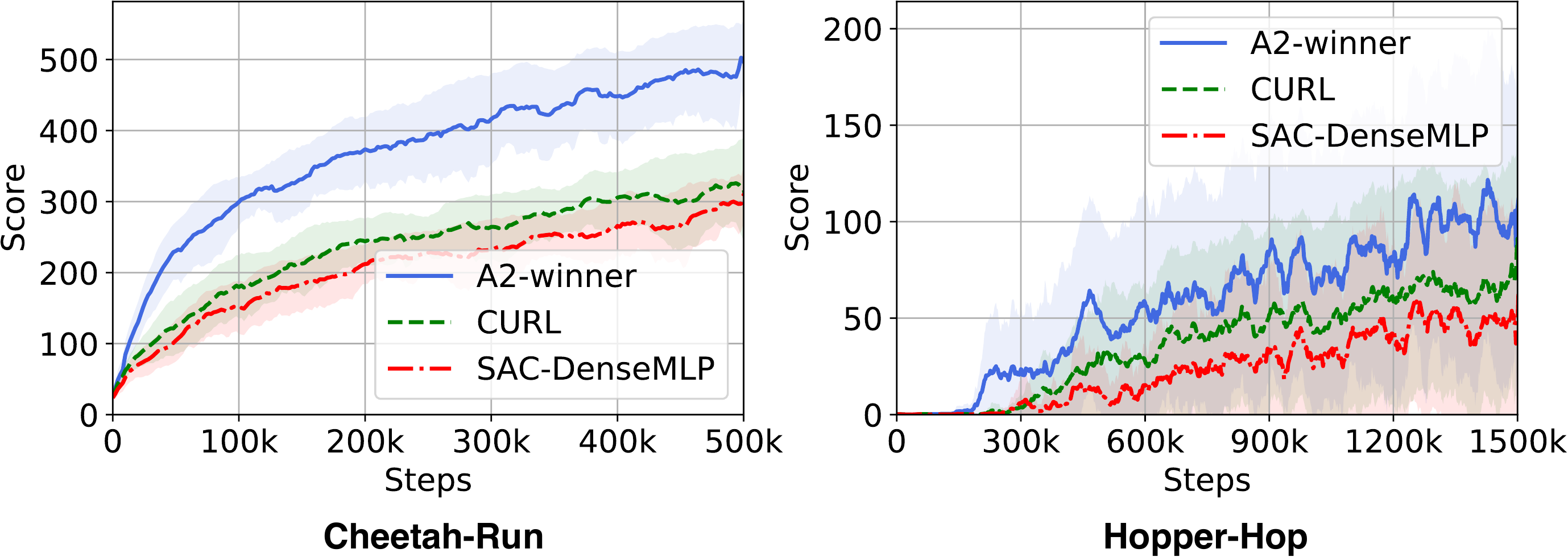}
     \caption{
     Comparison of \texttt{A2-winner} and baselines in partially observable vector-based DMC environments.}
     \label{fig:method_pomdp}
    \vspace{-10pt}
\end{wrapfigure}
\paragraph{Generalize to partially observable scenarios.}
Claiming the generality of a method based on conclusions drawn just on fully observable environments like DMC is very dangerous. 
Therefore, we conduct an ablation study on the Partially Observable Markov Decision Process (POMDP) setting to see whether \texttt{A2-winner} is able to perform well in POMDP. We random mask 20\% of the state dimensions (e.g., 15 dimensions -> 12 dimensions) to form a POMDP environment in DMC. As demonstrated in \Cref{fig:method_pomdp}, \texttt{A2-winner} consistently outperforms CURL and SAC-DenseMLP in the POMDP setting in Hopper-Hop and Cheetah-Run, showing that \texttt{A2-winner} is not only effective in fully observable environments but also partially observable environments. 

\minisection{To search or not?}
As shown above, the searched result \texttt{A2-winner} can generalize well to all kinds of different settings. A natural question here is, however, for a new type of domain, why not perform a new evolution search, instead of simply using the previously searched result?
To compare these two solutions, we conduct another evolutionary search similar to \Cref{sec: Evolution on Image-based RL} but replaced the three image-based tasks with three vector-based ones (marked by $\dag$ in \Cref{tab:state_generalization}) from scratch. More details are summarized in \Cref{Evolution on Vector-based RL}. We name the searched result as ``\texttt{A2-winner-v}''. As shown in \Cref{tab:state_generalization}, \texttt{A2-winner-v} is a very strong-performing loss for vector-based tasks, even stronger than \texttt{A2-winner}.
Actually, \texttt{A2-winner-v} is able to outperform baselines in 16 out of 18 environments (with 15 unseen \textit{test environments}), while \texttt{A2-winner} only outperforms baselines in 12 out of 18 environments.
However, please note that it costs another 5k GPU hours (on NVIDIA P100) to search for \texttt{A2-winner-v} while there is no additional cost to directly use \texttt{A2-winner}. It is a trade-off between lower computational cost and better performance. 
\vspace{-5pt}
\section{Analysis of Auxiliary Loss Functions}
\label{Analysis of Auxiliary Loss Functions}
\vspace{-5pt}
In this section, we analyze all the loss functions we have evaluated during the evolution procedure as a whole dataset in order to gain some insights into the role of auxiliary loss in RL performance. By doing so, we hope to shed light on future auxiliary loss designs.
We will also release this ``dataset'' publicly to facilitate future research.

\minisection{Typical patterns.}
We say that an auxiliary loss candidate has a certain pattern if the pattern's \textit{source} is a subset of the candidate's \textit{source}, and the pattern's \textit{target} is a subset of the candidate's \textit{target}. 
For instance, a loss candidate of $\bm{\{s_t, a_t\} \rightarrow \{s_{t+1}, s_{t+2}\}}$ has the pattern $\{s_t, a_t\} \rightarrow \{s_{t+1}\}$, and does not have the pattern $\{a_t, s_{t+1}\} \rightarrow \{s_t\}$. We then try to analyze whether a certain pattern is helpful to representation learning in RL in expectation.

Specifically, we analyze the following patterns: (i) forward dynamics $\{s_t, a_t\} \rightarrow \{s_{t+1}\}$; (ii) inverse dynamics $\{a_t, s_{t+1}\} \rightarrow \{s_t\}$; (iii) reward prediction $\{s_t, a_t\} \rightarrow \{r_t\}$; (iv) action inference $\{s_t, s_{t+1}\} \rightarrow \{a_t\}$ and (v) state reconstruction in the latent space $\{s_t\} \rightarrow \{s_{t}\}$. For each of these patterns, we categorize all the loss functions we have evaluated into (i) \textit{with} or (ii) \textit{without} this pattern. 
We then calculate the average RL performances of these two categories, summarized in \Cref{tab:analyze_losses}. 
Some interesting observations are as follows. 
\vspace{-6pt}
\begin{itemize}[leftmargin=20pt]
    \item[(i)] Forward dynamics is helpful in most tasks and improves RL performance on Reacher-Easy (image) and Cheetah-Run (vector) significantly (p-value$<$0.05).
    \item[(ii)] State reconstruction in the latent space improves RL performance in image-based tasks but undermines vector-based tasks. The improvements in image-based tasks could be attributed to the combination of augmentation techniques, which, combined with reconstruction loss, enforces the extraction of meaningful features. In contrast, no augmentation is used in the vector-based setting, and thus the encoder learns no useful representations. This also explains why CURL performs poorly in vector-based experiments.
    \item[(iii)] In the vector-based setting, some typical human-designed patterns (e.g., reward prediction, inverse dynamics, and action inference) can be very detrimental to RL performance, implying that some renowned techniques in loss designs might not work well under atypical settings.
\end{itemize}
\vspace{-6pt}

\begin{table}[t]
\caption{Statistical analysis on auxiliary loss functions. The number reported is the difference of the expected RL score when the auxiliary losses \textit{have} one pattern compared to those \textit{do not have}. The corresponding p-value from the t-test is also reported. Positive numbers indicate that this pattern is beneficial. If the performance gain is statistically significant, the number is marked with the asterisk, indicating it is very likely to be helpful. 
Negative numbers indicate this pattern is detrimental. 
}
\begin{subtable}[t]{\textwidth}
\begin{center}
\resizebox{\textwidth}{!}{
\begin{threeparttable}
\begin{tabular}{c|ccccc}
\toprule
& \multicolumn{5}{c}{The score difference between average performances w/ and w/o typical patterns (w/ - w/o) }\\
\hline
 &  Forward dynamics  &  Inverse dynamics  & Reward prediction  & Action inference & State reconstruction \\
\hline
Cheetah-Run (Image) & $+1.28^{  } ~~~~~$ & $-3.51^{  } ~~~~~$ & $-31.16^{ ** } ~~$ & $-75.95^{ ** } ~~$ & $+42.44^{ ** } ~~$ \\
Reacher-Easy (Image) & $+28.25^{ * } ~~~$ & $+8.36^{  } ~~~~~$ & $+37.80^{ ** } ~~$ & $+3.35^{  } ~~~~~$ & $+70.72^{ ** } ~~$ \\
Walker-Walk (Image) & $+22.20^{  } ~~~~$ & $-48.59^{ ** } ~~$ & $-8.11^{  } ~~~~~$ & $+29.86^{ * } ~~~$ & $+13.93^{  } ~~~~$ \\
Cheetah-Run (Vector) & $+94.18^{ ** } ~~$ & $-23.66^{ ** } ~~$ & $-33.28^{ ** } ~~$ & $-109.33^{ ** } ~$ & $-50.15^{ ** } ~~$ \\
Hopper-Hop (Vector) & $+15.50^{ ** } ~~$ & $-16.47^{ ** } ~~$ & $-11.30^{ * } ~~~$ & $-32.10^{ ** } ~~$ & $-25.67^{ ** } ~~$ \\
Quadruped-Run (Vector) & $-28.07^{  } ~~~~$ & $-18.19^{  } ~~~~$ & $-114.23^{ ** } ~$ & $-105.37^{ ** } ~$ & $-82.06^{ ** } ~~$ \\
\bottomrule
\end{tabular}
\begin{tablenotes}
    \item[] $*$: p-value $<$ 0.05. $**$: p-value $<$ 0.01
\end{tablenotes}
\end{threeparttable}
}
\label{tab:typical_patterns}
\end{center}
\end{subtable}
\begin{subtable}[t]{\textwidth}
\begin{center}
\resizebox{\textwidth}{!}{
\begin{threeparttable}
\begin{tabular}{c|ccc}
\toprule
& \multicolumn{3}{c}{
The score difference between two sets varying the number of elements in source and target
}\\
\hline
 &   \quad  \quad State, $n_{target} > n_{source}$  \quad \quad &   \quad \quad Action, $n_{target} > n_{source}$  \quad \quad &   \quad \quad Reward, $n_{target} > n_{source}$  \quad  \quad \\
\hline
Cheetah-Run (Image) & $+80.09^{ ** } ~~$ & $+13.62^{  } ~~~~$ & $+3.33^{  } ~~~~~$ \\
Reacher-Easy (Image) & $+1.98^{  } ~~~~~$ & $-12.72^{  } ~~~~$ & $+65.66^{ ** } ~~$ \\
Walker-Walk (Image) & $+73.56^{ ** } ~~$ & $+42.22^{ * } ~~~$ & $-41.90^{ * } ~~~$ \\
Cheetah-Run (Vector) & $+188.06^{ ** } ~$ & $-102.62^{ ** } ~$ & $-93.94^{ ** } ~~$ \\
Hopper-Hop (Vector) & $+19.80^{ ** } ~~$ & $-29.70^{ ** } ~~$ & $-5.03^{  } ~~~~~$ \\
Quadruped-Run (Vector) & $+75.17^{ ** } ~~$ & $-4.31^{  } ~~~~~$ & $-46.60^{ * } ~~~$ \\
\bottomrule
\end{tabular}
\begin{tablenotes}
    \item[] $*$: p-value $<$ 0.05. $**$: p-value $<$ 0.01
\end{tablenotes}
\end{threeparttable}
}
\end{center}
\label{tab:prediction_targets}
\end{subtable}
\label{tab:analyze_losses}
\vspace{-15pt}
\end{table}
\minisection{Number of Sources and Targets.}
We further investigate whether it is more beneficial to use a small number of sources to predict a large number of targets ($n_{target} > n_{source}$, e.g., using $s_t$ to predict $s_{t+1},s_{t+2}, s_{t+3}$), or the other way around ($n_{target} < n_{source}$, e.g., using $s_t, s_{t+1}, s_{t+2}$ to predict $s_{t+3}$). 
Statistical results are shown in \Cref{tab:analyze_losses}, where we find that auxiliary losses with more states on the \emph{target} side have a significant advantage over losses with more states on the \emph{source} side. This result echoes recent works~\citep{stooke2021decoupling, schwarzer2020spr}: predicting more states leads to strong performance gains.

\section{Related Work}
\vspace{-10pt}
\label{sec:related_work}
\minisection{Reinforcement Learning with Auxiliary Losses.}
Usage of auxiliary tasks for learning better state representations and  improving the sample efficiency of RL agents, especially on image-based tasks, has been explored in many recent works. 
A number of manually designed auxiliary objectives are shown to boost RL performance, including observation reconstruction~\citep{yarats2019sac_ae}, reward prediction~\citep{DBLP:conf/iclr/JaderbergMCSLSK17}, dynamics prediction~\citep{de2018integrating} and contrastive learning objectives~\citep{laskin2020curl, schwarzer2020spr, stooke2021decoupling}. It is worth noting that most of these works focus on image-based settings, and only a limited number of works study the vector-based setting~\citep{munk2016ml_ddpg, ota2020ofe}. 
Although people may think that vector-based settings can benefit less from auxiliary tasks due to their lower-dimensional state space, we show in our paper that there is still much potential for improving their performance with better learned representations.

Compared to the previous works, we point out two major advantages of our approach. 
(i) Instead of handcrafting an auxiliary loss with expert knowledge, \method automatically searches for the best auxiliary loss, relieving researchers from such tedious work. 
(ii) \method is a principled approach that can be used in arbitrary RL settings. We discover great auxiliary losses that bring significant performance improvement in image-based and the rarely studied vector-based settings.

\minisection{Automated Reinforcement Learning.}
RL training is notoriously sensitive to hyper-parameters and environment changes~\citep{henderson2018thatmatters}. Recently, many works attempted to take techniques in AutoML to alleviate human intervention, for example, hyper-parameter optimization~\citep{espeholt2018impala, paul2019fast, DBLP:conf/nips/XuHHOSS20, DBLP:conf/nips/ZahavyXVHOHSS20self_tuning}, reward search~\citep{faust2019evolvingrewards, DBLP:conf/nips/DiscoveryAuxi} and network architecture search~\citep{DBLP:conf/iclr/designRNA, DBLP:conf/iclr/FrankeKBH21}. 
In contrast to these methods which optimize a new configuration for each environment, we search for auxiliary loss functions that generalize across different settings such as (i) different robots of control; (ii) different data types of observation; (iii) partially observable settings; (iv) different network architectures; (v) different benchmark domains.

\minisection{Automated Loss Design.} 
In the AutoML community, it has become a trend to design good loss functions that can
outperform traditional and handcrafted ones. 
To be specific, to resolve computer vision tasks, AM-LFS~\citep{li2019lfs} defines the loss function search space as a parameterized probability distribution of the hyper-parameters of softmax loss. A recent work, AutoLoss-Zero~\citep{li2021autoloss_zero}, proposes to search loss functions with primitive mathematical operators. 

For RL, existing works focus on searching for a better RL objective, EPG~\citep{DBLP:conf/nips/HouthooftEPG18} and MetaGenRL~\citep{DBLP:conf/iclr/KirschSS20} define the search space of loss functions as parameters of a low complexity neural network. Recently, \citep{co2020evolvingRL} defines the search space of RL loss functions as a directed acyclic graph and discovers two DQN-like regularized RL losses. Note that none of these works investigates auxiliary loss functions, which are crucial to facilitate representation learning in RL and to make RL successful in highly complex environments. To the best of our knowledge, our work is the first attempt to search for auxiliary loss functions that can significantly improve RL performance.

\vspace{-5pt}
\section{Conclusion and Future Work}
\label{sec:conclusion}
\vspace{-5pt}
We present \method, a principled and universal framework for automated auxiliary loss design for RL. 
By searching on \textit{training environments} with this framework, we discover a top-performing auxiliary loss function \texttt{A2-winner} that generalizes well to a diverse set of \textit{test environments}. Furthermore, we present 
an in-depth investigation of the statistical relations between auxiliary loss patterns and RL performance. 
We hope our studies provide insights that will deepen the understanding of auxiliary losses in RL, and shed light on how to make RL more efficient and practical. 
Limitations of our current work lie in that searching requires an expensive computational cost.
In the future, we plan to incorporate more delicate information such as higher-order information \cite{ghorbani2019investigation} of the inner-loop RL training procedure to derive more efficient auxiliary loss search methods.

\clearpage

\bibliographystyle{plain}
\bibliography{ref}

\clearpage

\appendix

\section{Algorithm}
\begin{algorithm}[htb]
    \caption{Automated Auxiliary Loss Search}
    \label{alg:method}
    \begin{algorithmic}[1]
      \STATE {\bfseries Initialization:} Randomly generate (bootstrapping) P auxiliary loss functions $\{ \mathcal{L}_p \}_{\text{stage-}1}^P$ and P parameterized policy $\{ \pi_{{\omega}_p}\}_{\text{stage-}1}^P$;
       
      \FOR{$i = 1, 2, \dotsc N$}
      \STATE Optimize policies $\{ \pi_{{\omega}_p}\}_{\text{stage-}i}^P$ with RL loss $\mathcal{L}_{\text{RL}}$ and corresponding Auxiliary loss $\{ \mathcal{L}_p \}_{\text{stage-}i}^P$.
      \STATE Evaluate performance (AULC scores) of each RL agent $\{\mathcal{E}_p\}_{\text{stage-}i}^P$ and select top T candidates $\{ \mathcal{L}_t \}_{\text{stage-}i}^T$.
      \STATE Apply mutations and loss rejection check (introduced in \Cref{sec:evolution}) on top T candidates $\{ \mathcal{L}_t \}_{\text{stage-}i}^T$ to generate new auxiliary loss candidates $\{ \mathcal{L}_p \}_{\text{stage-}(i+1)}^P$
      \ENDFOR
      \STATE Cross validate (introduced in \Cref{sec:search-reasults}) top-performing candidates during evolution to get the optimal auxiliary loss $\mathcal{L}^*$.
     \RETURN {$\mathcal{L}^*$}
    \end{algorithmic}
\end{algorithm}

\section{Examples of Loss Functions}
\label{appendix: loss_example}
We show examples of existing 
$\mathcal{L}_{\text{RL}}$ and $\mathcal{L}_{\text{Aux}}$ below.
\paragraph{RL loss instances.}
RL losses are the basic objectives for solving RL problems. For example, when solving discrete control tasks, the Deep Q Networks (DQN)~\cite{mnih2013dqn} only fit the Q function, where $\mathcal{L}_{\text{RL}}$ is minimizing the error between $Q_\omega$ and $Q_{\hat{\omega}}$ (target Q network):
\begin{equation}
\mathcal{L}_{\text{RL}}= \mathcal{L}_{\text{RL},Q}(\omega; \mathcal{E})=
\mathbb{E}_{s,r \sim \mathcal{E}, a \sim \pi} (Q_\omega (s_t, a_t) - (r_t + \gamma \max_{a}Q_{\hat{\omega}}(s_{t+1}, a_t)))^2
\label{eq:dqn}
\end{equation}
However, for continuous action space, the agent is always required to optimize a policy function alongside the Q loss as in \cref{eq:dqn}. For instance, Soft Actor Critic (SAC)~\cite{haarnoja2018sac} additionally optimizes the policy by policy gradient like:
\begin{equation}
\begin{aligned}
\mathcal{L}_{\text{RL}} = \mathcal{L}_{\text{RL},Q}+\mathcal{L}_{\text{RL},\pi},\quad 
\mathcal{L}_{\text{RL},\pi}(\omega; \mathcal{E}) = 
\mathbb{E}_{s,r \sim \mathcal{E}, a \sim \pi} (-\min_{i=1,2} Q_{\hat{\omega}_i} (s_t, a_t) + \alpha \log \pi_{\omega}(a_t | s_t))
\end{aligned}
\label{eq:sac}
\end{equation}
\paragraph{Auxiliary loss instances.}
Besides $\mathcal{L}_{\text{RL}}$, adding an auxiliary loss $\mathcal{L}_{\text{Aux}}$ helps to learn informative state representation for the best learning efficiency and final performance. 
For example, auxiliary loss of forward dynamics measures the mean squared error of state in the latent space:
\begin{equation}
\mathcal{L}_{\text{Aux}}(\theta; \mathcal{E}) = 
\|h(g_\theta(s_t), a_t) - g_{\hat{\theta}}(s_{t+1})\|_2,
\label{eq:forward_dynamics}
\end{equation}
where $h$ denotes a predictor network. 
Another instance, Contrastive Unsupervised RL (CURL)~\cite{laskin2020curl} designs the auxiliary loss by contrasitive similarity relations:
\begin{equation}
\mathcal{L}_{\text{Aux}}(\theta; \mathcal{E}) = 
\frac{\exp(g_\theta(s_t')^\top W g_{\hat{\theta}}(s_{t_+}'))}{\exp(g_\theta(s_t')^\top W g_{\hat{\theta}}(s_{t_+}'))  + \sum_{i=0}^{K-1}\exp (g_\theta(s_t')^\top W g_{\hat{\theta}}(s_i'))},
\label{eq:curl}
\end{equation}
where $s_t'$ and $s_{t_+}'$ are states of the same state $s_t$ after different random augmentations, and $W$ is a learned parameter matrix.
\clearpage
\section{Implementation Details}

\label{sec:implementation_details}

\subsection{Architecture}

\subsubsection{State Encoder Architectures}
\label{sec:implementation_encoder_architectures}

\begin{figure}[!htb]
     \centering
     \includegraphics[width=1.0\textwidth]{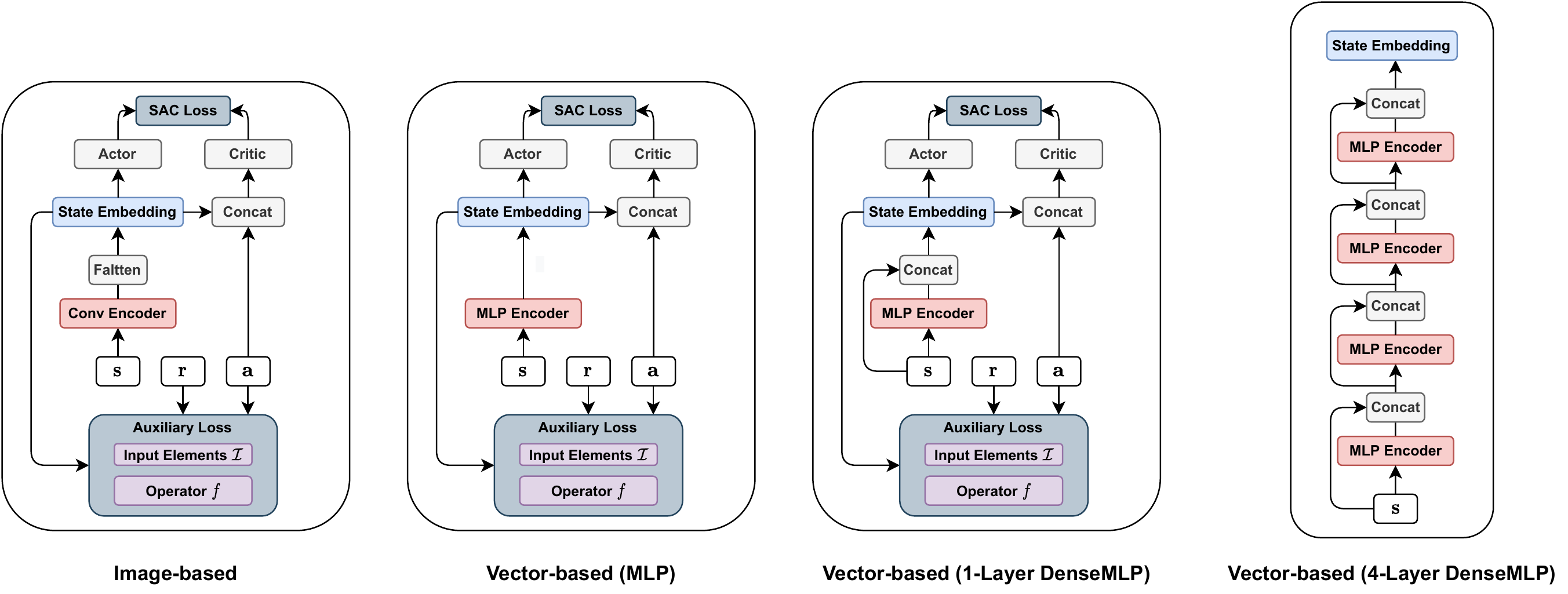}
     \caption{
     Network structures of image-based RL and vector-based RL with auxiliary losses.}
     \label{fig:network_structure}
\end{figure}

In \Cref{fig:network_structure}, we demonstrate the overall architecture when auxiliary loss is used. The architecture is generally identical to architectures adopted in CURL~\citep{laskin2020curl}. ``Image-based'' and ``1-layer DenseMLP'' are the architectures we used in our experiments. ``MLP'' and ``4-layer DenseMLP'' are for ablations. Ablation details are given in \Cref{sec:ablation_encoder_architectures}.

\subsubsection{Siamese Network}
\label{sec:ablation_target_encoders}

For a fair comparison with baseline methods, we follow the same Siamese network structure for representation learning as CURL~\citep{laskin2020curl}. As shown in \Cref{fig:auxi_computation_graph}, when computing targets $\hat{y}$ for auxiliary losses, we map states to state embeddings with a target encoder. We stop gradients from target encoder $\hat{\theta}$ and update $\hat{\theta}$ in the exponential moving averaged (EMA) manner where $\hat{\theta}' = \tau\theta + (1-\tau)\hat{\theta}$. This step, i.e., to freeze the gradients of the target encoder, is necessary when the loss is computed without negative samples. Otherwise, encoders will collapse to generate the same representation for any input. We have verified this in our early experiments. 
\subsection{Loss Operators}

\label{sec:implementation_operators}
\paragraph{Instance Discrimination}
Our implementation is based on InfoNCE loss~\citep{oord2018cpc}:
\begin{equation}
    L = \log \frac{\exp(\phi( y, \hat{y}))}{\exp(\phi(y, \hat{y})) + \sum_{i=0}^{K-1}\exp (\phi(y,y_i))}
\end{equation}
The instance discrimination loss can be interpreted as a log-loss of a K-way softmax classifier whose label is $\hat{y}$.
The difference between discrimination-based loss operators lies in the discrimination objective $\phi$ used to measure agreement between $(y,\hat{y})$ pairs. Inner employs inner product $\phi(y,\hat{y}) = y^{\top}\hat{y}$ while Bilinear employs bilinear product $\phi(y,\hat{y}) =yW\hat{y}$, where $W$ is a learnable parameter matrix. Cosine uses cosine distance $\phi(y,\hat{y}) =\frac{y^{\top}\hat{y}}{\|y\|\cdot\|\hat{y}\|}$ for further matrix calculation. As for losses implemented with cross entropy calculation without negative samples, we only take diagonal elements of matrix $M$ where $M_{i,j} = \phi(y_j, \hat{y_i})$ for cross entropy calculation.

\paragraph{Mean Squared Error}
The implementation of MSE-based loss operators are straightforward. MSE loss operator = $(y-\hat{y})^2$ while normalized MSE = $(\frac{y}{\|y\|} - \frac{\hat{y}}{\|\hat{y}\|})^2$. When combined with negative samples, MSE loss operator (with negative pairs) = $(y- \hat{y})^2 - (y - y_i)^2$ while normalized MSE (with negative pairs) = $(\frac{y}{\|y\|} - \frac{\hat{y}}{\|\hat{y}\|})^2 - (\frac{y}{\|y\|} - \frac{y_i}{\|y_i\|})^2$.

\subsection{Evolution Strategy}
\label{sec:evolution_details}
\paragraph{Horizon-changing Mutations} There are two kinds of mutations that can change horizon length. One is to decrease horizon length. Specifically, we remove the last time step, i.e., $(s_{t+k}, a_{t+k}, r_{t+k})$ if the target horizon length is $k$. The other is to increase horizon length, in which we append three randomly generated bits to the given masks at the end. We do not shorten the horizon when it becomes too small (less than 1) or lengthen the horizon when it is too long (exceeding 10).

\paragraph{Mutating Source and Target Masks}
When mutating a candidate, the mutation on the \emph{source} and the \emph{target} are independent except for horizon change mutation where two masks should either both increase horizon or decrease horizon.

\paragraph{Initialization} At each initialization, we randomly generate 75 auxiliary loss functions (every bit of masks are generated from $\text{Bernoulli}(p)$ where $p=0.5$.) and generate 25 auxiliary loss functions with prior probability, which makes the auxiliary loss have some features like forward dynamics prediction or reward prediction. The prior probability for generating the pattern of forward dynamics prediction is: 
(i) every bit of states from \emph{target} is generated from $\text{Bernoulli}(p)$ where $p=0.2$; 
(ii) every bit of actions from \emph{source} is generated from $\text{Bernoulli}(p)$ where $p=0.8$; 
(iii) every bit of states from \emph{target} is generated by flipping the states of \emph{source}; 
(iv) The other bits are generated from $\text{Bernoulli}(p)$ where $p=0.5$. 
The prior probability for generating the pattern of reward prediction is: 
(i) every bit of rewards from \emph{target} is generated from $\text{Bernoulli}(p)$ where $p=0.8$; 
(ii) Every bit of states and actions from \emph{target} is 0; 
(iii) The other bits are generated from $\text{Bernoulli}(p)$ where $p=0.5$.

\subsection{Training Details}
\subsubsection{Hyper-parameters in the Image-based Setting} 
\label{sec:hyper_image}
We use the same hyper-parameters for \method, SAC-wo-aug, SAC and CURL during the search phase to ensure a fair comparison. When evaluating the searched auxiliary loss, we use a slightly larger setting (e.g., larger batch size) to train RL agents sufficiently. A full list is shown in \Cref{tab:hyper_image}.

\begin{table}[htbp]
\begin{center}
\caption{Hyper-parameters used in image-based environments.}
\label{tab:hyper_image}
\resizebox{0.95\textwidth}{!}{
\begin{tabular}{c|c|c}
\toprule
Hyper-parameter & During Evolution & Final Evaluation of \texttt{A2-winner}  \\
\hline
Random crop & \makecell[t]{False for SAC-wo-aug; \\ True for others} &  True \\
Observation rendering & \makecell[t]{(84, 84) for SAC-wo-aug;\\ (100, 100) for others} & (100, 100) \\
Observation downsampling & (84, 84) &  (84, 84)\\
Replay buffer size & 100000 & 100000\\
Initial steps & 1000 & 1000\\
Stacked frames & 3 & 3\\
Actoin repeat & \makecell[t]{ 4 (Cheetah-Run, Reacher-Easy) \\  2 (Walker-Walk);} &  \makecell[t]{8 (Cartpole-Swingup);\\ 4 (Others) \\ 2 (Walker-Walk, Finger-Spin)  \\ } \\
Hidden units (MLP) & 1024 & 1024 \\
Hidden units (Predictor MLP) & 256 & 256\\
Evaluation episodes & 10 & 10\\
Optimizer & Adam & Adam \\
($\beta_1, \beta_2$) for actor/critic/encoder & (.9, .999) & (.9, .999) \\
($\beta_1, \beta_2$) for entropy $\alpha$ & (.5, .999)  & (.5, .999) \\
Learning rate for actor/critic & 1e-3 & \makecell[t]{2e-4 (Cheetah-Run);\\ 1e-3 (Others)}\\
Learning rate for encoder & 1e-3 & \makecell[t]{3e-3 (Cheetah-Run, Finger-Spin, Walker-Walk);\\ 1e-3 (Others)}\\
Learning for $\alpha$ & 1e-4  & 1e-4\\
Batch size for RL loss & 128 & 512\\
Batch size for auxiliary loss  & 128 & \makecell[t]{ 128 (Walker-Walk)\\256 (Cheetah-Run, Finger-Spin) \\  512 (Others);}\\
Auxiliary Loss multipilier $\lambda$ & 1 & 1\\
Q function EMA $\tau$ & 0.01 & 0.01\\
Critic target update freq & 2 & 2 \\
Convolutional layers & 4 & 4 \\
Number of filters & 32 & 32 \\
Non-linearity & ReLU & ReLU \\ 
Encoder EMA $\tau$ & 0.05 & 0.05 \\
Latent dimension & 50 & 50 \\
Discount $\gamma$ & .99 & .99 \\
Initial temperature & 0.1 & 0.1 \\

\bottomrule
\end{tabular}
}
\end{center}
\end{table}

\subsubsection{Hyper-parameters in the Vector-based Setting} \label{sec:hyper_evolution_state}

We use the same hyper-parameters for \method, SAC-Identity, SAC-DenseMLP and CURL-DenseMLP, shown in \Cref{tab:hyper_state}. Since training in vector-based environments is substantially faster than in image-based environments, there is no need to balance training cost and agent performance. We use this setting for both the search and final evaluation phases.

\begin{table}[htbp]
\begin{center}
\caption{Hyper-parameters used in vector-based environments.}
\label{tab:hyper_state}
\resizebox{0.6\textwidth}{!}{
\begin{tabular}{c|c}
\hline
Replay buffer size & 100000\\
Initial steps & 1000\\
Action repeat & 4 \\
Hidden units (MLP) & 1024\\
Hidden units (Predictor MLP) & 256\\
Evaluation episodes & 10\\
Optimizer & Adam\\
($\beta_1, \beta_2$) for actor/critic/encoder & (.9, .999)\\
($\beta_1, \beta_2$) for entropy $\alpha$ & (.5, .999)\\
Learning rate for actor/critic/encoder & \makecell[t]{2e-4 (Cheetah-Run); \\ 1e-3 (Others) } \\
Learning for $\alpha$ & 1e-4\\
Batch size & 512\\
Auxiliary Loss multipilier $\lambda$ & 1\\
Q function EMA $\tau$ & 0.01\\
Critic target update freq & 2\\
DenseMLP Layers & 1\\
Non-linearity & ReLU\\
Encoder EMA $\tau$ & 0.05\\
Latent dimension of DenseMLP & 40\\
Discount $\gamma$ & .99\\
Initial temperature & 0.1\\

\hline
\end{tabular}
}

\end{center}
\end{table}

\subsection{Baselines Implementation}
\label{sec:baselines}
\paragraph{Image-based Setting}
These following baselines are chosen because they are competitive methods for benchmarking control from pixels.
CURL~\citep{laskin2020curl} is the main baseline to compare within the image-based setting, which is considered to be the state-of-the-art image-based RL algorithm. CURL learns state representations with a contrastive auxiliary loss. 
PlaNet~\citep{hafner2019planet} and Dreamer~\citep{hafner2019dreamer} are model-based methods that generate synthetic rollouts with a learned world model.
SAC+AE~\citep{yarats2019sac_ae} uses a reconstruction auxiliary loss of images to boost RL training.
SLAC~\citep{lee2019slac} leverages forward dynamics to construct a latent space for RL agents. Note that there are two versions of SLAC with different gradient Updates per agent step: SLACv1 (1:1) and SLACv2(3:1). We adopt SLACv1 for comparison since all methods only make one gradient update per agent step.
Image SAC is just vanilla SAC~\citep{haarnoja2018sac} agents with images as inputs.

\paragraph{Vector-based Setting}
As for the vector-based setting, we compare \method with SAC-Identity, SAC and CURL. SAC-Identity is the vanilla vector-based SAC where states are directly fed to actor/critic networks. SAC and CURL use the same architecture of 1-layer densely connected MLP as a state encoder. Note that both \method and baseline methods use the same hyper-parameter reported in \Cref{tab:hyper_state} without additional hyper-parameter tuning.

\clearpage
\section{Additional Experiment Results}
\label{Further Experiment Results}

\subsection{Search Space Pruning}
\paragraph{Results of Search Space Pruning}
\label{Results of Search Space Pruning}
Considering that the loss space is huge, an effective optimization strategy is required.
Directly grid-searching over the whole space is infeasible because of unacceptable computational costs. Thus some advanced techniques such as space pruning and an elaborate search strategy are necessary.
Our search space can be seen as a combination of the space for the input $\mathcal{I}$ and the space for the operator $f$. 
Inspired by AutoML works~\citep{dai2020fbnetv3,ying2019nasbench101} that search for hyper-parameters first and then neural architectures, we approximate the joint search of input and operator in \Cref{eq:nested_optimization} in a two-step manner. The optimal auxiliary loss $\{\mathcal{I}^*, f^*\}$ can be optimized as:
\begin{equation}
\begin{split} 
\max_{\mathcal{L}} \mathcal{R}(M_{\omega^* (\mathcal{L})}; \mathcal{E}) & = \max_{\mathcal{I}, f} \mathcal{R}(M_{\omega^* (\mathcal{I}, f)}; \mathcal{E}) \approx \max_{\mathcal{I}} \mathcal{R}(M_{\omega^* (\mathcal{I}, f^*)}; \mathcal{E}) \\
\text{where} \quad f^* & \approx \argmax_{f} \mathbb{E}_\mathcal{I} [\mathcal{R}(M_{\omega^* (\mathcal{I}, f)}; \mathcal{E})]
\end{split}
\label{eq:pruning-approx}
\end{equation}
To decide the best loss operator, for every $f$ in the \textit{operator} space, we estimate $\mathbb{E}_\mathcal{I} [\mathcal{R}(M_{\omega^* (\mathcal{I}, f)}; \mathcal{E})]$ with a random sampling strategy. 
We run 15 trials for each loss operator to estimate performance expectation. For each of 10 possible $f$ in the search space (5 operators with optional negative samples), we run 5 trials on each of the 3 image-based environments (used in evolution) with the same \textit{input elements} $\{s_t, a_t\}\rightarrow\{s_{t+1}\}$, as we found that forward dynamics is a reasonable representative of our search space with highly competitive performance.
Surprisingly, as summarized in \Cref{tab:similarity_metrics}, the simplest MSE without negative samples outperforms all other loss operators with complex designs. Therefore, this loss operator is chosen for the rest of this paper. 
\begin{table}[htbp]
\caption{Normalized episodic rewards (mean \& standard deviation for 5 seeds) of 3 environments used in evolution
on image-based DMControl500K with different loss operators.}
\begin{center}
\resizebox{0.9\textwidth}{!}{
\begin{tabular}{c|ccccc}
\toprule
Loss operator and discrimination &  Inner &  Bilinear & Cosine & MSE & N-MSE \\
\hline
w/ negative samples & $0.979 \pm 0.344 $ & $0.953 \pm 0.329 $ & $0.872 \pm 0.412$ & $0.124 \pm 0.125$ & $0.933 \pm 0.360$\\
w/o negative samples & $0.669 \pm 0.311 $ & $0.707 \pm 0.299 $ & $0.959 \pm 0.225$ & $\bf{1.000 \pm 0.223}$ & $0.993 \pm 0.229$\\
\bottomrule
\end{tabular}
}
\label{tab:similarity_metrics}
\end{center}
\end{table}

\begin{figure}[b]
     \centering
     \includegraphics[width=0.6\textwidth]{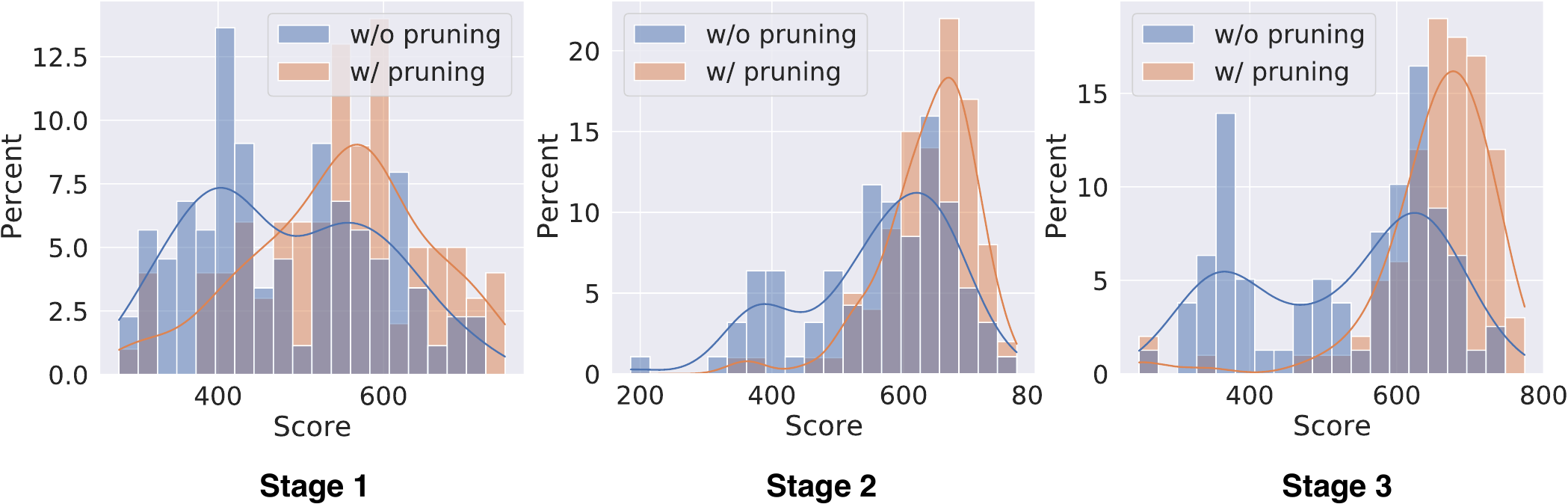}
     \caption{
     Comparison of evolution with and without pruning by performance histogram. 
     }
     \label{fig:hist_pruning}
\end{figure}

\paragraph{Ablation Study on Search Space Pruning}
\label{Ablation Study on Search Space Pruning}
As introduced in \Cref{Results of Search Space Pruning}, we decompose the full search space into \textit{operator} and \textit{input elements}.
Here we try to directly apply the evolution strategy to the whole space without the pruning step. The comparison results are shown in \Cref{fig:hist_pruning}. We can see that pruning improves the evolution process, making it easier to find good candidates. 

\subsection{Learning Curves for \method on Image-based DMControl}
\label{sec:learning_curves_image}
\begin{figure}[h]
     \centering
     \includegraphics[width=1.0\textwidth]{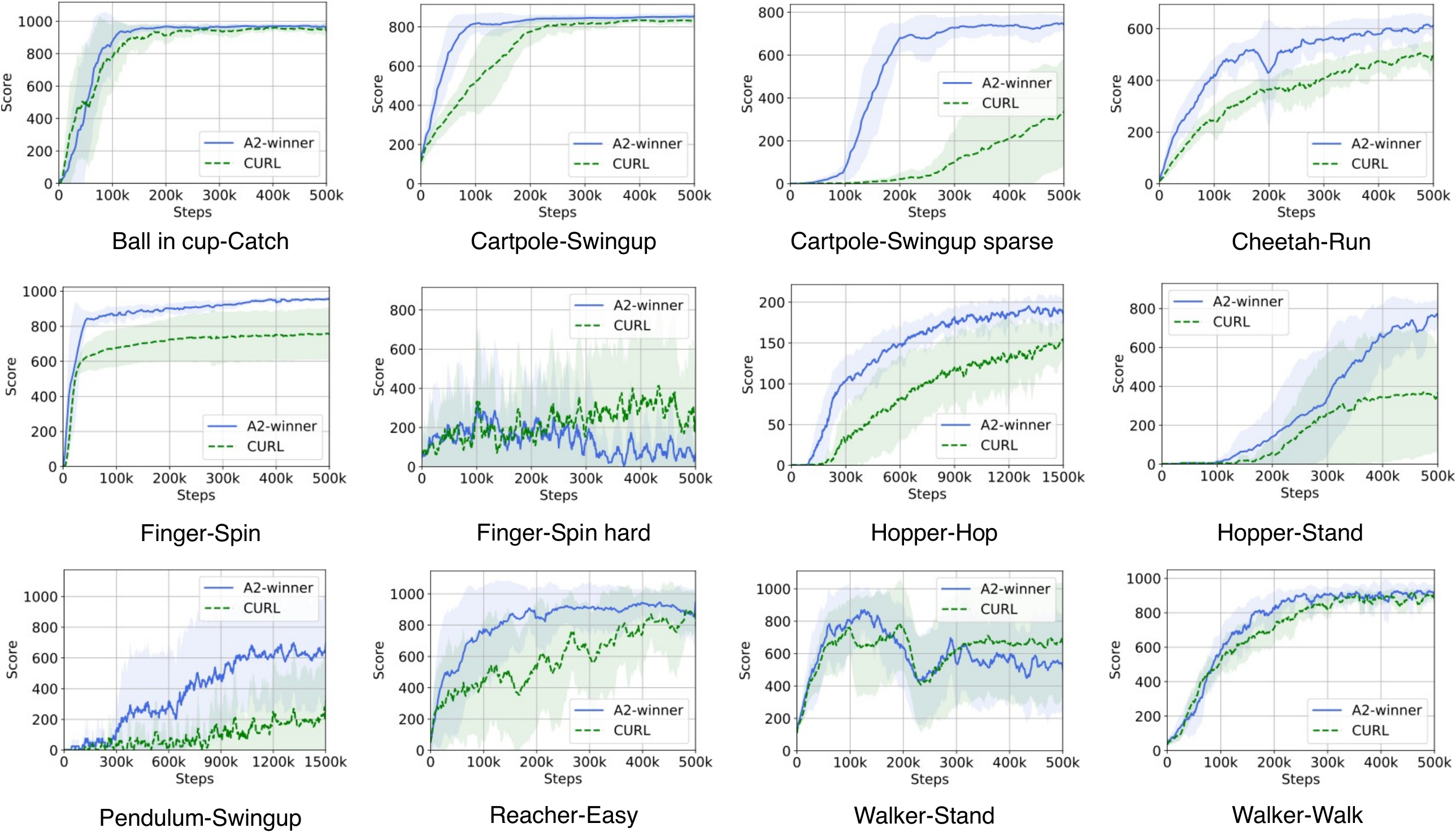}
     \caption{
     Learning curves of \texttt{A2-winner} and CURL on 12 DMC environments. Shadow represents the standard deviation over five random seeds. The curves are uniformly smoothed for visual display. The y-axis represents episodic reward and x-axis represents interaction steps. 
     }
     \label{fig:image_12_curves}
\end{figure}

We benchmark the performance of \method to the best-performing image-based baseline (CURL). As shown in \Cref{fig:image_12_curves}, the sample efficiency of \method outperforms CURL in 10 out of 12 environments. Note that the learning curves of CURL may not match the data in \Cref{tab:image_generalization}. This is because we use the data reported in the CURL paper for tabular while we rerun CURL for learning curves plotting, where we find the performance of our rerunning CURL is sightly below the CURL paper.

\subsection{Effectiveness of AULC scores}
\label{Effectiveness of AULC scores}
\begin{figure}[h]
     \centering
     \includegraphics[width=0.4\textwidth]{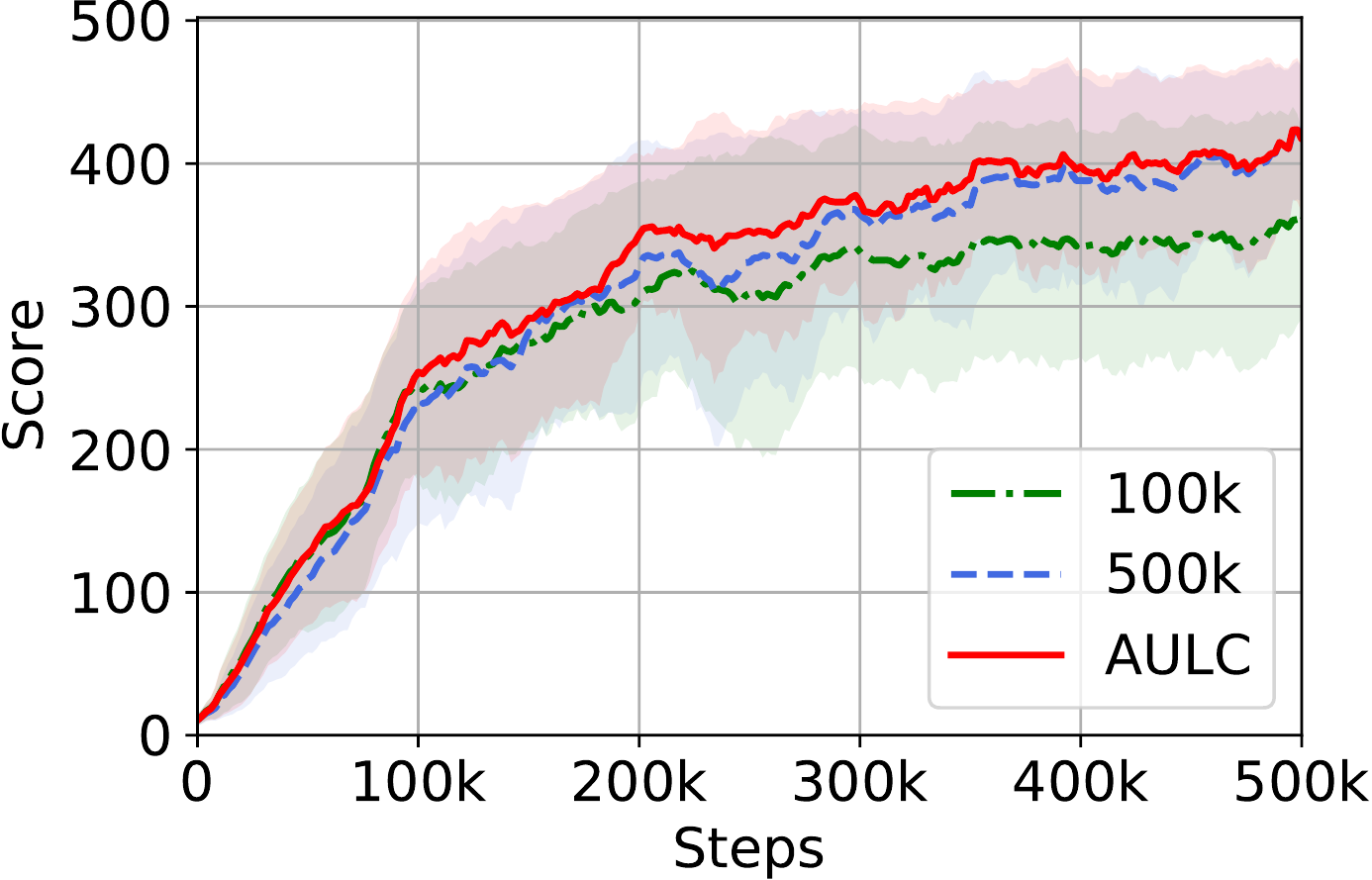}
     \caption{
     Learning curves of top-10 loss candidates selected with different metrics.}
     \label{fig:ablation_evolution_metrics_top10}
\end{figure}

To illustrate why we use the \emph{area under learning curve} (AULC) instead of other metrics, we select top-10 candidates with different evolution metrics. In practice, AULC is calculated as the sum of scores of all checkpoints during training.
\Cref{fig:ablation_evolution_metrics_top10} demonstrates the usage of AULC score could well balance both sample efficiency and final performance. The learning curves of the top-10 candidates selected by AULC score look better than the other two metrics (that select top candidates simply with 100k step score or 500k step score). 

\subsection{Comparing Auxiliary Loss with Data Augmentation}
\label{RL with Auxiliary Loss and Augmentation}
\begin{figure}[h]
     \centering
     \includegraphics[width=0.8\textwidth]{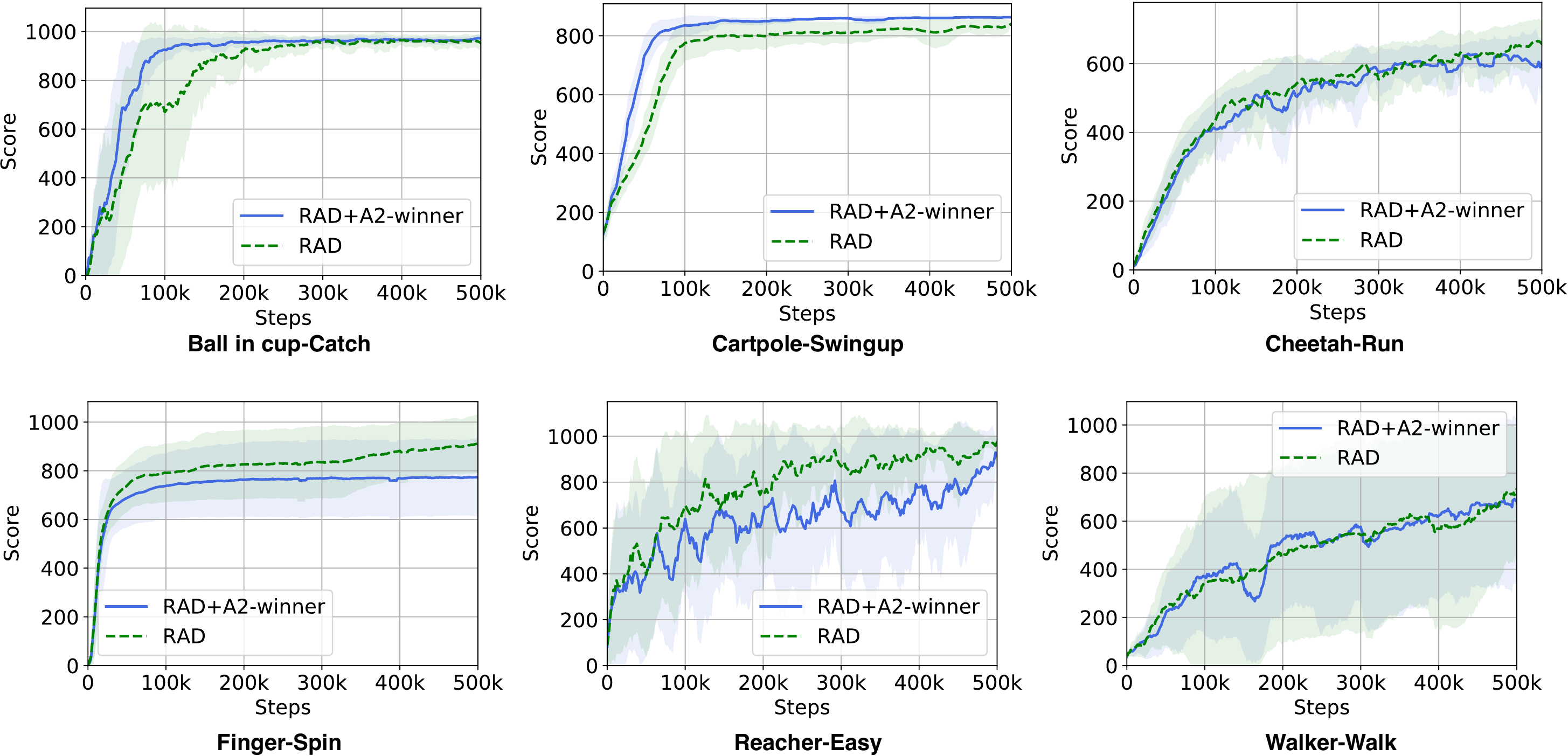}
     \caption{
     Comparison of learning curves of \method and RAD in image-based DMC environments.}
     \label{fig:rad_aarl_image}
\end{figure}

\begin{figure}[h]
     \centering
     \includegraphics[width=0.8\textwidth]{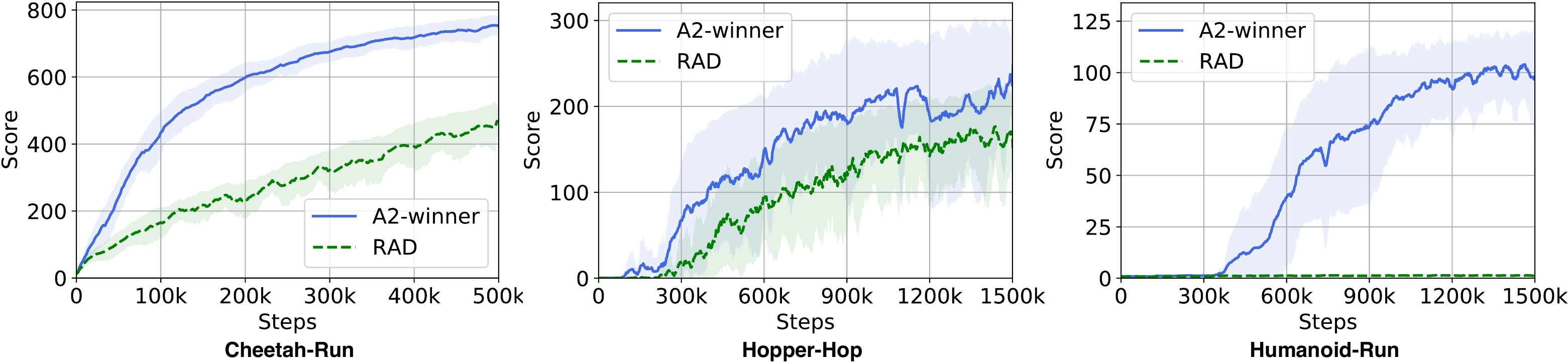}
     \caption{
     Comparison of learning curves of \method and RAD in vector-based DMControl environments.}
     \label{fig:rad_aarl_state}
\end{figure}

Besides auxiliary losses, data augmentation has been shown to be a strong technique for data-efficient RL, especially in image-based environments\citep{laskin2020rad, kostrikov2020drq}. RAD~\citep{laskin2020rad} can be seen as a version of CURL without contrastive loss but with a better image transformation function for data augmentation. 
We compare \texttt{A2-winner} with RAD in both image-based and vector-based DMControl environments. The learning curves in image-based environments are shown in \Cref{fig:rad_aarl_image}, where no statistically significant difference is observed. As readers may notice, the scores on RAD paper~\citep{laskin2020rad} are higher than the RAD and \texttt{A2-winner} learning curves reported. To avoid a misleading conclusion that RAD is much stronger than \texttt{A2-winner} , we would like to emphasize some key differences between RAD and our implementationt: 1) \textit{Large Conv encoder output dim} (RAD: 47, \method/CURL: 25); 2) \textit{Larger image size} (RAD: 108, \method/CURL: 100); 3) \textit{Larger encoder feature dim} (RAD: 64, \method/CURL: 50). We use the hyper-parameters used in CURL for consistency of scores reported in our paper.
However, in vector-based environments, as shown in \Cref{fig:rad_aarl_state}, \texttt{A2-winner} greatly outperforms RAD. Due to the huge difference between images and proprioceptive features, RAD could not transfer augmentation techniques like random crop and transforms used for images to vectors. Though RAD designs noise and random scaling for proprioceptive features, \texttt{A2-winner}  shows much better performance on vector-based settings. These results show that recent progress in using data augmentation for RL is still limited to image-based RL while using auxiliary loss functions for RL is able to boost RL across environments with totally different data types of observation. Besides comparing auxiliary losses with data augmentation in DMC, we also provide experimental results in Atari~\citep{bellemare2013arcade}. As shown in \Cref{tab:atari}, \texttt{A2-winner} significantly outperforms DrQ~\citep{kostrikov2020drq}.

\subsection{Evolution on Vector-based RL}
\label{Evolution on Vector-based RL}
\begin{wrapfigure}{r}{0.20\textwidth}
    \vspace{-20pt}
    \centering
    \includegraphics[width=0.15\textwidth]{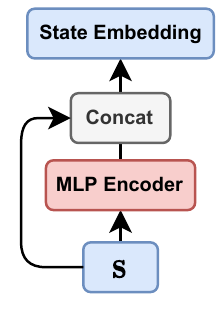}
    \captionsetup{font=footnotesize}
    \caption{Network architecture of 1-layer DenseMLP state encoder.}
    \label{fig:densemlp_1layer}
    \vspace{-10pt}
\end{wrapfigure}
\paragraph{Experiment Settings}
As for vector-based RL, we use a 1-layer densely connected MLP as the state encoder as shown in \Cref{fig:densemlp_1layer} due to the low-dimensional state space. 
So, for this setting, we focus on this simple encoder structure.
Additional ablations on state encoder architectures are given in \Cref{sec:ablation_encoder_architectures}.
In the search phase, we compare \method to SAC-Identity, SAC-DenseMLP, CURL-DenseMLP. 
To ensure a fair comparison, all SAC related hyper-parameters are the same as those reported in the CURL paper. Details can be found in \Cref{sec:hyper_evolution_state}.
SAC-Identity is vanilla SAC with no state encoder, while the other three methods (\method, SAC-DenseMLP, CURL-DenseMLP) use the same encoder architecture. Different from the image-based setting, there is no data augmentation in the vector-based setting. Note that many environments that are challenging in image-based settings become easy to tackle with vector-based inputs. Therefore we apply our search framework to more challenging environments for vector-based RL, including Cheetah-Run, Hopper-Hop and Quadruped-Run.

\begin{figure}[t]
     \centering
     \includegraphics[width=1.0\textwidth]{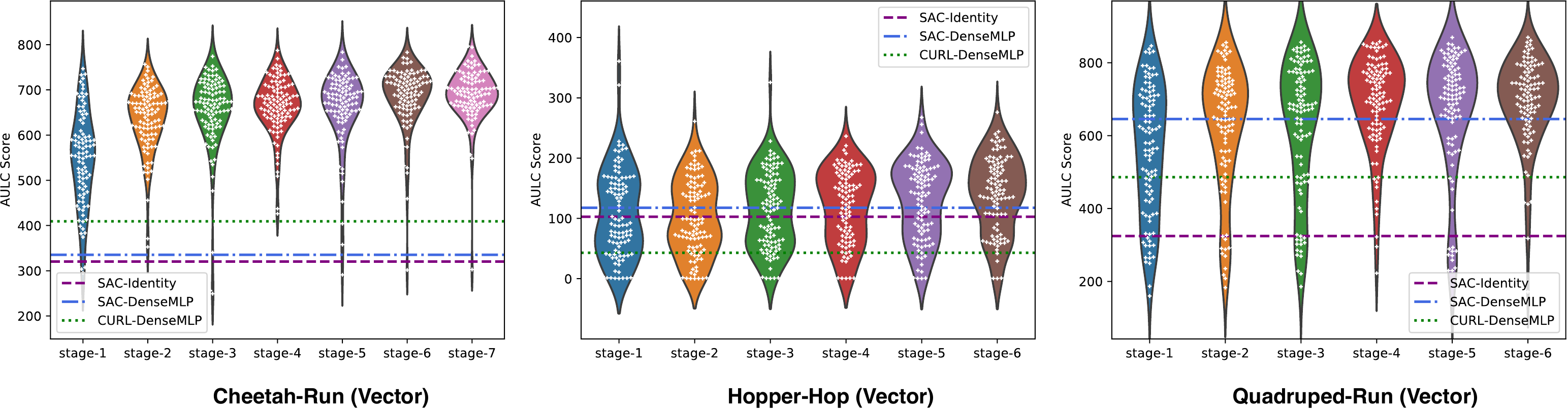}
     \caption{
     Evolution process in in the three training (vector-based) environments. Every white dot represents a loss candidate, and the score of y-axis shows its corresponding approximated AULC score. The horizontal lines show the scores of baselines. The AULC score is approximated with the average evaluation score at 300k, 600k, 900k, 1200k, 1500k time steps (Cheetah-Run at 100k, 200k, 300k, 400K).}
     \label{fig:state_evolution_process}
\end{figure}

\paragraph{Search Results} 
Similar to image-based settings, we approximate AULC with the average score agents achieved at 300k, 600k, 900k, 1200k, and 1500k time steps\footnote{\label{fn:note1}As for Cheetah-Run, we still use average score agents achieved at 100k, 200k, 300k, 400K and 500k time steps since agents converge close to optimal score within 500k time steps.}.
For each environment, we early stop the experiment when the budget of 1,500 GPU hours is exhausted.
The evolution process is shown in \Cref{fig:state_evolution_process}, where we find a large portion of candidates outperform baselines (horizontal dashed lines). The performance improvement is especially significant on Cheetah-Run, where almost all candidates in the population greatly outperform all baselines by the end of the first stage.
Similar to image-based settings, we also use cross-validation to select the best loss function, which we call ``\texttt{A2-winner-v}'' here (all the top candidates during evolution are reported in \Cref{sec:top_losses})

\subsection{Encoder Architecture Ablation for Vector-based RL}
\label{sec:ablation_encoder_architectures}
\begin{table}[h]
\caption{Normalized episodic rewards of \method (mean \& standard deviation for 5 seeds of 6 environments) on v DMControl100K with different encoder architectures.}
\begin{center}
\resizebox{0.90\textwidth}{!}{
\begin{tabular}{cccc}
\toprule
\method-MLP (1-layer) & \method-MLP (4-layer) & \method-DenseMLP (1-layer) & \method-DenseMLP (4-layer) \\
\hline
0.919 $\pm$ 0.217 & 0.544 $\pm$ 0.360 & \textbf{1.000 $\pm$ 0.129} & 0.813 $\pm$ 0.218 \\

\bottomrule
\end{tabular}
}
\label{tab:ablation_encoder_architecture}
\end{center}
\end{table}
As shown in \Cref{fig:densemlp_1layer}, we choose a 1-layer densely connected MLP as the state encoder for vector-based RL. We conduct an ablation study on different encoder architectures in the vector-based setting. The results are summarized in \Cref{tab:ablation_encoder_architecture}, where \method with 4-layer encoders consistently perform worse than 1-layer encoders. We also note that dense connection is helpful in the vector-based setting compared with naive MLP encoders.

\subsection{Visualization of Loss Landscape}
\label{sec:landscape}
\begin{figure}[htbp]
     \centering
     \includegraphics[width=1.0\textwidth]{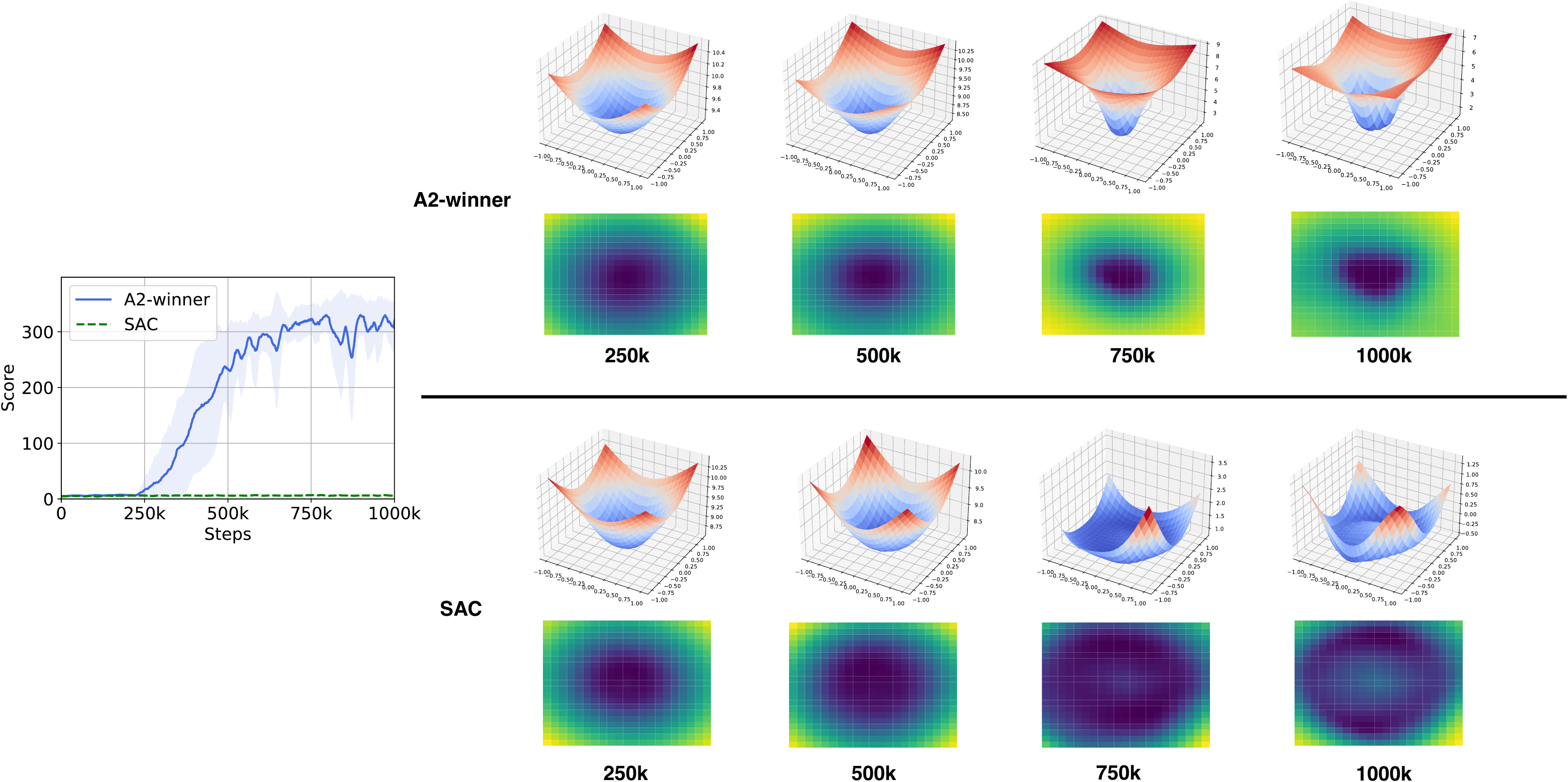}
     \caption{
     Left: Learning curves of \texttt{A2-winner} and SAC on vector-based Humanoid-Stand. Right: Critic Loss Landscape of \texttt{A2-winner} (upper right ) and SAC (lower right) at 250k, 500k, 750k and 1000k time steps, trained on vector-based Humanoid-Stand. The first row shows 3D surface plots, and the second row shows heatmap plots of loss landscapes.}
     \label{fig:loss_landscape}
\end{figure}

In an effort to reveal why auxiliary losses are helpful to RL, we draw the loss landscape of critic loss of both \texttt{A2-winner} and SAC using the technique in \cite{li2018visualizing,ota2021training}. We choose Humanoid-Stand as the testing environment since we observe the most significant advantage of \texttt{A2-winner} over SAC on complex robotics tasks like Humanoid. Note that the only difference between \texttt{A2-winner} and SAC is whether using auxiliary loss or not. As shown in \Cref{fig:loss_landscape}, the critic loss landscape of \texttt{A2-winner} appears to be convex during training, while the loss landscape of SAC becomes more non-convex as training proceeds. The auxiliary loss of \texttt{A2-winner} is able to efficiently boost Q learning (gaining near 300 reward at 500k steps), while SAC suffers from the poor results of critic learning (gaining near 0 reward even at 1000k time steps). This result shows that such an auxiliary loss might make learning easier from an optimization perspective. 

\subsection{Histogram of Auxiliary Loss Analysis}
\label{sec:histogram}
\begin{figure*}[htbp]
    \centering
    \begin{subfigure}[b]{\textwidth}
        \raisebox{-\height}{\includegraphics[width=1\textwidth]{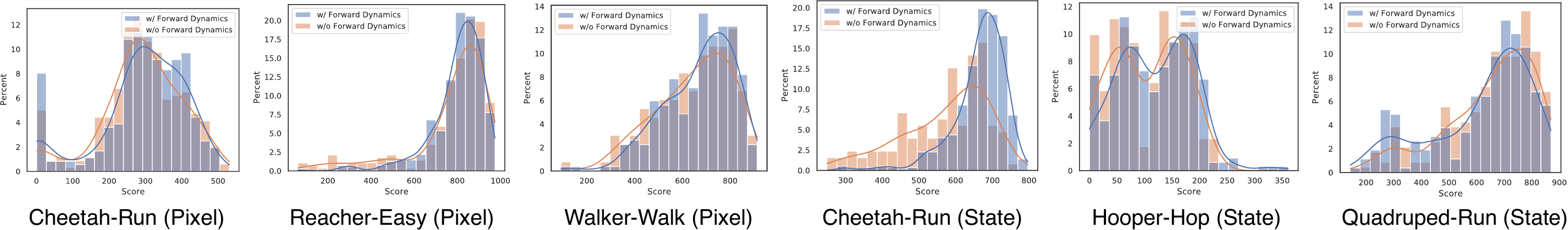}}
        \caption{Forward dynamics}
    \end{subfigure}
    \hfill
    \centering
    \begin{subfigure}[b]{\textwidth}
        \raisebox{-\height}{\includegraphics[width=1\textwidth]{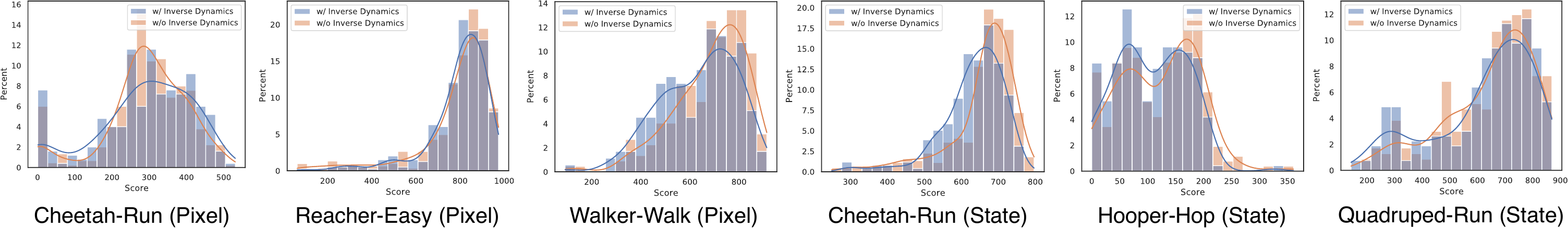}}
        \caption{Inverse dynamics}
    \end{subfigure}
    \hfill
    \centering
    \begin{subfigure}[b]{\textwidth}
        \raisebox{-\height}{\includegraphics[width=1\textwidth]{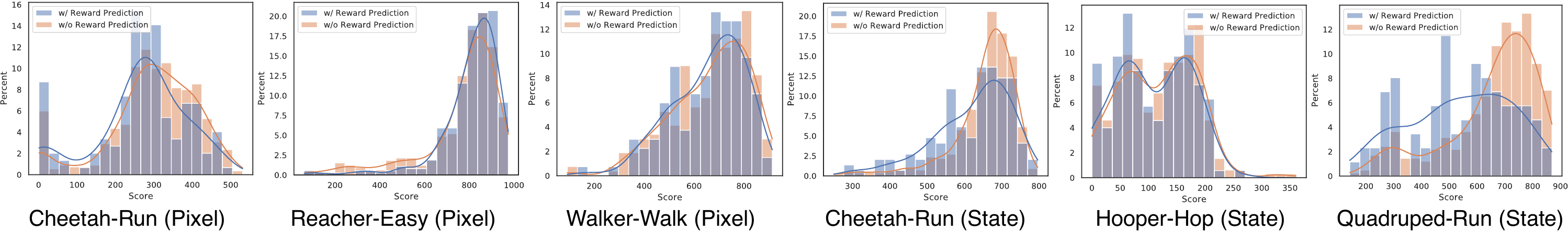}}
        \caption{Reward prediction}
    \end{subfigure}
    \hfill
    \centering
    \begin{subfigure}[b]{\textwidth}
        \raisebox{-\height}{\includegraphics[width=1\textwidth]{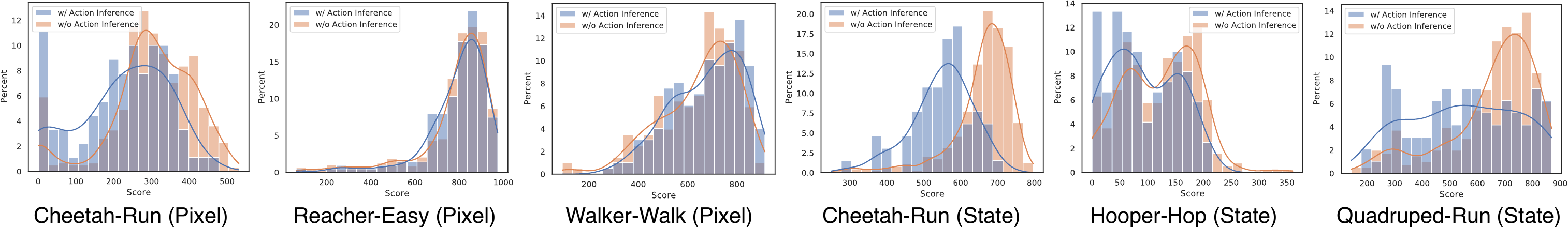}}
        \caption{Action inference}
    \end{subfigure}
    \hfill
    \centering
    \begin{subfigure}[b]{\textwidth}
        \raisebox{-\height}{\includegraphics[width=1\textwidth]{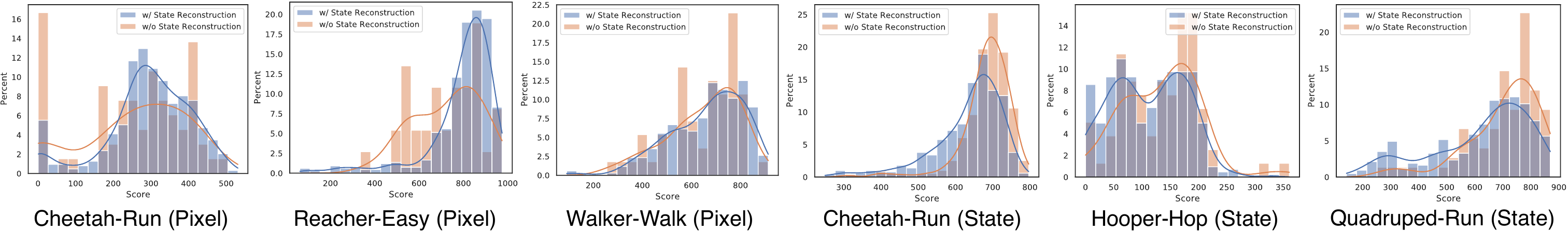}}
        \caption{State reconstruction}
    \end{subfigure}
    \hfill
    \centering
    \begin{subfigure}[b]{\textwidth}
        \raisebox{-\height}{\includegraphics[width=1\textwidth]{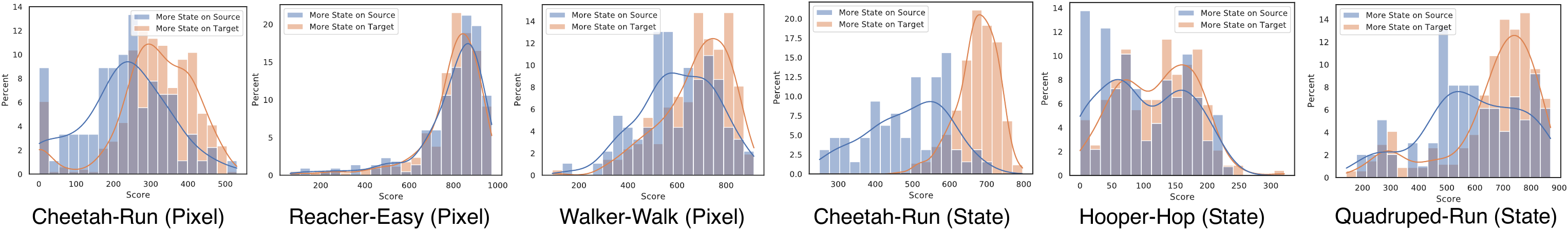}}
        \caption{State source \& target}
    \end{subfigure}
    \hfill
    \centering
    \begin{subfigure}[b]{\textwidth}
        \raisebox{-\height}{\includegraphics[width=1\textwidth]{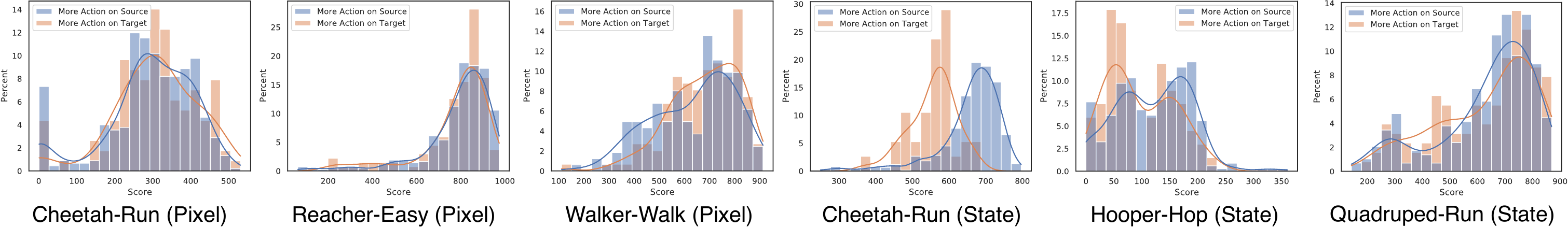}}
        \caption{Action source \& target}
    \end{subfigure}
    \hfill
    \centering
    \begin{subfigure}[b]{\textwidth}
        \raisebox{-\height}{\includegraphics[width=1\textwidth]{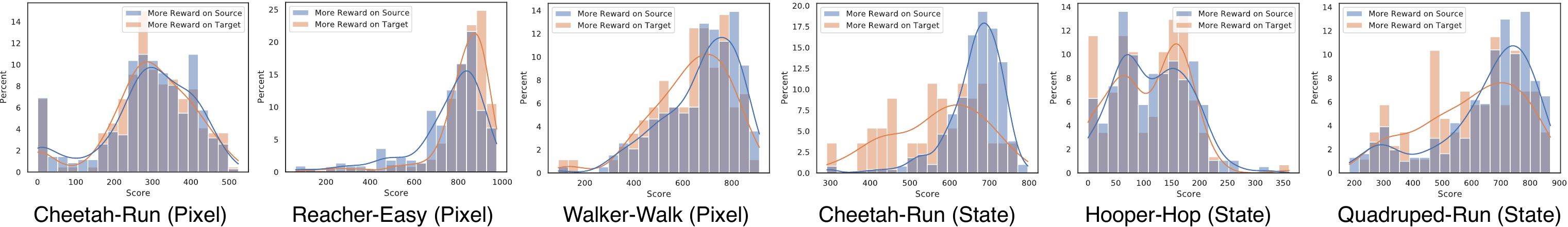}}
        \caption{Reward source \& target}
    \end{subfigure}
    \vspace{0.5cm}
    \caption{Histogram of statistical analysis of auxiliary loss candidates in six evolution processes. The x-axis represents approximated AULC score while the y-axis represents the percentage of the corresponding bin of population. Best viewed in color.}
    \label{fig:analyze_histogram}
\end{figure*}

The histogram of each pattern analysis is shown in \Cref{fig:analyze_histogram}.

\subsection{Comparing \texttt{A2-winner} with Advanced Human-designed Auxiliary Losses}
Besides CURL, many recent works (e.g., SPR~\citep{schwarzer2020spr} and ATC~\citep{stooke2021decoupling}) also proposed advanced auxiliary losses that achieve strong performance. Surprisingly, we find that both SPR and ATC designed similar patterns as we conclude in \Cref{Analysis of Auxiliary Loss Functions}, like forward dynamics and $n_{\text{target}} > n_{\text{source}}$. Particularly, in ATC, they train the encoder only with ATC loss, and we find the performance of \texttt{A2-winner} has better performances than the results reported in their paper: we are 2$\times$ more sample efficient to reach 800 scores on Cartpole-Swingup, 2$\times$ more sample efficient to reach 100 scores on Hopper-Hop, and 3$\times$ more sample efficient to reach 600 scores on Cartpole-Swingup sparse (see Figure 2 of \citep{stooke2021decoupling}). As for SPR, we find they have superior performance on Atari games benchmark, as shown in \Cref{tab:atari_appendix}, where \texttt{A2-winner} outperform all baselines except SPR. 
However, note that, our \texttt{A2-winner} is only searched on a small set of DMC benchmarks and can still generalize well to discrete-control tasks of Atari, while SPR is designed and only evaluated on Atari environments.
In addition, we believe such a gap can arise from different base RL algorithm implementation (\method is based on Efficient Rainbow DQN while SPR adopts Categorical DQN) and different hyper-parameters. 

\begin{table}[t]
\caption{Mean and Median scores (normalized by human score and random score) achieved by \method and baselines on 26 Atari games benchmarked at 100k time-steps (Atari100k).}
\label{tab:atari_appendix}
\begin{center}
\resizebox{0.99\linewidth}{!}{
\begin{tabular}{c|cccccccc|cc}
\toprule
Metric &  \texttt{A2-winner}  &  CURL  &  Eff. Rainbow & DrQ~\citep{kostrikov2020drq} & SimPLe & DER & OTRainbow & SPR & Random & Human \\
\midrule
Mean Human-Norm'd & 0.568 & 0.381 & 0.285 & 0.357 & 0.443 & 0.285 & 0.264 & \textbf{0.704} & 0.000 & 1.000\\
Median Human-Norm'd & 0.317 & 0.175 & 0.161 & 0.268 & 0.144 & 0.161 & 0.204 & \textbf{0.415} & 0.000 & 1.000 \\
\bottomrule
\end{tabular}}
\end{center}
\vspace{-14pt}
\end{table}

\subsection{The Trend of Increasing Performance during Evolution}
\label{appendix:trend}

\begin{table}[h]
\caption{Average AULC scores of populations of each stage.}
\label{tab:average_aulc_each_stage}
\begin{center}
\resizebox{0.99\linewidth}{!}{
\begin{tabular}{c|ccccccc|c}
\toprule
&  stage-1  &  stage-2 & stage-3 & stage-4 & stage-5 & stage-6 & stage-7 & SAC (baseline) \\
\midrule
Cheetah-Run & 191.75  & 252.51   & 258.09   & 284.53   & 349.52   & 351.51   & 352.57   & 285.82 \\
Reacher-Easy & 674.87  & 782.75   & 812.61 & 823.04 & 810.15 & 811.88 & 827.19 & 637.60 \\
Walker-Walk & 599.38   &  633.75   &  716.18 &  702.49  &  N/A       &  N/A       &  N/A       & 675.84 \\
\bottomrule
\end{tabular}}
\end{center}
\end{table}

\begin{table}[h]
\caption{Average AULC scores of Top-5 candidates of each stage.}
\label{tab:average_aulc_top5_each_stage}
\begin{center}
\resizebox{0.99\linewidth}{!}{
\begin{tabular}{c|ccccccc|c}
\toprule
&  stage-1  &  stage-2 & stage-3 & stage-4 & stage-5 & stage-6 & stage-7 & SAC (baseline) \\
\midrule
Cheetah-Run & 398.18  & 424.27  & 428.08   & 485.54   & 487.94   & 482.65   & 498.46  & 285.82 \\
Reacher-Easy & 931.27  & 950.61   & 943.83   & 938.91   & 954.77  & 955.02  & 969.43  & 637.60 \\
Walker-Walk & 834.09   &  883.77   &  896.52 &  880.73  &  N/A       &  N/A       &  N/A    & 675.84 \\
\bottomrule
\end{tabular}}
\end{center}
\end{table}

To illustrate the trend of increasing performance during evolution, we provide the average AULC score of populations of each stage in \Cref{tab:average_aulc_each_stage}. As for comparing evoluationary search with random sampling, we can take the \textit{stage-1 of each evolution procedure} as random sampling. As shown in \Cref{tab:average_aulc_each_stage}, the average performance of the stage-1 population (i.e., random sampling) is even worse than SAC in Cheetah-Run and Walker-Walk. Nevertheless, as evolution continues, the performance of the evolved population in the following stages improves significantly, surpassing the score of SAC. 

To illustrate the trend of increasing performance of best individuals during evolution, we provide the average AULC score of the top 5 candidates of the population at each stage in \Cref{tab:average_aulc_top5_each_stage}. As shown in \Cref{tab:average_aulc_top5_each_stage}, there is an obvious trend that the performance of the best individuals in the population at each stage continues to improve and also outperformed the baseline by a large margin during the evolution across all the three training environments.

\clearpage

\section{Search Space Complexity Analysis}
\label{sec:calculation_complexity}

The search space size is calculated by the size of \textit{input element} space multiplying by the size of the loss operator space.

For \textit{input elements}, the search space for input elements is a pair of binary masks $(m,\hat{m})$, each of which is up to length $(3k+3)$ if the length of an interaction data sequence, i.e., horizon, is limited to $k$ steps. 
In our case, we set the maximum horizon length $k_{\text{max}} = 10$. 
we calculate separately for each possible horizon length $k$. When length is $k$, the interaction sequence length $(s_t, a_t, r_t, \cdots, s_{t+k})$ has length $(3k+3)$. For binary mask $\hat{m}$, there are $2^{3k+3}$ different options. There are also $2^{3k+3}$ distinct binary mask $m$ to select targets. Therefore, there are $2^{6k+6}$ combinations when horizon length is fixed to $k$. As our maximum horizon is 10, we enumerate $k$ from $1$ to $10$, resulting in $\sum_{i=1}^{10} 2^{6i+6}$.

For \textit{operator}, we can learn intuitively from \Cref{tab:similarity_metrics} that there are 5 different similarity measures with or without negative samples, resulting in $5 \times 2 = 10$ different loss operators.

In total, the size of the entire space is

$$10 \times \sum_{i=1}^{10} 2^{6i+6}\approx 7.5 \times 10^{20}.$$

\clearpage
\section{Top-performing Auxiliary Losses}
\label{sec:top_losses}
\subsection{\texttt{A2-winner} and \texttt{A2-winner-v}}

We introduce all the top-performing auxiliary losses during evolution in this section. Note that MSE is chosen (details are given \Cref{Results of Search Space Pruning}) as the loss operator for all the auxliary losses reported below.
The source $\text{seq}_{source}$ and target $\text{seq}_{target}$ of auxiliary loss of \texttt{A2-winner} are: 
\begin{equation}
\{s_{t+1}, a_{t+1}, a_{t+2}, a_{t+3}\} \rightarrow \{r_{t}, r_{t+1}, s_{t+2}, s_{t+3}\},
\end{equation}
where \texttt{A2-winner} is the third-best candidate of stage 4 in Cheetah-Run (Image). 

The source $\text{seq}_{source}$ and target $\text{seq}_{target}$ of auxiliary loss of \texttt{A2-winner-v} are:
\begin{equation}
\begin{split}
\{s_{t}, a_{t}, a_{t+1}, s_{t+2}, a_{t+2}, a_{t+3}, r_{t+3}, a_{t+4}, r_{t+4}, a_{t+5}, a_{t+7}, s_{t+8}, a_{t+8}, r_{t+8}\} \\
\rightarrow \{s_{t+1}, s_{t+3}, a_{t+4}, s_{t+6}, s_{t+9}\},
\end{split}
\end{equation}
where \texttt{A2-winner-v} is the fourth-best candidate of stage 4 in Cheetah-Run (Vector). 

These two losses are chosen because they are the best-performing loss functions during cross-validation.

\subsection{During Evolution}
We report all the top-5 auxiliary loss candidates during evolution in this section.

\begin{table}[h]
\caption{Top-5 candidates of each stage in Cheetah-Run (Image) evolution process}
\begin{center}
\begin{subtable}[t]{\textwidth}
\resizebox{\textwidth}{!}{
\begin{threeparttable}
\begin{tabular}{c|l}
\toprule
\multicolumn{2}{c}{ Cheetah-Run (Image) }\\
\hline
Stage-1 & \makecell[l]{$\{r_{t}, s_{t+1}, a_{t+1}, r_{t+1}, a_{t+2}, r_{t+2}, a_{t+3}, r_{t+3}\} \rightarrow \{s_{t}, a_{t}, s_{t+2}, s_{t+3}, s_{t+4}\}$ \\ $\{s_{t}, a_{t}, r_{t}\} \rightarrow \{s_{t+1}\}$ \\ $\{s_{t}, a_{t}, a_{t+1}, r_{t+2}\} \rightarrow \{s_{t}, a_{t}, s_{t+1}, a_{t+1}, r_{t+1}, s_{t+2}, r_{t+2}, s_{t+3}\}$ \\ $\{s_{t}, r_{t}, a_{t+1}, a_{t+2}, a_{t+3}, r_{t+3}, r_{t+4}, a_{t+5}, r_{t+5}, s_{t+6}, s_{t+7}\} \rightarrow \{s_{t}, a_{t}, s_{t+1}, s_{t+2}, r_{t+2}, r_{t+3}, s_{t+4}, r_{t+5}, s_{t+6}, a_{t+6}, s_{t+7}\}$ \\ $\{s_{t}, a_{t}, s_{t+1}, a_{t+1}, s_{t+2}, r_{t+2}\} \rightarrow \{s_{t}, s_{t+1}, r_{t+1}, s_{t+2}, r_{t+2}, s_{t+3}\}$}\\
\hline
Stage-2 & \makecell[l]{$\{s_{t}, a_{t+1}, r_{t+2}, s_{t+4}, r_{t+4}\} \rightarrow \{s_{t+2}, a_{t+3}, r_{t+3}, a_{t+4}, s_{t+5}\}$ \\ $\{s_{t}, a_{t}, a_{t+2}, r_{t+2}\} \rightarrow \{s_{t}, r_{t}, s_{t+1}, s_{t+2}, r_{t+2}\}$ \\ $\{a_{t}, r_{t}, s_{t+1}, r_{t+1}, s_{t+2}, a_{t+2}, r_{t+2}, a_{t+3}, a_{t+4}\} \rightarrow \{s_{t+1}, s_{t+2}, s_{t+3}, a_{t+3}, s_{t+4}\}$ \\ $\{s_{t}, a_{t}, r_{t}, a_{t+1}, r_{t+1}, a_{t+2}, r_{t+2}, a_{t+3}, a_{t+4}\} \rightarrow \{s_{t}, s_{t+1}, s_{t+2}, s_{t+4}, s_{t+5}\}$ \\ $\{r_{t}, s_{t+1}, r_{t+1}\} \rightarrow \{s_{t}, a_{t}, a_{t+1}, s_{t+2}\}$}\\
\hline
Stage-3 & \makecell[l]{$\{s_{t}, a_{t}, a_{t+2}, r_{t+2}\} \rightarrow \{s_{t}, s_{t+1}, s_{t+2}, r_{t+2}\}$ \\ $\{s_{t}, r_{t}, a_{t+1}, a_{t+3}, r_{t+3}, r_{t+4}, a_{t+5}, r_{t+5}, s_{t+6}, a_{t+6}, s_{t+7}\} \rightarrow \{s_{t}, a_{t}, s_{t+1}, s_{t+2}, r_{t+2}, r_{t+3}, s_{t+4}, r_{t+5}, s_{t+6}, s_{t+7}, a_{t+7}\}$ \\ $\{s_{t}, a_{t}, a_{t+1}, r_{t+1}, a_{t+2}, r_{t+2}, a_{t+3}, s_{t+4}, a_{t+4}\} \rightarrow \{s_{t+1}, s_{t+2}, s_{t+4}, s_{t+5}\}$ \\ $\{s_{t}, a_{t}, a_{t+1}, r_{t+1}, r_{t+2}, s_{t+3}, a_{t+3}, r_{t+4}\} \rightarrow \{s_{t+1}, s_{t+2}, r_{t+3}, s_{t+4}, a_{t+4}, s_{t+5}\}$ \\ $\{r_{t}, s_{t+1}\} \rightarrow \{s_{t}, a_{t}, a_{t+1}, s_{t+2}\}$}\\
\hline
Stage-4 & \makecell[l]{$\{s_{t}, s_{t+1}, a_{t+2}, r_{t+2}, s_{t+3}, s_{t+4}\} \rightarrow \{a_{t+1}, s_{t+2}, r_{t+2}, s_{t+4}, a_{t+4}, s_{t+5}\}$ \\ $\{s_{t}, a_{t}, a_{t+1}, r_{t+1}, r_{t+2}, s_{t+3}, a_{t+3}\} \rightarrow \{s_{t+1}, s_{t+2}, r_{t+3}, s_{t+4}\}$ \\ $\{s_{t}\} \rightarrow \{s_{t}, r_{t}, s_{t+1}, r_{t+1}, s_{t+2}, a_{t+2}, r_{t+2}\}$ \\ $\{s_{t}, r_{t}, a_{t+1}, s_{t+2}, a_{t+2}, r_{t+2}, a_{t+3}, r_{t+3}, a_{t+4}\} \rightarrow \{s_{t}, a_{t}, s_{t+1}, r_{t+1}, r_{t+3}, s_{t+4}, a_{t+4}, s_{t+5}\}$ \\ $\{r_{t}, s_{t+1}, a_{t+1}\} \rightarrow \{s_{t}, a_{t}, a_{t+1}, s_{t+2}\}$}\\
\hline
Stage-5$^*$ & \makecell[l]{$\{a_{t}, r_{t}, s_{t+1}, a_{t+1}, r_{t+1}, a_{t+2}, r_{t+2}, a_{t+3}, r_{t+3}\} \rightarrow \{r_{t}, s_{t+1}, a_{t+1}, s_{t+2}, s_{t+4}\}$ \\ $\{s_{t}, a_{t+1}, r_{t+2}, a_{t+3}, s_{t+4}, r_{t+4}\} \rightarrow \{s_{t+1}, s_{t+2}, a_{t+3}, s_{t+4}, a_{t+4}, s_{t+5}\}$ \\ $^\dag\{s_{t+1}, a_{t+1}, a_{t+2}, a_{t+3}\} \rightarrow \{r_{t}, r_{t+1}, s_{t+2}, s_{t+3}\}$ \\ $\{s_{t}\} \rightarrow \{s_{t}, r_{t}, s_{t+1}, r_{t+1}, s_{t+2}\}$ \\ $\{s_{t}\} \rightarrow \{s_{t}, r_{t}, s_{t+1}, r_{t+1}, s_{t+2}, r_{t+2}\}$}\\
\hline
Stage-6 & \makecell[l]{$\{a_{t+1}, r_{t+1}, s_{t+2}, r_{t+2}, a_{t+3}, a_{t+4}\} \rightarrow \{r_{t}, s_{t+1}, r_{t+1}, r_{t+3}, s_{t+4}, a_{t+4}, s_{t+5}\}$ \\ $\{s_{t}, a_{t+1}, a_{t+3}, s_{t+4}, r_{t+4}\} \rightarrow \{s_{t+1}, s_{t+2}, a_{t+3}, s_{t+4}, a_{t+4}, s_{t+5}\}$ \\ $\{s_{t}, a_{t+1}, r_{t+1}, s_{t+2}, a_{t+2}, r_{t+2}, s_{t+3}, a_{t+3}, r_{t+3}\} \rightarrow \{a_{t}, s_{t+2}, r_{t+2}, a_{t+3}, s_{t+4}, a_{t+4}, s_{t+5}\}$ \\ $\{s_{t}, a_{t+1}, r_{t+2}, a_{t+3}, s_{t+4}, r_{t+4}\} \rightarrow \{s_{t+1}, s_{t+2}, a_{t+3}, s_{t+4}, a_{t+4}, s_{t+5}, a_{t+5}\}$ \\ $\{s_{t}, a_{t+1}, a_{t+2}, r_{t+2}, s_{t+3}, a_{t+3}, s_{t+4}, r_{t+4}\} \rightarrow \{s_{t+1}, a_{t+1}, s_{t+2}, r_{t+2}, s_{t+4}, a_{t+4}, r_{t+4}, s_{t+5}\}$}\\
\hline
Stage-7 & \makecell[l]{$\{s_{t}, a_{t}, r_{t}, a_{t+1}, r_{t+2}, a_{t+3}, s_{t+4}\} \rightarrow \{a_{t}, r_{t}, s_{t+2}, r_{t+3}, s_{t+4}, a_{t+4}, r_{t+4}, s_{t+5}\}$ \\ $\{s_{t}, r_{t+2}, a_{t+3}, s_{t+4}\} \rightarrow \{a_{t}, s_{t+1}, s_{t+2}, r_{t+2}, a_{t+3}, r_{t+3}, s_{t+4}, a_{t+4}, r_{t+4}, s_{t+5}\}$ \\ $\{s_{t+1}, a_{t+2}, r_{t+2}, s_{t+3}, a_{t+3}, r_{t+3}, s_{t+4}\} \rightarrow \{r_{t}, a_{t+1}, r_{t+1}, s_{t+2}, r_{t+2}, s_{t+4}, a_{t+4}, s_{t+5}\}$ \\ $\{s_{t}, a_{t+1}, s_{t+2}, a_{t+2}, r_{t+2}, s_{t+3}, a_{t+3}, r_{t+3}, a_{t+4}\} \rightarrow \{a_{t}, s_{t+1}, s_{t+2}, r_{t+2}, s_{t+3}, r_{t+3}, s_{t+4}, a_{t+4}, s_{t+5}\}$ \\ $\{s_{t}, a_{t+1}, r_{t+2}, a_{t+3}, s_{t+4}, r_{t+4}\} \rightarrow \{s_{t+1}, s_{t+2}, r_{t+2}, a_{t+3}, s_{t+4}, a_{t+4}, s_{t+5}, a_{t+5}\}$}\\
\bottomrule
\end{tabular}
\begin{tablenotes}
    \item[] $*$: Used for cross-validation. $\dag$: \texttt{A2-winner}. 
\end{tablenotes}
\end{threeparttable}
}
\end{subtable}

\end{center}

\label{tab:top_losses_cheetah_run_image}
\end{table}

\begin{table}[h]
\caption{Top-5 candidates of each stage in Reacher-Easy (Image) evolution process}
\begin{center}
\begin{subtable}[t]{\textwidth}
\resizebox{\textwidth}{!}{
\begin{tabular}{c|l}
\toprule
\multicolumn{2}{c}{ Reacher-Easy (Image) }\\
\hline
Stage-1 & \makecell[l]{$\{s_{t+1}, a_{t+1}\} \rightarrow \{r_{t}, r_{t+1}\}$ \\ $\{r_{t}, s_{t+1}, a_{t+1}, s_{t+2}, a_{t+2}, r_{t+2}, a_{t+3}, a_{t+4}, r_{t+4}, s_{t+5}, a_{t+5}, r_{t+5}, s_{t+6}, a_{t+6}, r_{t+6}, r_{t+7}, s_{t+8}, a_{t+8}, s_{t+9}\} \rightarrow \{a_{t+1}, r_{t+1}, s_{t+2}, s_{t+3}, r_{t+3}, s_{t+4}, a_{t+4}, a_{t+5}, a_{t+6}, r_{t+6}, s_{t+7}, a_{t+7}, r_{t+7}, a_{t+8}, r_{t+8}, s_{t+10}\}$ \\ $\{s_{t}, r_{t}, s_{t+1}, a_{t+2}, s_{t+4}, a_{t+4}, r_{t+5}, s_{t+6}, a_{t+6}, a_{t+7}, s_{t+9}, s_{t+10}\} \rightarrow \{s_{t}, s_{t+1}, r_{t+1}, s_{t+2}, r_{t+2}, r_{t+3}, s_{t+4}, a_{t+4}, r_{t+4}, s_{t+5}, s_{t+6}, a_{t+6}, r_{t+7}, r_{t+8}, s_{t+10}\}$ \\ $\{s_{t}, r_{t}, s_{t+1}, a_{t+1}, s_{t+2}, s_{t+3}, a_{t+3}, a_{t+4}, s_{t+5}, r_{t+5}, a_{t+6}, r_{t+6}, s_{t+7}\} \rightarrow \{a_{t+1}, s_{t+2}, a_{t+2}, r_{t+2}, s_{t+3}, a_{t+3}, r_{t+3}, s_{t+4}, a_{t+4}, r_{t+4}, s_{t+6}, a_{t+6}\}$ \\ $\{s_{t+1}, r_{t+1}, a_{t+2}, r_{t+2}, r_{t+3}, s_{t+4}, a_{t+4}, r_{t+4}, s_{t+5}, s_{t+6}, r_{t+6}\} \rightarrow \{r_{t}, s_{t+1}, s_{t+2}, a_{t+2}, r_{t+2}, s_{t+4}, s_{t+5}, r_{t+5}, s_{t+6}, a_{t+6}, s_{t+7}, s_{t+8}\}$}\\
\hline
Stage-2 & \makecell[l]{$\{s_{t}, s_{t+1}, a_{t+1}\} \rightarrow \{r_{t}, r_{t+1}\}$ \\ $\{s_{t}, r_{t}, a_{t+2}, a_{t+3}, s_{t+4}, a_{t+4}, r_{t+4}, a_{t+5}, r_{t+5}, s_{t+6}, a_{t+7}, a_{t+8}\} \rightarrow \{r_{t}, a_{t+1}, s_{t+2}, r_{t+2}, s_{t+3}, a_{t+3}, r_{t+3}, a_{t+4}, r_{t+4}, a_{t+5}, r_{t+5}, a_{t+6}, s_{t+7}, s_{t+9}\}$ \\ $\{s_{t}, a_{t}, r_{t}, s_{t+1}, a_{t+2}, a_{t+3}, s_{t+5}\} \rightarrow \{a_{t}, s_{t+2}, a_{t+2}, r_{t+2}, s_{t+3}, a_{t+3}, a_{t+4}, r_{t+4}\}$ \\ $\{s_{t}, a_{t}, s_{t+1}, a_{t+3}, s_{t+4}, a_{t+4}, a_{t+5}, s_{t+6}, a_{t+6}, a_{t+7}, r_{t+7}, s_{t+9}, a_{t+9}\} \rightarrow \{s_{t}, a_{t}, r_{t}, s_{t+1}, r_{t+1}, s_{t+2}, r_{t+3}, s_{t+4}, r_{t+4}, s_{t+6}, a_{t+6}, r_{t+7}, a_{t+8}, s_{t+10}\}$ \\ $\{s_{t}, a_{t}, r_{t}, a_{t+1}, a_{t+3}, r_{t+3}, s_{t+4}, a_{t+4}, s_{t+5}, a_{t+5}, r_{t+5}, a_{t+6}, r_{t+6}, a_{t+7}\} \rightarrow \{s_{t+1}, s_{t+2}, a_{t+2}, s_{t+3}, s_{t+6}, a_{t+6}, s_{t+7}, s_{t+8}\}$}\\
\hline
Stage-3 & \makecell[l]{$\{a_{t}, s_{t+1}, a_{t+1}, s_{t+2}, a_{t+2}\} \rightarrow \{r_{t}, a_{t+1}, r_{t+1}, r_{t+2}\}$ \\ $\{s_{t}, a_{t}, s_{t+1}, s_{t+2}, a_{t+2}, a_{t+3}, s_{t+4}, r_{t+4}, a_{t+5}, r_{t+5}, r_{t+6}, a_{t+7}\} \rightarrow \{s_{t}, a_{t}, r_{t}, s_{t+2}, s_{t+3}, a_{t+3}, s_{t+4}, a_{t+4}, a_{t+5}, r_{t+5}, s_{t+6}, s_{t+7}, s_{t+8}\}$ \\ $\{s_{t}, a_{t}, s_{t+1}, a_{t+1}, r_{t+1}, s_{t+2}, a_{t+2}, a_{t+3}, s_{t+4}, r_{t+6}, a_{t+7}\} \rightarrow \{r_{t}, a_{t+2}, r_{t+2}, a_{t+3}, r_{t+3}, s_{t+4}, a_{t+4}, a_{t+5}, s_{t+6}, a_{t+6}, s_{t+7}, r_{t+7}, s_{t+8}\}$ \\ $\{s_{t}, a_{t}, s_{t+1}, r_{t+1}, a_{t+2}, a_{t+3}, s_{t+4}, r_{t+4}, s_{t+6}, r_{t+6}, a_{t+7}\} \rightarrow \{s_{t}, r_{t}, r_{t+1}, a_{t+2}, r_{t+2}, a_{t+3}, r_{t+3}, s_{t+4}, a_{t+4}, a_{t+5}, r_{t+5}, s_{t+6}, s_{t+7}, s_{t+8}\}$ \\ $\{s_{t}, a_{t}, r_{t}, a_{t+1}, r_{t+1}, a_{t+2}, s_{t+6}, s_{t+7}, a_{t+7}, s_{t+8}\} \rightarrow \{s_{t}, r_{t}, s_{t+2}, a_{t+2}, r_{t+2}, a_{t+3}, r_{t+3}, s_{t+8}, a_{t+8}\}$}\\
\hline
Stage-4 & \makecell[l]{$\{s_{t}, a_{t}, r_{t}, a_{t+1}, a_{t+3}, r_{t+3}, s_{t+4}, s_{t+5}, a_{t+5}, r_{t+5}, a_{t+6}, r_{t+6}, a_{t+7}\} \rightarrow \{s_{t+1}, s_{t+2}, s_{t+3}, r_{t+4}, r_{t+5}, r_{t+7}, s_{t+8}\}$ \\ $\{a_{t}, s_{t+1}, a_{t+1}, s_{t+2}\} \rightarrow \{r_{t}, r_{t+1}, a_{t+2}, r_{t+2}\}$ \\ $\{s_{t}, a_{t}, r_{t}, a_{t+1}, r_{t+3}, s_{t+4}, s_{t+5}, s_{t+6}, r_{t+6}, a_{t+7}\} \rightarrow \{s_{t+1}, s_{t+2}, a_{t+3}, s_{t+4}, r_{t+4}, s_{t+5}, s_{t+6}, s_{t+7}, r_{t+7}, s_{t+8}\}$ \\ $\{s_{t}, a_{t}, r_{t+1}, a_{t+2}, a_{t+3}, s_{t+4}, r_{t+4}, r_{t+5}, s_{t+6}, r_{t+6}, a_{t+7}\} \rightarrow \{s_{t}, r_{t}, r_{t+1}, a_{t+2}, r_{t+2}, a_{t+3}, r_{t+3}, s_{t+4}, a_{t+4}, a_{t+5}, r_{t+5}, s_{t+6}, s_{t+7}, s_{t+8}\}$ \\ $\{a_{t}, s_{t+1}, a_{t+1}, s_{t+2}, a_{t+2}\} \rightarrow \{r_{t}, r_{t+1}, r_{t+2}\}$}\\
\hline
Stage-5 & \makecell[l]{$\{s_{t}, a_{t}, a_{t+1}, a_{t+2}, a_{t+3}, s_{t+4}, s_{t+5}, a_{t+6}, r_{t+6}, a_{t+7}, s_{t+8}\} \rightarrow \{r_{t}, a_{t+2}, r_{t+2}, a_{t+3}, s_{t+4}, s_{t+5}, a_{t+5}, s_{t+6}, a_{t+6}, s_{t+7}, r_{t+7}, s_{t+8}\}$ \\ $\{s_{t}, a_{t}, s_{t+1}, a_{t+1}, r_{t+1}\} \rightarrow \{r_{t}, s_{t+2}\}$ \\ $\{s_{t}, a_{t}, r_{t}, a_{t+1}, r_{t+3}, s_{t+4}, s_{t+5}, s_{t+6}, r_{t+6}, a_{t+7}\} \rightarrow \{s_{t+1}, s_{t+2}, a_{t+3}, s_{t+4}, r_{t+4}, s_{t+5}, r_{t+6}, s_{t+7}, r_{t+7}, s_{t+8}\}$ \\ $\{s_{t}, a_{t}, r_{t}, a_{t+1}, r_{t+1}, s_{t+2}, a_{t+2}, a_{t+3}, s_{t+4}, s_{t+5}, a_{t+5}, r_{t+6}, a_{t+7}\} \rightarrow \{s_{t}, r_{t}, a_{t+1}, r_{t+1}, a_{t+2}, r_{t+2}, s_{t+3}, a_{t+3}, s_{t+4}, s_{t+6}, a_{t+6}, r_{t+6}, s_{t+7}\}$ \\ $\{s_{t}, a_{t}, s_{t+1}, r_{t+1}, a_{t+2}, s_{t+3}, s_{t+4}, r_{t+6}, a_{t+7}, s_{t+8}\} \rightarrow \{s_{t}, r_{t}, r_{t+1}, a_{t+2}, r_{t+2}, a_{t+3}, s_{t+4}, a_{t+4}, a_{t+5}, s_{t+6}, r_{t+6}, s_{t+7}, s_{t+8}\}$}\\
\hline
Stage-6 & \makecell[l]{$\{s_{t}, a_{t}, r_{t}, a_{t+1}, a_{t+3}, r_{t+3}, s_{t+4}, a_{t+4}, s_{t+5}, a_{t+5}, r_{t+5}, s_{t+6}, a_{t+6}, r_{t+6}\} \rightarrow \{s_{t+1}, s_{t+2}, a_{t+2}, s_{t+3}, r_{t+4}, s_{t+6}\}$ \\ $\{s_{t}, a_{t}, r_{t}, s_{t+1}, a_{t+1}, r_{t+1}, s_{t+2}, a_{t+2}, s_{t+3}, a_{t+4}, r_{t+5}, r_{t+6}, s_{t+7}, s_{t+8}\} \rightarrow \{s_{t}, r_{t}, r_{t+2}, a_{t+3}, r_{t+3}, s_{t+4}, r_{t+4}, a_{t+5}, r_{t+5}, s_{t+6}, a_{t+6}, r_{t+6}, s_{t+7}, a_{t+7}, r_{t+7}, s_{t+8}\}$ \\ $\{s_{t}, a_{t}, r_{t}, r_{t+1}, a_{t+2}, a_{t+3}, s_{t+4}, s_{t+5}, r_{t+5}\} \rightarrow \{s_{t}, r_{t}, r_{t+1}, a_{t+2}, a_{t+3}, r_{t+3}, a_{t+4}, r_{t+4}, a_{t+5}, s_{t+6}, a_{t+6}, r_{t+6}, s_{t+7}, s_{t+8}\}$ \\ $\{s_{t}, a_{t}, s_{t+1}, a_{t+3}, a_{t+4}, a_{t+5}, s_{t+6}, a_{t+6}, s_{t+7}, a_{t+7}, r_{t+7}, s_{t+9}, a_{t+9}\} \rightarrow \{s_{t}, a_{t}, r_{t}, s_{t+1}, r_{t+1}, s_{t+2}, s_{t+3}, r_{t+3}, s_{t+4}, a_{t+4}, r_{t+4}, s_{t+6}, a_{t+6}, r_{t+7}, a_{t+8}, a_{t+9}, s_{t+10}\}$ \\ $\{s_{t}, a_{t}, r_{t}, a_{t+1}, r_{t+1}, a_{t+2}, a_{t+3}, s_{t+4}, s_{t+5}, a_{t+5}, r_{t+6}, a_{t+7}\} \rightarrow \{r_{t}, s_{t+1}, r_{t+1}, a_{t+2}, r_{t+2}, s_{t+3}, s_{t+4}, s_{t+6}, a_{t+6}, r_{t+6}, s_{t+7}, s_{t+8}\}$}\\
\hline
Stage-7 & \makecell[l]{$\{s_{t}, a_{t}, s_{t+1}, a_{t+1}, r_{t+1}, s_{t+2}, a_{t+2}, r_{t+5}, r_{t+6}, a_{t+7}, s_{t+8}\} \rightarrow \{r_{t}, a_{t+2}, r_{t+2}, r_{t+3}, s_{t+4}, a_{t+4}, a_{t+5}, a_{t+6}, r_{t+6}, s_{t+7}, s_{t+8}\}$ \\ $\{s_{t}, a_{t}, a_{t+1}, s_{t+2}, a_{t+2}, a_{t+3}, s_{t+4}, a_{t+5}, r_{t+5}, r_{t+6}, a_{t+7}\} \rightarrow \{s_{t+1}, r_{t+2}, a_{t+3}, r_{t+3}, s_{t+4}, a_{t+5}, r_{t+5}, s_{t+6}, a_{t+6}, s_{t+7}, r_{t+7}\}$ \\ $\{s_{t}, a_{t}, r_{t}, a_{t+2}, a_{t+3}, s_{t+4}, s_{t+5}, r_{t+5}, a_{t+6}\} \rightarrow \{s_{t}, r_{t}, r_{t+1}, s_{t+2}, a_{t+2}, a_{t+3}, r_{t+3}, r_{t+4}, s_{t+6}, s_{t+8}\}$ \\ $\{s_{t}, a_{t}, r_{t}, a_{t+1}, a_{t+3}, r_{t+3}, s_{t+4}, a_{t+4}, s_{t+5}, a_{t+5}, r_{t+5}, s_{t+6}, a_{t+6}, r_{t+6}\} \rightarrow \{r_{t}, s_{t+1}, s_{t+2}, a_{t+2}, s_{t+3}, r_{t+4}, s_{t+6}\}$ \\ $\{s_{t}, a_{t}, r_{t}, a_{t+1}, r_{t+1}, a_{t+3}, r_{t+3}, s_{t+4}, a_{t+4}, s_{t+5}, r_{t+5}, s_{t+6}, a_{t+7}, r_{t+7}\} \rightarrow \{s_{t}, r_{t}, s_{t+1}, r_{t+1}, a_{t+2}, s_{t+3}, a_{t+3}, r_{t+4}, s_{t+6}, s_{t+8}\}$}\\
\bottomrule
\end{tabular}
}
\end{subtable}

\end{center}

\label{tab:top_losses_reacher_easy_image}
\end{table}

\begin{table}[h]
\caption{Top-5 candidates of each stage in Walker-Walk (Image) evolution process}
\begin{center}
\begin{subtable}[t]{\textwidth}
\resizebox{\textwidth}{!}{
\begin{tabular}{c|l}
\toprule
\multicolumn{2}{c}{ Walker-Walk (Image) }\\
\hline
Stage-1 & \makecell[l]{$\{s_{t}, a_{t}, s_{t+2}, a_{t+2}, r_{t+2}, s_{t+3}, a_{t+4}, a_{t+5}, s_{t+6}, a_{t+6}, a_{t+7}, s_{t+8}, r_{t+8}\} \rightarrow \{s_{t}, s_{t+1}, r_{t+1}, s_{t+2}, r_{t+2}, r_{t+3}, a_{t+4}, r_{t+4}, a_{t+5}, r_{t+5}, a_{t+6}, r_{t+6}, r_{t+7}, s_{t+8}, a_{t+8}, s_{t+9}\}$ \\ $\{s_{t}, a_{t}, a_{t+1}, r_{t+1}\} \rightarrow \{a_{t}, s_{t+1}, r_{t+1}\}$ \\ $\{r_{t}, a_{t+1}, s_{t+2}, a_{t+2}, r_{t+3}, s_{t+5}, a_{t+5}, a_{t+6}, r_{t+7}, a_{t+8}\} \rightarrow \{s_{t}, a_{t}, s_{t+1}, s_{t+3}, a_{t+3}, s_{t+4}, a_{t+4}, s_{t+6}, s_{t+7}, a_{t+7}, s_{t+8}, s_{t+9}, a_{t+9}, s_{t+10}\}$ \\ $\{s_{t}, r_{t}, s_{t+1}, a_{t+1}, s_{t+2}, a_{t+2}, r_{t+2}, s_{t+3}, a_{t+4}, r_{t+4}, s_{t+5}, a_{t+5}, a_{t+6}\} \rightarrow \{s_{t}, a_{t}, r_{t}, s_{t+1}, s_{t+2}, s_{t+3}, s_{t+4}, r_{t+4}, a_{t+5}, r_{t+5}, s_{t+6}, s_{t+7}\}$ \\ $\{s_{t}, r_{t}, a_{t+1}, s_{t+2}, a_{t+2}, s_{t+3}, a_{t+3}, a_{t+4}, s_{t+5}, a_{t+5}, r_{t+5}\} \rightarrow \{s_{t}, a_{t}, r_{t}, a_{t+1}, r_{t+1}, s_{t+2}, s_{t+3}, a_{t+3}, s_{t+5}, a_{t+5}, r_{t+5}, s_{t+6}\}$}\\
\hline
Stage-2 & \makecell[l]{$\{s_{t}, r_{t}, s_{t+1}, a_{t+1}, r_{t+1}, s_{t+3}, a_{t+3}, a_{t+4}\} \rightarrow \{a_{t}, r_{t}, s_{t+1}, s_{t+2}, s_{t+3}, r_{t+3}, a_{t+4}, s_{t+5}\}$ \\ $\{s_{t}, r_{t}, s_{t+2}, a_{t+2}, r_{t+2}, s_{t+3}, r_{t+3}\} \rightarrow \{a_{t}, s_{t+1}, a_{t+1}, s_{t+2}, a_{t+2}, r_{t+2}, r_{t+3}, s_{t+4}\}$ \\ $\{r_{t}, a_{t+1}, s_{t+2}, r_{t+2}, s_{t+3}, r_{t+3}\} \rightarrow \{a_{t}, s_{t+1}, a_{t+1}, s_{t+2}, a_{t+2}, r_{t+2}, a_{t+3}, r_{t+3}, s_{t+4}\}$ \\ $\{s_{t}, r_{t}, s_{t+2}, s_{t+3}, r_{t+3}, s_{t+4}\} \rightarrow \{s_{t}, r_{t}, s_{t+1}, a_{t+2}, r_{t+2}, s_{t+3}, s_{t+4}, a_{t+4}, s_{t+5}\}$ \\ $\{s_{t}, r_{t}, s_{t+1}, s_{t+2}, s_{t+4}, a_{t+4}, a_{t+5}, s_{t+6}, r_{t+6}, s_{t+7}\} \rightarrow \{s_{t}, r_{t}, a_{t+1}, r_{t+1}, s_{t+2}, r_{t+2}, s_{t+3}, s_{t+4}, a_{t+4}, r_{t+5}, s_{t+6}, a_{t+7}, r_{t+7}, s_{t+8}\}$}\\
\hline
Stage-3 & \makecell[l]{$\{s_{t}, a_{t}, s_{t+1}, a_{t+1}, r_{t+1}, s_{t+2}, r_{t+2}, s_{t+3}, a_{t+3}, r_{t+3}\} \rightarrow \{s_{t+2}, a_{t+2}, s_{t+3}, r_{t+3}, s_{t+4}\}$ \\ $\{s_{t}, r_{t}, s_{t+1}, a_{t+1}, r_{t+1}, s_{t+2}, r_{t+2}, a_{t+3}, s_{t+4}, a_{t+4}\} \rightarrow \{a_{t}, r_{t}, s_{t+1}, s_{t+2}, a_{t+2}, s_{t+3}, r_{t+3}, a_{t+4}, s_{t+5}\}$ \\ $\{s_{t}, a_{t}, a_{t+1}, r_{t+1}\} \rightarrow \{s_{t+2}\}$ \\ $\{s_{t}, r_{t}, s_{t+1}, a_{t+1}, r_{t+1}, s_{t+3}, a_{t+3}, a_{t+4}, r_{t+4}\} \rightarrow \{s_{t}, a_{t}, r_{t}, s_{t+1}, s_{t+2}, a_{t+2}, s_{t+3}, r_{t+3}, a_{t+4}, s_{t+5}\}$ \\ $\{s_{t}, r_{t}, s_{t+2}, a_{t+2}, r_{t+2}, s_{t+3}, r_{t+3}\} \rightarrow \{a_{t}, s_{t+1}, a_{t+1}, s_{t+2}, a_{t+2}, r_{t+2}, r_{t+3}, s_{t+4}\}$}\\
\hline
Stage-4 & \makecell[l]{$\{s_{t}, a_{t}, a_{t+1}\} \rightarrow \{s_{t+1}, a_{t+1}, s_{t+2}\}$ \\ $\{s_{t}, r_{t}, r_{t+1}, s_{t+2}, s_{t+3}, r_{t+3}, r_{t+4}\} \rightarrow \{s_{t}, a_{t}, r_{t}, s_{t+1}, r_{t+1}, s_{t+2}, a_{t+2}, s_{t+3}, r_{t+3}, s_{t+4}, a_{t+4}, s_{t+5}\}$ \\ $\{s_{t}, s_{t+2}, s_{t+3}, a_{t+3}, r_{t+3}, s_{t+4}, a_{t+4}\} \rightarrow \{s_{t}, a_{t}, r_{t}, a_{t+2}, r_{t+2}, a_{t+4}, s_{t+5}\}$ \\ $\{s_{t}, r_{t}, s_{t+1}, a_{t+1}, r_{t+1}, s_{t+2}, a_{t+3}, s_{t+4}, a_{t+4}\} \rightarrow \{a_{t}, r_{t}, s_{t+1}, s_{t+2}, a_{t+2}, r_{t+3}, s_{t+5}\}$ \\ $\{s_{t}, r_{t}, s_{t+1}, a_{t+1}, r_{t+1}, r_{t+2}, s_{t+3}, a_{t+3}, s_{t+4}, a_{t+4}\} \rightarrow \{a_{t}, r_{t}, s_{t+1}, s_{t+2}, a_{t+2}, s_{t+3}, r_{t+3}, a_{t+4}, s_{t+5}\}$}\\
\bottomrule
\end{tabular}
}
\end{subtable}

\end{center}

\label{tab:top_losses_walker_walk_image}
\end{table}

\begin{table}[h]
\caption{Top-5 candidates of each stage in Cheetah-Run (Vector) evolution process}
\begin{center}
\begin{subtable}[t]{\textwidth}
\resizebox{\textwidth}{!}{
\begin{threeparttable}
\begin{tabular}{c|l}
\toprule
\multicolumn{2}{c}{ Cheetah-Run (Raw) }\\
\hline
Stage-1 & \makecell[l]{$\{s_{t}, a_{t}, r_{t}, a_{t+1}, r_{t+1}\} \rightarrow \{s_{t+1}, s_{t+2}\}$ \\ $\{a_{t}, r_{t}, s_{t+2}, a_{t+2}, a_{t+3}, r_{t+3}\} \rightarrow \{s_{t}, s_{t+1}, a_{t+1}, s_{t+3}, s_{t+4}\}$ \\ $\{a_{t}, a_{t+1}, s_{t+2}, a_{t+2}, r_{t+2}, a_{t+3}, r_{t+3}, s_{t+4}, r_{t+4}, a_{t+5}, r_{t+5}, a_{t+7}, r_{t+7}, s_{t+8}, a_{t+8}, r_{t+8}\} \rightarrow \{s_{t}, s_{t+1}, s_{t+3}, a_{t+4}, s_{t+5}, s_{t+6}, a_{t+6}, s_{t+7}, s_{t+9}\}$ \\ $\{a_{t+1}, a_{t+2}, s_{t+3}, a_{t+3}, a_{t+4}, a_{t+5}, r_{t+5}, a_{t+6}, r_{t+7}\} \rightarrow \{s_{t}, a_{t}, s_{t+1}, s_{t+2}, s_{t+4}, s_{t+5}, s_{t+6}, s_{t+7}, a_{t+7}, s_{t+8}\}$ \\ $\{s_{t}, a_{t}, a_{t+1}, a_{t+2}, r_{t+3}, a_{t+4}, r_{t+4}, s_{t+5}, a_{t+5}, a_{t+6}, s_{t+7}, a_{t+7}, s_{t+8}, a_{t+8}, r_{t+8}\} \rightarrow \{s_{t+1}, s_{t+2}, s_{t+3}, a_{t+3}, s_{t+4}, s_{t+6}, s_{t+9}\}$}\\
\hline
Stage-2 & \makecell[l]{$\{s_{t}, a_{t}, r_{t}, a_{t+1}, r_{t+1}\} \rightarrow \{s_{t+1}, s_{t+2}\}$ \\ $\{s_{t}, a_{t}, r_{t}, a_{t+1}, r_{t+1}\} \rightarrow \{s_{t+1}, s_{t+2}\}$ \\ $\{s_{t}, a_{t}, a_{t+1}, r_{t+1}, a_{t+2}, r_{t+2}, a_{t+3}, r_{t+3}, a_{t+4}, r_{t+4}, s_{t+5}, a_{t+5}, r_{t+5}, a_{t+6}, a_{t+7}, a_{t+8}, r_{t+8}\} \rightarrow \{a_{t}, s_{t+1}, s_{t+2}, a_{t+2}, s_{t+3}, s_{t+4}, s_{t+6}, s_{t+9}\}$ \\ $\{s_{t}, a_{t}, a_{t+1}, s_{t+2}, a_{t+2}, a_{t+3}, r_{t+3}, a_{t+4}, r_{t+4}, a_{t+5}, a_{t+7}, s_{t+8}, a_{t+8}, r_{t+8}\} \rightarrow \{s_{t+1}, s_{t+3}, a_{t+4}, s_{t+6}, s_{t+9}\}$ \\ $\{s_{t}, a_{t}, r_{t}\} \rightarrow \{r_{t}, s_{t+1}\}$}\\
\hline
Stage-3 & \makecell[l]{$\{s_{t}, a_{t}, r_{t}, a_{t+1}, r_{t+1}\} \rightarrow \{s_{t+1}\}$ \\ $\{s_{t}, a_{t}\} \rightarrow \{r_{t}, s_{t+1}\}$ \\ $\{s_{t}, a_{t}, r_{t}, a_{t+1}\} \rightarrow \{s_{t+2}\}$ \\ $\{s_{t}, a_{t}, r_{t}\} \rightarrow \{s_{t+1}\}$ \\ $\{s_{t}, a_{t}, a_{t+1}, r_{t+1}\} \rightarrow \{s_{t}, s_{t+1}, s_{t+2}\}$}\\
\hline
Stage-4$^*$ & \makecell[l]{$\{s_{t}, a_{t}, r_{t}, a_{t+1}\} \rightarrow \{s_{t+1}, s_{t+2}\}$ \\ $\{s_{t}, a_{t}, a_{t+1}, r_{t+1}, a_{t+2}, r_{t+2}, r_{t+3}, r_{t+4}, s_{t+5}, a_{t+5}, r_{t+5}, a_{t+6}, a_{t+7}, a_{t+8}, r_{t+8}\} \rightarrow \{a_{t}, s_{t+1}, s_{t+2}, a_{t+2}, s_{t+3}, s_{t+4}, s_{t+6}, s_{t+9}\}$ \\ $\{s_{t}, a_{t}, a_{t+1}, a_{t+2}, r_{t+2}, r_{t+3}, a_{t+4}, r_{t+4}, a_{t+5}, r_{t+5}, a_{t+6}, a_{t+7}, a_{t+8}, r_{t+8}\} \rightarrow \{a_{t}, s_{t+1}, s_{t+2}, a_{t+2}, s_{t+3}, s_{t+4}, s_{t+6}, a_{t+8}, s_{t+9}\}$ \\ $^\dag\{s_{t}, a_{t}, a_{t+1}, s_{t+2}, a_{t+2}, a_{t+3}, r_{t+3}, a_{t+4}, r_{t+4}, a_{t+5}, a_{t+7}, s_{t+8}, a_{t+8}, r_{t+8}\} \rightarrow \{s_{t+1}, s_{t+3}, a_{t+4}, s_{t+6}, s_{t+9}\}$ \\ $\{s_{t}, a_{t}, a_{t+1}, a_{t+2}, r_{t+3}, a_{t+4}, r_{t+4}, s_{t+5}, a_{t+5}, s_{t+7}, a_{t+7}, s_{t+8}, a_{t+8}, r_{t+8}\} \rightarrow \{s_{t+1}, s_{t+2}, s_{t+3}, a_{t+3}, s_{t+4}, s_{t+6}, a_{t+8}, r_{t+8}, s_{t+9}\}$}\\
\hline
Stage-5$^*$ & \makecell[l]{$\{s_{t}, a_{t}, r_{t}, a_{t+1}\} \rightarrow \{s_{t+1}, r_{t+1}\}$ \\ $\{s_{t}, a_{t}, a_{t+1}, r_{t+1}\} \rightarrow \{s_{t+1}, a_{t+1}, s_{t+2}, a_{t+2}, r_{t+2}, s_{t+3}\}$ \\ $\{s_{t}, a_{t}, r_{t}, a_{t+1}, r_{t+1}\} \rightarrow \{s_{t+1}\}$ \\ $\{s_{t}, a_{t}, a_{t+1}, a_{t+2}, r_{t+2}, r_{t+3}, a_{t+4}, r_{t+4}, s_{t+5}, a_{t+5}, r_{t+5}, a_{t+6}, a_{t+7}, a_{t+8}, r_{t+8}\} \rightarrow \{s_{t+1}, s_{t+2}, s_{t+3}, s_{t+4}, s_{t+6}, a_{t+8}, s_{t+9}\}$ \\ $\{s_{t}, a_{t}, r_{t}, a_{t+1}\} \rightarrow \{s_{t+1}, s_{t+2}\}$}\\
\hline
Stage-6 & \makecell[l]{$\{s_{t}, a_{t}, a_{t+1}, a_{t+2}, r_{t+2}, a_{t+3}, r_{t+3}, r_{t+4}, s_{t+5}, a_{t+5}, r_{t+5}, a_{t+6}, a_{t+7}, a_{t+8}, r_{t+8}\} \rightarrow \{s_{t}, a_{t}, s_{t+1}, s_{t+2}, a_{t+2}, s_{t+3}, a_{t+3}, a_{t+4}, s_{t+5}, s_{t+6}, a_{t+8}, s_{t+9}\}$ \\ $\{s_{t}, a_{t}, a_{t+1}\} \rightarrow \{r_{t}, s_{t+1}, r_{t+1}, s_{t+2}\}$ \\ $\{s_{t}, a_{t}, a_{t+1}, a_{t+2}, r_{t+2}, a_{t+3}, r_{t+3}, a_{t+4}, r_{t+4}, s_{t+5}, a_{t+5}, a_{t+6}, a_{t+7}, s_{t+8}, a_{t+8}, r_{t+8}\} \rightarrow \{a_{t}, s_{t+1}, s_{t+2}, s_{t+3}, s_{t+6}, a_{t+8}, r_{t+8}, s_{t+9}\}$ \\ $\{s_{t}, a_{t}, r_{t}, a_{t+1}\} \rightarrow \{s_{t+1}, s_{t+2}\}$ \\ $\{s_{t}, a_{t}, r_{t}, a_{t+1}\} \rightarrow \{s_{t+1}, s_{t+2}\}$}\\
\hline
Stage-7 & \makecell[l]{$\{s_{t}, a_{t}, r_{t}, a_{t+1}, r_{t+1}\} \rightarrow \{s_{t+1}, s_{t+2}\}$ \\ $\{s_{t}, a_{t}, r_{t}, a_{t+1}, r_{t+1}\} \rightarrow \{s_{t+1}, r_{t+1}, s_{t+2}\}$ \\ $\{s_{t}, a_{t}, r_{t}, a_{t+1}\} \rightarrow \{r_{t}, s_{t+1}, s_{t+2}\}$ \\ $\{s_{t}, a_{t}, a_{t+1}\} \rightarrow \{s_{t+1}, a_{t+1}, s_{t+2}\}$ \\ $\{s_{t}, a_{t}, a_{t+1}, a_{t+2}, r_{t+2}, a_{t+3}, r_{t+3}, a_{t+4}, r_{t+4}, s_{t+5}, a_{t+5}, a_{t+6}, a_{t+7}, a_{t+8}, r_{t+8}\} \rightarrow \{s_{t+1}, s_{t+2}, s_{t+3}, a_{t+3}, a_{t+4}, s_{t+6}, s_{t+8}, s_{t+9}\}$}\\
\bottomrule
\end{tabular}
\begin{tablenotes}
    \item[] $*$: Used for cross-validation. $\dag$: \texttt{A2-winner-v}. 
\end{tablenotes}
\end{threeparttable}
}
\end{subtable}

\end{center}

\label{tab:top_losses_cheetah_run_state}
\end{table}

\begin{table}[h]
\caption{Top-5 candidates of each stage in Hopper-Hop (Vector) evolution process}
\begin{center}
\begin{subtable}[t]{\textwidth}
\resizebox{\textwidth}{!}{
\begin{tabular}{c|l}
\toprule
\multicolumn{2}{c}{ Hopper-Hop (Raw) }\\
\hline
Stage-1 & \makecell[l]{$\{s_{t}, a_{t}\} \rightarrow \{r_{t}, s_{t+1}\}$ \\ $\{a_{t}, r_{t}, s_{t+2}, a_{t+2}, r_{t+2}, s_{t+3}, a_{t+3}, r_{t+3}, s_{t+5}, a_{t+5}, r_{t+5}, a_{t+6}, a_{t+7}, r_{t+7}, a_{t+8}, r_{t+8}\} \rightarrow \{s_{t}, s_{t+1}, a_{t+1}, s_{t+4}, a_{t+4}, s_{t+6}, s_{t+7}, s_{t+8}, s_{t+9}\}$ \\ $\{s_{t}, a_{t}, s_{t+2}, a_{t+3}, r_{t+4}, a_{t+5}, r_{t+5}, s_{t+6}, a_{t+6}, r_{t+7}, r_{t+8}\} \rightarrow \{s_{t}, r_{t}, s_{t+1}, a_{t+1}, r_{t+1}, s_{t+2}, s_{t+3}, s_{t+4}, a_{t+6}, s_{t+7}, a_{t+7}, r_{t+7}, a_{t+8}, s_{t+9}\}$ \\ $\{s_{t}, a_{t}, s_{t+2}, a_{t+3}, r_{t+3}, a_{t+5}\} \rightarrow \{s_{t}, a_{t+1}, s_{t+2}, s_{t+3}, r_{t+3}, s_{t+4}, r_{t+4}, r_{t+5}, s_{t+6}\}$ \\ $\{s_{t}, r_{t}\} \rightarrow \{s_{t}, r_{t}, s_{t+1}\}$}\\
\hline
Stage-2 & \makecell[l]{$\{s_{t}, a_{t}, s_{t+1}, a_{t+1}\} \rightarrow \{s_{t+1}, s_{t+2}\}$ \\ $\{s_{t}, a_{t}, r_{t}, s_{t+2}, r_{t+2}\} \rightarrow \{r_{t+1}, s_{t+2}, a_{t+2}, s_{t+3}\}$ \\ $\{s_{t}, a_{t}, s_{t+1}, a_{t+1}, a_{t+4}, s_{t+5}, a_{t+5}, s_{t+6}, a_{t+6}\} \rightarrow \{s_{t+2}, r_{t+2}, r_{t+3}, r_{t+4}, s_{t+5}, r_{t+5}, s_{t+6}, a_{t+6}, r_{t+6}\}$ \\ $\{r_{t}, a_{t+1}, r_{t+1}, s_{t+2}, a_{t+2}, s_{t+3}, a_{t+3}, a_{t+4}, r_{t+4}\} \rightarrow \{s_{t}, a_{t}, r_{t}, s_{t+1}, s_{t+2}, s_{t+3}, s_{t+4}, r_{t+4}\}$ \\ $\{s_{t}, a_{t+1}, s_{t+2}, a_{t+2}, r_{t+2}\} \rightarrow \{s_{t+1}, r_{t+1}, a_{t+2}, s_{t+3}\}$}\\
\hline
Stage-3 & \makecell[l]{$\{s_{t}, a_{t}, s_{t+1}, a_{t+1}\} \rightarrow \{s_{t}, s_{t+2}\}$ \\ $\{s_{t}\} \rightarrow \{s_{t}, a_{t}, r_{t}, s_{t+1}\}$ \\ $\{s_{t}, a_{t}, r_{t}, s_{t+2}, a_{t+2}, r_{t+2}\} \rightarrow \{s_{t}, s_{t+1}, r_{t+1}, s_{t+2}, a_{t+2}, r_{t+2}, s_{t+3}\}$ \\ $\{s_{t}, a_{t}, a_{t+1}\} \rightarrow \{s_{t+1}, s_{t+2}\}$ \\ $\{s_{t}, a_{t}, a_{t+1}\} \rightarrow \{s_{t}, s_{t+1}, s_{t+2}\}$}\\
\hline
Stage-4 & \makecell[l]{$\{s_{t}, r_{t}, s_{t+2}, a_{t+2}, r_{t+2}\} \rightarrow \{r_{t+1}, a_{t+2}, s_{t+3}\}$ \\ $\{s_{t}, a_{t}, s_{t+1}, a_{t+1}\} \rightarrow \{s_{t+2}\}$ \\ $\{s_{t}, r_{t}, s_{t+1}, a_{t+1}, r_{t+1}, s_{t+2}, a_{t+2}, r_{t+2}\} \rightarrow \{s_{t}, r_{t+1}, a_{t+2}, s_{t+3}\}$ \\ $\{s_{t}, a_{t+1}, s_{t+2}, a_{t+2}, r_{t+2}\} \rightarrow \{s_{t+1}, r_{t+1}, a_{t+2}, s_{t+3}\}$ \\ $\{s_{t}, a_{t}, s_{t+1}, a_{t+1}\} \rightarrow \{s_{t+1}, s_{t+2}\}$}\\
\hline
Stage-5 & \makecell[l]{$\{s_{t}, a_{t}, r_{t}, a_{t+1}, s_{t+2}, a_{t+2}, r_{t+2}\} \rightarrow \{s_{t}, r_{t+1}, a_{t+2}, s_{t+3}\}$ \\ $\{s_{t}, a_{t+1}, r_{t+1}, s_{t+2}, a_{t+2}, r_{t+2}\} \rightarrow \{s_{t}, r_{t+1}, s_{t+2}, a_{t+2}, s_{t+3}\}$ \\ $\{s_{t}, a_{t}, r_{t+1}\} \rightarrow \{s_{t}, s_{t+1}, s_{t+2}\}$ \\ $\{s_{t}, r_{t}, a_{t+1}, s_{t+2}, r_{t+2}\} \rightarrow \{s_{t+1}, a_{t+2}, s_{t+3}\}$ \\ $\{s_{t}, a_{t}, s_{t+1}, a_{t+1}\} \rightarrow \{r_{t+1}, s_{t+2}\}$}\\
\hline
Stage-6 & \makecell[l]{$\{s_{t}, a_{t}, a_{t+1}\} \rightarrow \{s_{t}, s_{t+1}, s_{t+2}\}$ \\ $\{s_{t}, r_{t}, s_{t+2}, a_{t+2}, r_{t+2}\} \rightarrow \{s_{t}, s_{t+1}, s_{t+2}, s_{t+3}\}$ \\ $\{s_{t}, a_{t}, a_{t+1}\} \rightarrow \{s_{t}, s_{t+1}, s_{t+2}\}$ \\ $\{s_{t}, a_{t}, r_{t}, s_{t+1}, a_{t+1}, r_{t+1}\} \rightarrow \{r_{t}, s_{t+1}, s_{t+2}\}$ \\ $\{s_{t}, a_{t}, r_{t}, a_{t+1}, s_{t+2}\} \rightarrow \{s_{t}, s_{t+1}, r_{t+1}\}$}\\
\bottomrule
\end{tabular}
}
\end{subtable}

\end{center}

\label{tab:top_losses_hopper_hop_state}
\end{table}

\begin{table}[h]
\caption{Top-5 candidates of each stage in Quadruped-Run (Vector) evolution process}
\begin{center}
\begin{subtable}[t]{\textwidth}
\resizebox{\textwidth}{!}{
\begin{tabular}{c|l}
\toprule
\multicolumn{2}{c}{ Quadruped-Run (Raw) }\\
\hline
Stage-1 & \makecell[l]{$\{a_{t}, r_{t}, s_{t+1}, s_{t+2}, a_{t+2}\} \rightarrow \{s_{t}, a_{t+1}, s_{t+3}\}$ \\ $\{r_{t}, s_{t+1}, s_{t+3}, r_{t+3}\} \rightarrow \{s_{t}, a_{t}, r_{t}, s_{t+1}, a_{t+1}, r_{t+1}, s_{t+2}, r_{t+2}, s_{t+3}, a_{t+3}, s_{t+4}\}$ \\ $\{a_{t}, a_{t+1}, r_{t+1}, s_{t+2}, r_{t+2}, s_{t+3}, a_{t+3}, r_{t+3}\} \rightarrow \{s_{t}, s_{t+1}, a_{t+2}, s_{t+4}\}$ \\ $\{s_{t}, a_{t}, r_{t+1}, a_{t+2}, s_{t+3}, a_{t+3}, r_{t+3}, s_{t+4}, s_{t+5}\} \rightarrow \{a_{t}, a_{t+1}, r_{t+1}, a_{t+2}, r_{t+3}, a_{t+4}\}$ \\ $\{s_{t}, a_{t}, r_{t}, s_{t+1}, a_{t+1}, s_{t+3}\} \rightarrow \{r_{t+1}, r_{t+2}\}$}\\
\hline
Stage-2 & \makecell[l]{$\{a_{t}, r_{t}, a_{t+2}, r_{t+2}, s_{t+3}, a_{t+3}, r_{t+3}, a_{t+4}\} \rightarrow \{s_{t}, s_{t+1}, a_{t+1}, s_{t+2}, s_{t+4}, s_{t+5}\}$ \\ $\{s_{t}, a_{t}, a_{t+1}, r_{t+1}, r_{t+2}, a_{t+3}, a_{t+4}, r_{t+4}, a_{t+5}, r_{t+5}, s_{t+6}, r_{t+6}, a_{t+7}, a_{t+8}, r_{t+8}, s_{t+9}\} \rightarrow \{s_{t+1}, s_{t+2}, a_{t+2}, s_{t+3}, s_{t+4}, s_{t+5}, s_{t+7}, s_{t+8}\}$ \\ $\{a_{t+1}, r_{t+1}, s_{t+2}, a_{t+3}, r_{t+3}\} \rightarrow \{s_{t}, a_{t}, r_{t}, a_{t+1}, a_{t+3}, s_{t+4}\}$ \\ $\{a_{t}, a_{t+1}, s_{t+2}, a_{t+2}, a_{t+3}, a_{t+4}, a_{t+5}, r_{t+5}, a_{t+6}, r_{t+6}, a_{t+7}, s_{t+8}, a_{t+8}\} \rightarrow \{s_{t}, s_{t+1}, s_{t+3}, s_{t+4}, s_{t+5}, s_{t+6}, s_{t+7}, s_{t+8}, s_{t+9}\}$ \\ $\{s_{t}, s_{t+1}, a_{t+1}, s_{t+2}, a_{t+2}, s_{t+3}, s_{t+4}\} \rightarrow \{s_{t}, a_{t}, s_{t+1}, s_{t+2}, a_{t+2}\}$}\\
\hline
Stage-3 & \makecell[l]{$\{a_{t}, a_{t+1}, a_{t+3}, r_{t+3}, r_{t+4}, a_{t+5}, a_{t+7}, r_{t+7}, s_{t+8}, a_{t+8}\} \rightarrow \{s_{t}, s_{t+1}, s_{t+2}, s_{t+3}, s_{t+4}, a_{t+4}, s_{t+5}, a_{t+5}, r_{t+5}, s_{t+6}, a_{t+6}, r_{t+6}, s_{t+7}, s_{t+9}\}$ \\ $\{a_{t}, a_{t+1}, a_{t+3}, r_{t+3}, a_{t+5}, a_{t+7}, s_{t+8}, a_{t+8}\} \rightarrow \{s_{t}, a_{t}, s_{t+1}, a_{t+2}, s_{t+3}, s_{t+4}, a_{t+4}, s_{t+5}, s_{t+6}, a_{t+6}, s_{t+7}, s_{t+9}\}$ \\ $\{a_{t}, r_{t}, r_{t+2}, s_{t+3}, a_{t+3}, r_{t+3}, r_{t+4}\} \rightarrow \{s_{t}, s_{t+1}, a_{t+1}, s_{t+2}, a_{t+2}, a_{t+3}, s_{t+4}, s_{t+5}\}$ \\ $\{a_{t}, a_{t+1}, r_{t+3}, a_{t+4}, r_{t+4}, a_{t+5}, a_{t+7}, r_{t+7}, s_{t+8}\} \rightarrow \{s_{t}, r_{t}, s_{t+1}, s_{t+3}, s_{t+4}, s_{t+5}, s_{t+6}, a_{t+6}, s_{t+7}, s_{t+8}, s_{t+9}\}$ \\ $\{s_{t}, a_{t}, r_{t}, r_{t+1}, a_{t+2}, r_{t+2}, s_{t+3}, a_{t+3}, r_{t+3}, s_{t+4}, s_{t+5}\} \rightarrow \{a_{t}, a_{t+1}, r_{t+1}, a_{t+2}, r_{t+3}\}$}\\
\hline
Stage-4 & \makecell[l]{$\{r_{t}, r_{t+1}, a_{t+2}, r_{t+2}, s_{t+3}, a_{t+3}, r_{t+3}, s_{t+4}, s_{t+5}\} \rightarrow \{a_{t+2}, r_{t+3}, a_{t+4}\}$ \\ $\{s_{t}, a_{t}, r_{t+1}, a_{t+2}, r_{t+2}, s_{t+3}, a_{t+3}, r_{t+3}, s_{t+4}, s_{t+5}\} \rightarrow \{a_{t}, a_{t+1}, a_{t+2}, r_{t+3}, a_{t+4}\}$ \\ $\{a_{t}, a_{t+1}, a_{t+3}, r_{t+3}, s_{t+4}\} \rightarrow \{s_{t+1}, r_{t+1}, s_{t+2}, a_{t+2}, r_{t+2}, s_{t+3}\}$ \\ $\{s_{t}, a_{t}, r_{t}, r_{t+1}, a_{t+2}, s_{t+3}, a_{t+3}, r_{t+3}, s_{t+4}, s_{t+5}\} \rightarrow \{a_{t}, a_{t+2}, r_{t+3}, a_{t+4}\}$ \\ $\{s_{t+2}, a_{t+2}, a_{t+3}\} \rightarrow \{s_{t}, a_{t}, a_{t+2}, s_{t+3}, a_{t+3}, s_{t+4}\}$}\\
\hline
Stage-5 & \makecell[l]{$\{a_{t+1}, r_{t+1}, s_{t+2}, a_{t+2}, r_{t+2}, a_{t+3}, r_{t+3}\} \rightarrow \{s_{t}, a_{t}, r_{t}, a_{t+1}, r_{t+1}, a_{t+2}, s_{t+3}, a_{t+3}\}$ \\ $\{a_{t}, r_{t}, r_{t+1}, a_{t+2}, r_{t+2}, s_{t+3}, a_{t+3}, r_{t+3}, r_{t+4}\} \rightarrow \{s_{t}, s_{t+1}, a_{t+1}, s_{t+2}, a_{t+2}, a_{t+3}, s_{t+4}, s_{t+5}\}$ \\ $\{a_{t}, a_{t+1}, a_{t+3}, r_{t+3}, a_{t+4}, r_{t+4}, a_{t+5}, a_{t+7}, r_{t+7}, s_{t+8}, a_{t+8}\} \rightarrow \{s_{t}, r_{t}, s_{t+1}, s_{t+2}, s_{t+3}, s_{t+4}, s_{t+5}, a_{t+5}, s_{t+6}, a_{t+6}, r_{t+6}, s_{t+7}, s_{t+8}\}$ \\ $\{a_{t}, a_{t+1}, s_{t+2}, a_{t+3}, r_{t+3}, a_{t+4}, a_{t+5}, s_{t+6}, a_{t+7}, s_{t+8}, a_{t+8}\} \rightarrow \{s_{t}, s_{t+1}, s_{t+2}, a_{t+2}, s_{t+3}, s_{t+4}, a_{t+4}, s_{t+5}, s_{t+6}, s_{t+7}, r_{t+7}, r_{t+8}\}$ \\ $\{s_{t}, a_{t}, r_{t}, r_{t+1}, a_{t+2}, s_{t+3}, a_{t+3}, r_{t+3}, s_{t+5}\} \rightarrow \{a_{t}, a_{t+2}, r_{t+3}, a_{t+4}\}$}\\
\bottomrule
\end{tabular}
}
\end{subtable}

\end{center}

\label{tab:top_losses_quadruped_run_state}
\end{table}

\end{document}